\documentclass{article}
\usepackage{iclr2027_conference,times}

\usepackage{amsmath,amsfonts,bm}

\def\eqref#1{equation~\ref{#1}}
\renewcommand{\eqref}[1]{Eq.~(\ref{#1})}

\def\1{\bm{1}}

\DeclareMathAlphabet{\mathsfit}{\encodingdefault}{\sfdefault}{m}{sl}
\SetMathAlphabet{\mathsfit}{bold}{\encodingdefault}{\sfdefault}{bx}{n}

\usepackage{hyperref}
\usepackage{url}
\usepackage{graphicx}
\usepackage{bbm}
\usepackage[table]{xcolor}
\usepackage{amsmath}       
\usepackage{amssymb}       
\usepackage{amsthm}        
\usepackage{subcaption}
\usepackage{algorithm}     
\usepackage{algpseudocode} 

\newtheorem{definition}{Definition}

\iclrpreprintcopy

\usepackage{booktabs}      

\title{Spectral Signatures of Memorization Basins in Diffusion Trajectories}

\author{
Achref Jaziri\textsuperscript{*} \\
Goethe University Frankfurt \\
Frankfurt am Main, Germany \\
\And
Martin Mundt \\
University of Bremen \\
Bremen, Germany \\
\And
Ahmed Imtiaz Humayun\textsuperscript{$\dagger$} \\
Google \\
New York, NY, USA \\
\And
Visvanathan Ramesh\textsuperscript{$\dagger$} \\
Goethe University Frankfurt \\
Frankfurt am Main, Germany
}

\begin{document}

\maketitle
\begingroup
\renewcommand{\thefootnote}{}
\footnotetext{%
\textsuperscript{*}Corresponding author: jaziri@em.uni-frankfurt.de.
\textsuperscript{$\dagger$}shared last authorship.
}
\endgroup
\begin{abstract}
Diffusion models are known to reproduce individual examples from their training data as a form of \textit{memorization} with implications for generalization, privacy, and downstream use. 
Recent work demonstrated that memorized training samples reside inside contiguous regions of the data/latent space known as \textit{basins of memorization} -- once a denoising trajectory enters a basin, under finite perturbations it irreversibly reproduces a memorized example.
Prior empirical work on such basins in trained models has primarily focused on classifier-free guidance based detection and mitigation strategies in text-to-image models, leaving basin entry in trained diffusion trajectories and the formation of such basins largely uncharacterized. 
In this paper, we present a geometric characterization of memorization basins for conditional, unconditional, and rectified-flow-based diffusion models. 
We observe a characteristic signature in the local spectrum of the function geometry that distinguishes basin entry: the emergence of a dominant singular mode followed by a sharp spectral recovery.
This signature appears at two timescales: along denoising trajectories during inference and over the course of training, with a markedly stronger presence along memorized trajectories.
We further investigate whether these geometric signatures can be used to intervene on memorization.  During training, we suppress memorization by acting on the dominant mode. At inference, we develop methods to identify trajectories entering memorization basins and steer them away. We find that these interventions remain effective even with low-dimensional approximations of the underlying geometry, and demonstrate that geometry-informed steering extends to large-scale text-to-image diffusion models. Together, our results corroborate that local spectral geometry provides a common basis for characterizing memorization and for suppressing, detecting, and steering away from it across training and generation.
\end{abstract}


\section{Introduction}
\label{sec:introduction}
Generative models are increasingly deployed as components of larger
computational systems, where their outputs inform subsequent predictions,
optimization, or decisions. Diffusion models exemplify this shift as they are deployed to  generate molecular and protein structures
\citep{wu2024protein}, candidate solutions for combinatorial optimization
\citep{sun2023difusco}, and actions and trajectories for planning and control
\citep{chi2025diffusion}.

As diffusion models take on this broader role, it is important to understand how much information from their training data is retained in their generated samples. Diffusion models can closely reproduce individual training examples \citep{carlini2023extracting}, raising concerns around privacy, copyright, and generalization. However, memorization is not inherently undesirable. More broadly in deep learning, retaining information about individual or atypical training examples can contribute positively to generalization \citep{feldman2020neural}. In diffusion models, memorization is also highly selective. These models reproduce only a subset of their training examples, with susceptibility to memorization depending on properties of both the data and the learned distribution \citep{carlini2023extracting,bonnaire2026diffusion}.

More generally, these empirical observations regarding memorization and memory in deep learning have been connected to model geometry. Previous work has shown that retained training information can be reflected in local model geometry \citep{nickl2023memory} and captured through compact spectral representations \citep{jung2026compact}. In generative models, the manifold memorization hypothesis further associates memorized samples with locally constrained regions of the learned distribution \citep{ross2025geometric}.

In this work, we posit that the constrained local geometry associated with memorization have a distinct temporal signature as trajectories evolve during denoising. We further investigate whether this geometry can be used to identify and intervene on memorized trajectories.  More specifically, we ask: \emph{when does a denoising trajectory become committed to a memorized outcome, what spectral signature emerges, and can it guide intervention?}

We adopt the framework of \emph{memorization basins}, defining them as regions from which a trajectory cannot escape a memorized outcome under bounded perturbations. Around basin entry, we empirically observe a distinct signature in the local Jacobian spectrum, accompanied by changes in perturbation sensitivity and generative dynamics. A related spectral pattern also emerges during training. At intermediate denoising steps, the input-output Jacobians develop a dominant singular mode, whereas low-noise spectra exhibit stronger rank recovery.

We further investigate whether these geometric properties can be used to suppress memorization during training and to detect or redirect memorized trajectories during generation. To this end, we propose GeoReg-DM, which regularizes local Jacobian geometry to suppress memorization during training. For inference, we introduce \emph{Jacobian Effective Rank} (JER), which uses the endpoint of the spectral signature to detect memorized generations, and GeoSteer-DM, which exploits dominant singular directions to redirect memorized trajectories.

Across diffusion and rectified-flow models, our results establish local spectral geometry as a unifying perspective on how memorization emerges and can be controlled across training and generation. 
\section{Related Work}

\paragraph{Memorization and geometry in generative models.}
Memorization is a longstanding phenomenon in overparameterized neural
networks, which can retain individual or atypical training examples while
still generalizing \citep{feldman2020neural}. In generative models,
memorization becomes directly observable through training-example reproduction
\citep{carlini2023extracting,somepalli2023diffusion}, with subsequent work
studying how it develops during diffusion training and how retained examples
are represented within the model
\citep{bonnaire2026diffusion,hintersdorf2024finding,kowalczuk2025finding}.
Local geometry provides a complementary perspective on generative
distributions \citep{meng2021estimating}, including intrinsic dimensionality,
manifold structure, and spectral regimes across diffusion time
\citep{stanczuk2024diffusion,humayun2025secrets,ventura2025manifolds}.
Memorization has specifically been associated with locally constrained or
sharp regions of the learned distribution
\citep{ross2025geometric,jeon2024understanding}, while related analyses study
denoiser-Jacobian eigenstructure \citep{kadkhodaie2024generalization},
variance-dependent loss of tangent dimensions \citep{achilli2024losing}, and
log-probability anisotropy \citep{asthana2026detecting}. We build on these
analyses by resolving spectral geometry along individual trajectories and
across training, and by using its singular directions for intervention.

\paragraph{Memorization mitigation and geometric control.}
Existing mitigation methods intervene during training through the data or
optimization procedure \citep{dutt2025memcontrol,shah2025does}, or during
generation through conditioning, attention, model predictions, initial noise,
and internal model components
\citep{somepalli2023diffusion,ren2024unveiling,zhao2026you,
wen2024detecting,han2026adjusting}. Separately, geometric editing methods use
Jacobian singular directions and learned subspaces to control generation
\citep{chen2024exploring,gaintseva2026midsteer,gaintseva2026casteer}.
Our approach connects these perspectives by using memorization-associated
local spectral geometry as an intervention target during training and as a
signal for detection and control during generation.

\section{THE DYNAMICS AND GEOMETRY OF MEMORIZATION}

\subsection{Memorization Basins}
\label{subsec:basins}
    
The manifold memorization hypothesis \citep{ross2025geometric} associates memorization with locally constrained regions of the learned distribution that restrict the effective degrees of freedom of generated samples. Statistical-physics analyses further identify a collapse transition beyond which reverse trajectories become attracted to individual training samples \citep{achilli2024losing}. Building on this picture, we characterize the denoising trajectory dynamics for a fixed learned model and sampler. Specifically, for perturbations bounded at each sampling step, we define a \emph{memorization basin} as a region from which no admissible continuation of a denoising trajectory can avoid memorization.

Let $\Phi_{t\rightarrow 0}:\mathbb{R}^{d}\rightarrow\mathbb{R}^{d}
$ denote the continuation of a fixed deterministic reverse process from
timestep $t$ to its final output, and let
$\mathcal{M}\subset\mathbb{R}^{d}$ denote the set of outputs classified as
memorized.

To formalize basin entry, we consider bounded perturbations to the remaining
reverse process. Let $T=t_N>\cdots>t_0=0$ denote the sampling schedule, and let $F_k$ map the state at $t_k$ to the next state at $t_{k-1}$. Starting from $x_t$, we allow each remaining
transition to be perturbed as
\[
x_{k-1}=F_k(x_k)+u_k,
\qquad
\|u_k\|_2\leq\delta.
\]
We denote the set of all such perturbation sequences by
$\mathcal U_t(\delta)$ and the resulting terminal output by
$\Phi^u_{t\rightarrow0}(x_t)$. The per-step bound $\delta$ therefore specifies the
magnitude of the admissible change at each remaining sampling step.

\begin{definition}[Memorization basin]
\label{def:basin}
For timestep $t$ and perturbation bound $\delta\geq 0$, the
\emph{memorization basin} is
\[
\mathcal{B}_t(\delta)
=
\left\{
x_t\in\mathbb{R}^{d}
:
\Phi^{u}_{t\rightarrow 0}(x_t)\in\mathcal{M}
\quad
\text{for all }u\in\mathcal{U}_t(\delta)
\right\}.
\]
\end{definition}

Thus, $x_t\in\mathcal{B}_t(\delta)$ when no continuation within the per-step bound  can escape $\mathcal M$. Basin membership is defined relative to the fixed sampler, sampling schedule, and bound $\delta$. Importantly, these basins are \emph{absorbing}: if an admissible continuation carries $x_t\in\mathcal{B}_t(\delta)$ to a later state $x_s$ in sampling order, then $x_s\in\mathcal{B}_s(\delta)$ (Appendix~\ref{app:basin_absorption}). Otherwise, an escape from $x_s$ could be concatenated with the preceding continuation to construct an escape from $x_t$. Basin entry therefore marks a persistent transition beyond which memorization cannot be avoided within bound $\delta$. This serves as the formal notion of commitment against which we evaluate the spectral signatures below.

\subsection{Spectral Geometry of Memorization}
\label{subsec:geometric_objects}
While membership in $\mathcal{B}_t(\delta)$ provides a rigorous condition for trajectory absorption,  evaluating it directly requires exhaustive perturbations of the remaining trajectory and denoising each of the perturbed trajectories. We therefore seek an easy to observe signature of a trajectory ``falling’’ into a memorization basin. Prior work shows that deep neural networks including diffusion models learn smooth functions which can be locally characterized using their input–output Jacobians \citep{balestriero2018spline,humayun2025secrets}. The Jacobian spectrum of denoisers have also previously been used to study inductive biases and generalization \citep{kadkhodaie2024generalization}. 
If $x_t\in\mathbb{R}^d$ denotes an intermediate state of a trajectory with marginal density $p_t$, the local geometry of $p_t$ along the trajectory can be characterized using the Jacobian of the score function which is equivalent to the Hessian of the log-density
$ H_t(x_t) = \nabla_{x_t}s_t(x_t) = \nabla_{x_t}^2\log p_t(x_t). $ 
\citet{jeon2024understanding} characterized memorization using the trace of the Hessian to identify regions with highly localized density.
We, however, are interested in the rate of change in the log-density landscape along different data space directions during denoising  -- which can be characterized using the Jacobian spectrum of the score function. 

For a trained denoiser approximating the score $s_t(x_t)$ via the forward process $x_t = \alpha_t x_0 + b_t \epsilon$, Tweedie's formula expresses the clean-estimate Jacobian as an affine function of the Hessian, $\frac{\partial \hat{x}_0}{\partial x_t} = \alpha_t^{-1}\big(I + b_t^2 H_t(x_t)\big)$. In an $\epsilon$-parameterized network, this relationship simplifies to:
\[
J_t = \frac{\partial \epsilon_\theta(x_t, t)}{\partial x_t} \approx -b_t H_t(x_t).
\]
with equality for the optimal model. Appendix~\ref{app:rf_curvature} derives the corresponding operator for rectified flow.
We analyze the directional structure of $J_t$ through its singular-value
decomposition $J_t=U_t\Sigma_tV_t^\top$, with $\sigma_1(t)\geq\cdots\geq\sigma_r(t)\geq0$, and summarize its spectral concentration and dimensionality using relative spectral gap, stable rank, and effective rank (defined in Appendix~\ref{app:spectral_behavioral_metrics}). Because $J_t$ serves as a curvature proxy, we also verify that it is nearly symmetric and sign-coherent near the data endpoint (empirically verified in Appendix~\ref{app:signed_curvature}).  

\begin{figure*}[t]
\centering

\begin{minipage}[t]{0.32\textwidth}
    \centering
    \includegraphics[width=\linewidth]
    {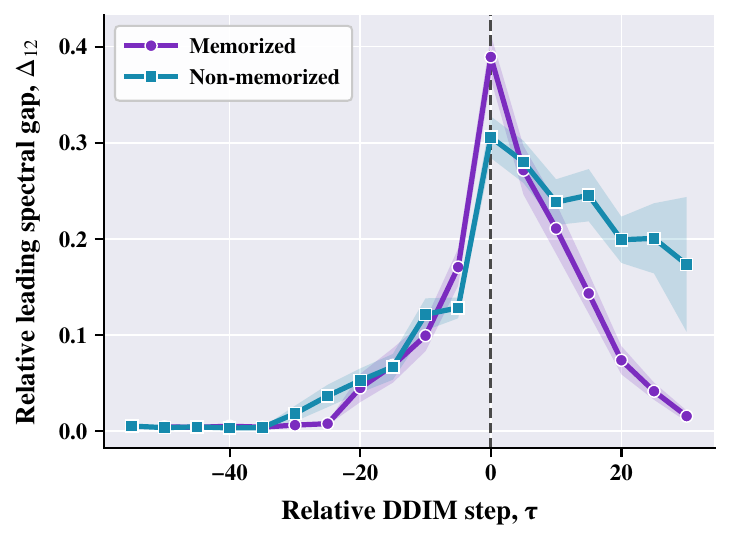}
\end{minipage}
\begin{minipage}[t]{0.32\textwidth}
    \centering
    \includegraphics[width=\linewidth]
    {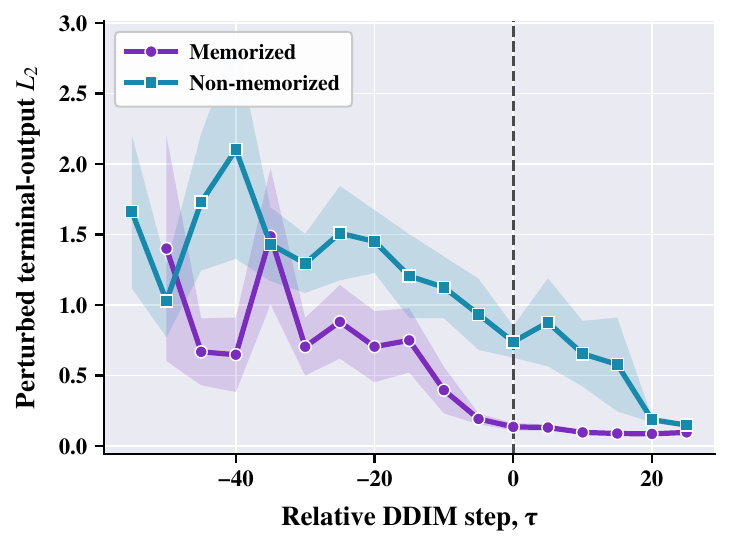}
\end{minipage}
\hfill
\begin{minipage}[t]{0.32\textwidth}
    \centering
    \includegraphics[width=\linewidth]
    {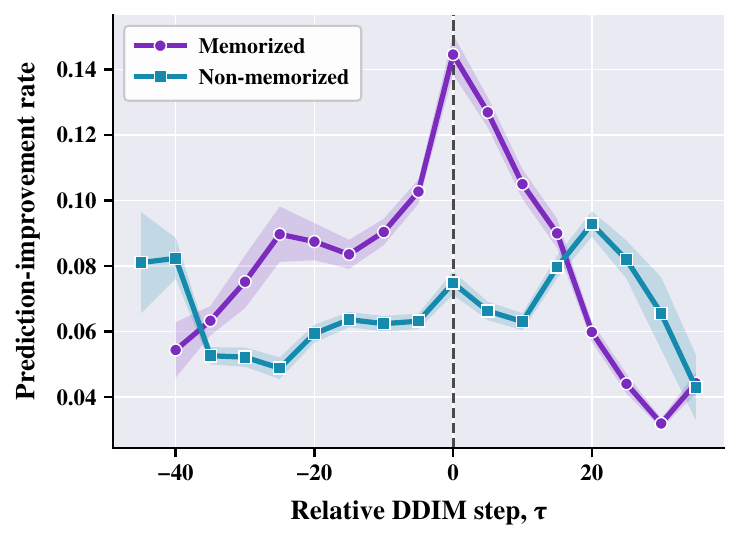}
\end{minipage}
\caption{\textbf{Spectral and behavioral signatures of memorization during inference.}
DDIM trajectories on CelebA are aligned at $\tau=t-t^*$, with $t^*$ defined
by the steepest increase in the leading spectral gap.
\textbf{(a)} Both groups exhibit spectral concentration around $\tau=0$,
with stronger concentration and subsequent recovery for memorized trajectories.
\textbf{(b)} Memorized trajectories become less sensitive to perturbations
after the transition.
\textbf{(c)} Their predicted clean samples approach the final output most
rapidly around the same region.}
\label{fig:alignment}

\end{figure*}

\paragraph{Spectral collapse and recovery of denoising trajectories.}
We adopt the CelebA setting of \citet{bonnaire2026diffusion}, training a U-Net diffusion model on 4,096 images and generating samples with DDIM. We classify a generation as memorized using the nearest-neighbor distance-ratio criterion of \citet{yoon2023diffusion}.  Along all denoising trajectories, the local Jacobian spectrum undergoes a phase transition, visible as a sharp rise in the relative spectral gap $\Delta_{12}$ (Figure~\ref{fig:alignment}a). In terms of stable rank, this transition appears as a spectral collapse followed by recovery (Appendix~\ref{app:singular_values}). The dynamic is driven largely by the top singular value. During collapse, $\sigma_1$ overshoots the remaining singular values and concentrates the spectrum around a single dominant mode. During recovery, the remaining singular values catch up and the spectrum relaxes toward a more isotropic structure as generation approaches low noise. Memorized trajectories undergo a stronger version of this cycle, with a sharper overshoot of $\sigma_1$ followed by a more pronounced recovery. As a result, $\Delta_{12}$ separates memorized from non-memorized trajectories at inference time (Figure~\ref{fig:alignment}a).

Because individual trajectories undergo collapse--recovery at different timesteps, we define $t^*$ as the state preceding the steepest increase in $\Delta_{12}$ and align trajectories by $\tau = t - t^*$. After $t^*$, memorized trajectories become substantially less sensitive to finite perturbations, and their predicted clean samples rapidly approach the final output (Figure~\ref{fig:alignment}b--c). Over the same region, bounded adversarial perturbations progressively lose their ability to redirect trajectories away from memorized outcomes (Appendix~\ref{app:adversarial_escape}). Together, these results suggest that the spectral transition at $t^*$ is an observable marker of declining escapeability, i.e., of the region where trajectories enter the memorization basin. 

\paragraph{Connecting dynamical collapse to local memorization geometry.} This collapse–recovery dynamic bridges trajectory-level views of memorization \citep{biroli2024dynamical,achilli2024losing} with local geometric accounts \citep{ross2025geometric,stanczuk2024diffusion}. While the geometry near a memorized endpoint approaches an isotropic $I/b_t$ structure (Appendix~\ref{app:curvature_uncertainty}), trajectories reach this regime by first passing through a phase of strong spectral concentration around basin entry. This intermediate regime isolates a dominant singular direction near $t^*$, providing the geometric direction that \textsc{GeoSteer-DM} later exploits to redirect memorized trajectories.

\begin{figure*}[h]
\centering
\begin{minipage}[t]{0.47\textwidth}
\centering
\includegraphics[width=\linewidth]
{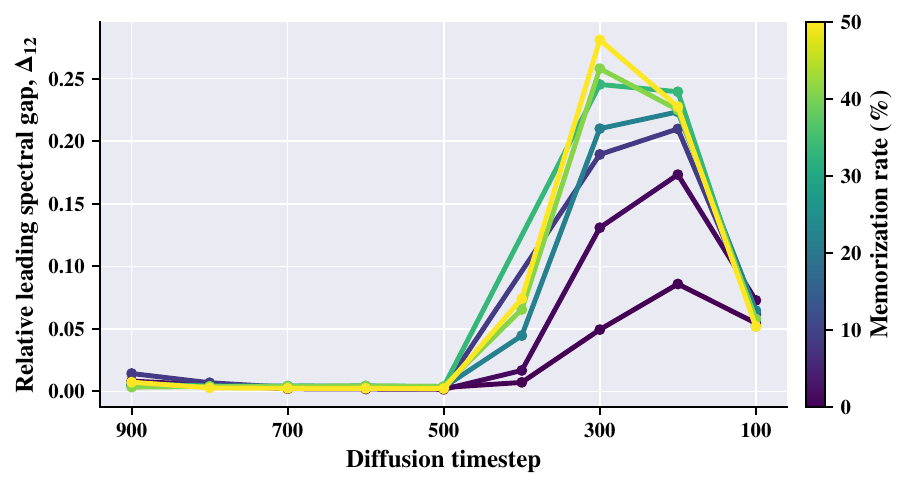}
\end{minipage}
\hfill
\begin{minipage}[t]{0.47\textwidth}
\centering
\includegraphics[width=\linewidth]
{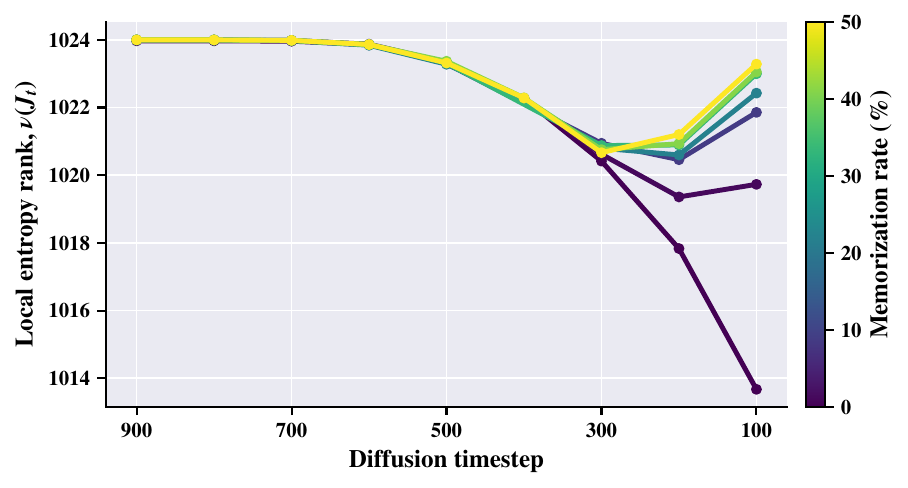}
\end{minipage}

\caption{\textbf{Training-time evolution of spectral concentration.}
Local Jacobian metrics for different training checkpoints colored by memorization rate.
\textbf{(a)} The relative leading spectral gap $\Delta_{12}$ rises sharply at intermediate noise, and falls again at low noise. Its peak grows with memorization rate and shifts toward higher noise in later checkpoints.
\textbf{(b)} Entropy effective rank reveals matching intermediate concentration followed by low-noise rank recovery in later checkpoints.}
\label{fig:train}

\end{figure*}

\paragraph{Spectral concentration emerges during training.}
Furthermore, the  geometric organization observed during denoising develops progressively during training (Figure~\ref{fig:train}). As training advances and population-level memorization increases, the local Jacobian becomes increasingly concentrated around a dominant mode at intermediate noise levels, reflected by a pronounced rise in the relative spectral gap $\Delta_{12}$ around $t\in[200,400]$ that grows with memorization rate (Figure~\ref{fig:train}a). At lower noise levels ($t\leq200$), the spectral gap closes and the local entropy rank $\nu(J_t)$ recovers toward a less concentrated spectrum at later checkpoints (Figure~\ref{fig:train}b). These observations provide a  training-time regularization target, directly motivating our approach \textsc{GeoReg-DM}.

\section{CONTROLLING MEMORIZATION THROUGH SPECTRAL GEOMETRY}

Our geometric analysis highlights two key points of intervention: regularizing the dominant intermediate-noise mode during training, and using local singular directions to detect and steer trajectories at inference time. We translate these insights into three complementary mechanisms: \textsc{GeoReg-DM} for training, JER for detection, and \textsc{GeoSteer-DM} for inference-time steering.

\subsection{GeoReg-DM: Spectral Regularization During Training}
\label{sec:georeg}

We introduce \textsc{GeoReg}, which suppresses the dominant local mode during training by penalizing the squared leading singular value $\sigma_1$ of a low-dimensional Jacobian sketch. Since computing $\sigma_1$ from the full $d\times d$ Jacobian at every
training step is prohibitively expensive, we  use a
low-dimensional sketch
$\widehat J_t=WJ_t$, where $W\in\mathbb{R}^{k\times d}$ has orthonormal
rows. The sketch is constructed directly through vector--Jacobian products
without materializing $J_t$ \citep{halko2011finding,woodruff2014sketching}.  Because $\widehat J_t$ has only
$k$ rows,  we could compute the leading squared singular value from the largest  eigenvalue of the $k\times k$ Gram matrix
$\widehat J_t\widehat J_t^\top$.

We apply the regularizer to a selected subset $\mathcal I$ of the minibatch,
\begin{equation}
\mathcal L_{\mathrm{geo}}
=
\frac{1}{|\mathcal I|}
\sum_{i\in\mathcal I}
\sigma_1\!\left(\widehat J_{t_i}^{(i)}\right)^2,
\label{eq:georeg_objective}
\end{equation}
Differentiating through the Jacobian sketch introduces mixed derivatives
$\partial^2\epsilon_\theta/(\partial\theta\,\partial x)$. To keep this
computation tractable, we regularize only a small subset of each minibatch,
selected using the input sensitivity of the ordinary denoising loss,
$\pi^{(i)}
=
\left\lVert
\nabla_{x_{t_i}^{(i)}}
\ell_{\mathrm{base}}^{(i)}
\right\rVert_2.
\label{eq:georeg_priority}
$

This quantity is simple to compute from the base-loss backward without constructing Jacobian sketches. After that, we allocate the geometric computation to the states with highest sensitivity. Full implementation details are given in Appendix~\ref{app:georeg}.

\subsection{JER: Efficient Detection from Local Spectral Geometry}
\label{sec:jer}
The spectral transitions observed in our analysis suggest that stable and effective rank provide a direct geometric signal for memorization detection. However, both depend on the Jacobian spectrum and are expensive to evaluate at inference time. We therefore propose the \emph{Jacobian Effective Rank} (JER), a local proxy for spectral dimensionality designed to retain this geometric signal without explicitly computing the Jacobian spectrum. To obtain an efficient estimator, we leverage a connection to R\'enyi-2 effective rank.

For a Jacobian $J$ with singular values $\{\sigma_i\}$, the R\'enyi-2 effective rank is
\begin{equation}
\operatorname{ER}_2(J) = \frac{\left(\sum_i \sigma_i\right)^2}{\sum_i \sigma_i^2} = \frac{\|J\|_*^2}{\|J\|_F^2}.
\label{eq:jer_exact}
\end{equation}
This quantity provides a useful bridge between the two rank measures underlying our geometric analysis: it satisfies $\operatorname{SR}(J) \leq \operatorname{ER}_2(J) \leq \nu(J)$ (Appendix~\ref{app:jer_derivation}), while depending only on the nuclear and Frobenius norms rather than the full singular-value distribution.

To approximate the remaining nuclear norm, we leverage our signed-curvature analysis (Appendix~\ref{app:signed_curvature}), which shows that the curvature spectrum is strongly sign-coherent near the data endpoint. We exploit this structure by replacing the nuclear norm with the oriented trace.

Building upon these approximations, we evaluate JER without constructing the Jacobian by leveraging stochastic trace estimation with Rademacher probes \citep{hutchinson1989stochastic}. For independent $z_1,\ldots,z_m$ and vector–Jacobian products $q_\ell = J^\top z_\ell$, we estimate $\widehat{\operatorname{tr}}_m(J) = m^{-1}\sum_\ell z_\ell^\top q_\ell$ and $\widehat F_m(J) = m^{-1}\sum_\ell \|q_\ell\|_2^2$, giving
\begin{equation}
S_{\mathrm{JER}}(J) = \frac{\left[a\,\widehat{\operatorname{tr}}_m(J)\right]_+^2}{\widehat F_m(J) + \varepsilon},
\label{eq:jer_score}
\end{equation}
where $a \in \{-1, +1\}$ selects the curvature orientation. At inference, JER is evaluated once at a fixed low-noise query state and thresholded using a calibration set. Details of the estimator and query construction are given in Appendix~\ref{app:jer_estimation}.

\subsection{GeoSteer-DM: Geometry-Informed Steering During Generation}
\label{sec:geosteer}

GeoSteer-DM  uses the local geometry to redirect a selected trajectory during
denoising. The spectral analysis above shows that memorization is accompanied
by the emergence of a dominant local Jacobian mode. We posit that steering the model away from this mode diminishes memorization. We use the leading
right singular vector as the minimal direction on which to intervene and redirect the trajectory
away from memorization.

Since different generative models parameterize their updates differently, we formulate our approach in terms of model-induced displacement. This allows us to generalize our method to DDIM (which predicts noise), Rectified Flow (which predicts velocity), and Classifier-Free Guidance (which combines conditional and unconditional predictions).

At sampler step $n$, let $x_{n+1}^{\mathrm{base}}$ denote the ordinary update
and $x_{n+1}^{0}$ a corresponding anchor update from the same current state.
We define the model-controlled displacement as
\begin{equation}
\delta_n
=
x_{n+1}^{\mathrm{base}}
-
x_{n+1}^{0}.
\label{eq:geosteer_delta_main}
\end{equation}
For unconditional DDIM and rectified flow, the anchor uses zero model output;
under classifier-free guidance, it uses the unconditional prediction, so that
$\delta_n$ isolates the contribution of conditional guidance. This provides
a common displacement to define our steering rule for different settings (sampler-specific formulations are given in
Appendix~\ref{app:geosteer}).

As a simple geometry-informed steering rule, we act on the leading right
singular direction of the local Jacobian. This direction captures the dominant local spectral mode and we use it to construct a minimal intervention that modifies only the dominant component.  Let
$J_{t_n}=U_{t_n}\Sigma_{t_n}V_{t_n}^{\top}$ and let $v_1(t_n)$ denote its
leading right singular vector. We define the corresponding projector
\[
P_n=v_1(t_n)v_1(t_n)^\top
\]
and decompose the model-controlled displacement as
\[
\delta_n
=
P_n\delta_n+(I-P_n)\delta_n.
\]
The two terms isolate the component along the leading geometric direction and
its orthogonal complement, respectively.

We choose a reflection as a simple norm-preserving intervention on this component,
\begin{equation}
\delta_n^{\mathrm{geo}}
=
(I-2P_n)\delta_n,
\qquad
x_{n+1}^{\mathrm{geo}}
=
x_{n+1}^{0}
+
\delta_n^{\mathrm{geo}}.
\label{eq:geosteer_reflection_main}
\end{equation}
This reflection is invariant to the sign of $v_1(t_n)$ and preserves the
displacement norm,
$\|\delta_n^{\mathrm{geo}}\|_2=\|\delta_n\|_2$,
so steering changes the direction of the model-induced update without
attenuating its magnitude.

\section{EXPERIMENTS}
\label{sec:experimental_setup}

We evaluate our framework in two complementary settings. We first train unconditional diffusion and rectified-flow models on a controlled CelebA subset, where memorization is measured by the nearest-neighbor gap ratio to the training set. We then test JER and GeoSteer-DM on Stable Diffusion v1.4, measuring memorization by SSCD similarity to known training references, and compare against existing inference-time baselines\footnote{Code will be released upon acceptance.}. Complete experimental details are found in Appendix~\ref{app:experiments}.
\subsection{Suppression, Detection, and Steering in unconditional diffusion}
\label{sec:results_georeg}

In the unconditional diffusion setting, we use our geometric framework to address three core questions: whether \textsc{GeoReg-DM} suppresses memorization while preserving coverage, whether JER detects memorized trajectories efficiently, and whether \textsc{GeoSteer-DM} redirects them.

\paragraph{Can GeoReg-DM suppress memorization without sacrificing distribution coverage?}
GeoReg-DM regularizes the local Jacobian geometry during training to suppress emerging memorization. We first evaluate it on three synthetic populations (Gaussian, Gaussian mixture, and 8D Swiss roll), where access to the true data distribution allows us to distinguish training-set copying from population coverage. Across all three settings, GeoReg-DM achieves the strongest reduction in copying while consistently preserving or improving recall (Table~\ref{tab:georeg_results}).  

This effect extends directly to image generation on CelebA32, where GeoReg-DM substantially reduces memorization and outperforms other approaches on both measures (Table~\ref{tab:georeg_results}). This decrease in memorization comes with only a modest increase in FID, which is expected in this setting: removing exact training-set reproductions can itself increase FID even when perceptual quality and sample diversity are preserved. We investigate this effect further, together with representative generations and additional perceptual-quality analyses, in Fig.\ref{fig:georeg_quality} and Appendix\ref{app:georeg_quality}. Together with the synthetic results, these findings show that GeoReg-DM suppresses memorization without simply contracting the learned distribution, and that this behavior transfers from controlled populations to image generation. To further highlight the impact of our regularization approach, we analyze the spectral geometry of the Jacobian after GeoReg regularization. We observe that the regularization affects the entire spectrum. Specifically, GeoReg significantly slows down concentration in the top singular value, as evidenced by  Figure~\ref{fig:georeg_effective_rank}.

\begin{table*}[ht]
\centering
\caption{\textbf{GeoReg reduces memorization across synthetic populations and
CelebA32.}
Results are averaged across five independent  seeds for each setting.
For synthetic populations, memorization is measured against the fixed training
set, while precision and recall are evaluated against independent population
samples. For CelebA32, $r$ denotes the nearest-neighbor gap ratio. Full
baseline definitions, training settings, and additional metrics are given in
Appendix~\ref{app:georeg_baselines}.}
\label{tab:georeg_results}
\small
\setlength{\tabcolsep}{2.8pt}
\resizebox{\textwidth}{!}{%
\begin{tabular}{lccc|ccc|ccc|ccc}
\toprule
& \multicolumn{3}{c|}{Gaussian80}
& \multicolumn{3}{c|}{Mixture32}
& \multicolumn{3}{c|}{SwissRoll8D128}
& \multicolumn{3}{c}{CelebA32} \\
\cmidrule(lr){2-4}
\cmidrule(lr){5-7}
\cmidrule(lr){8-10}
\cmidrule(lr){11-13}
Method
& Mem $\downarrow$ & Prec. $\uparrow$ & Rec. $\uparrow$
& Mem $\downarrow$ & Prec. $\uparrow$ & Rec. $\uparrow$
& Mem $\downarrow$ & Prec. $\uparrow$ & Rec. $\uparrow$
& Mem $\downarrow$ & Mean $r$ $\uparrow$ & FID $\downarrow$ \\
\midrule

Standard
& 31.85 & 0.978 & 0.807
& 46.85 & 0.997 & 0.757
& 16.10 & \textbf{0.928} & 0.816
& 8.80 & 0.766 & 54.89 \\

Tikhonov
& 25.15 & 0.981 & 0.921
& 43.25 & \textbf{0.998} & 0.812
& 13.90 & 0.918 & 0.839
& 6.50 & 0.780 & \textbf{49.03}  \\

Reduced LR
& 21.15 & \textbf{0.986} & 0.880
& 41.35 & 0.992 & 0.771
& 10.30 & 0.877 & \textbf{0.899}
& 6.80 & 0.785 & 52.21 \\

Ambient-style
& 24.35 & 0.985 & 0.858
& 47.00 & 0.993 & 0.600
& 17.30 & 0.917 & 0.789
& 4.90 & 0.808 & 52.15 \\

\rowcolor{gray!9}
GeoReg (Ours)
& \textbf{15.85} & 0.983 & \textbf{0.979}
& \textbf{19.25} & 0.980 & \textbf{0.963}
& \textbf{10.20} & 0.904 & 0.878
& \textbf{1.56} & \textbf{0.846} & 59.77 \\
\bottomrule
\end{tabular}%
}
\end{table*}

\paragraph{Local Jacobian geometry enables efficient post-hoc detection.}
As a baseline, we use temporal recovery, which follows directly from our spectral analysis and scores how strongly the spectrum recovers after its collapse along a trajectory. However, it requires full Jacobian spectra at multiple points along the trajectory, which makes it expensive. Comparing it with local terminal geometry on held-out CelebA trajectories shows that \textsc{JER} captures a strong memorization signal at substantially lower computational cost (Table~\ref{tab:jer_detection}). For DDIM, \textsc{JER} achieves the strongest overall detection, outperforming both local baselines and full temporal recovery. For rectified flow, temporal recovery  achieves the best detection accuracy, while our faster approach \textsc{JER} still outperforms all other local geometric detectors. The main distinction between JER and temporal recovery lies in the richer temporal information available to the latter. These results point to differences in how memorization is geometrically encoded across sampling methods. 

We also note that the spectral recovery baseline is intended only to validate our geometric observations, not as a practical method, since computing the geometry over the full trajectory is very slow in practice. Nevertheless, our empirical observations corroborate both our fast JER approach and our analyis from section \ref{subsec:geometric_objects}. 
. 


\begin{table*}[h]
\centering
\caption{\textbf{Efficient geometry-based memorization detection.}
Results on held-out CelebA trajectories. AUROC and AUPRC denote the
areas under the receiver-operating-characteristic and precision--recall
curves, respectively; MCC denotes the Matthews correlation coefficient.
Recall, specificity, and MCC are computed at operating points selected on a
disjoint calibration set and frozen before evaluation. For the unconditional
score-magnitude baseline, classifier orientation is also selected on the
calibration split and then frozen for held-out evaluation.}
\label{tab:jer_detection}
\small
\resizebox{\textwidth}{!}{%
\begin{tabular}{llcccccc}
\toprule
Sampler & Method
& AUROC $\uparrow$
& AUPRC $\uparrow$
& Recall $\uparrow$
& Spec. $\uparrow$
& MCC $\uparrow$
& Time $\downarrow$ \\
\midrule

DDIM
& FLIPD \cite{ross2025geometric}
& 0.845 & 0.722 & 0.437 & 0.932 & 0.443 & 42.05\,ms \\

& Normal bundle \cite{stanczuk2024diffusion}
& 0.902 & 0.773 & 0.452 & 0.952 & 0.493 & 922.19\,ms \\

\rowcolor{gray!8}
& Unconditional score magnitude
& 0.860 & 0.785 & 0.542 & 0.934 & 0.537 & \textbf{12.05\,ms} \\

\rowcolor{gray!8}
& Spectral temporal recovery
& 0.823 & 0.698 & 0.305 & \textbf{0.956} & 0.366 & 40.88\,s \\

\rowcolor{gray!8}
& \textbf{JER (ours)}
& \textbf{0.956}
& \textbf{0.895}
& \textbf{0.898}
& 0.886
& \textbf{0.763}
& 71.26\,ms \\

\midrule

RF
& FLIPD \cite{ross2025geometric}
& 0.720 & 0.733 & 0.310 & 0.894 & 0.248 & 42.48\,ms \\

& Normal bundle \cite{stanczuk2024diffusion}
& 0.760 & 0.746 & 0.338 & 0.897 & 0.278 & 922.12\,ms \\

\rowcolor{gray!8}
& Unconditional score magnitude
& 0.783 & 0.800 & 0.375 & 0.929 & 0.358 & \textbf{11.98\,ms} \\

\rowcolor{gray!8}
& Spectral temporal recovery
& \textbf{0.912}
& \textbf{0.931}
& \textbf{0.789}
& 0.879
& \textbf{0.667}
& 38.31\,s \\

\rowcolor{gray!8}
& \textbf{JER (ours)}
& 0.870
& 0.844
& 0.403
& \textbf{0.935}
& 0.392
& 57.41\,ms \\

\bottomrule
\end{tabular}%
}
\end{table*}

\begin{figure}[t]
    \centering

    \begin{minipage}[t]{0.38\linewidth}
        \centering
        \includegraphics[width=\linewidth]
        {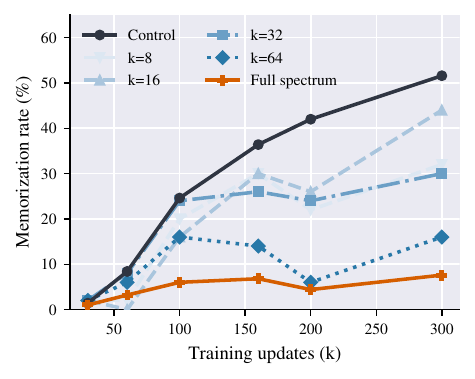}
    \end{minipage}
    \hspace{0.02\linewidth}
    \begin{minipage}[t]{0.38\linewidth}
        \centering
        \includegraphics[width=\linewidth]
        {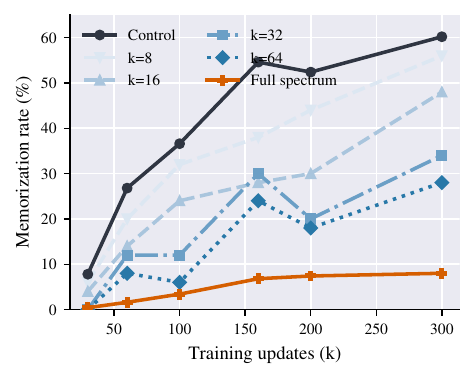}
    \end{minipage}

\caption{\textbf{\textsc{GeoSteer-DM} suppresses memorization across training checkpoints and geometric resolutions.}
Memorization rate of unmodified and steered sampling for (a) DDIM and (b) rectified flow. Full-Jacobian steering gives the strongest suppression. Low-dimensional Jacobian sketches retain much of this effect, with suppression generally strengthening as $k$ increases.}
    \label{fig:geosteer_resolution}
\end{figure}

\paragraph{Local geometry enables actionable, low-rank trajectory steering.}
Using the full Jacobian spectrum, \textsc{GeoSteer-DM} substantially suppresses
memorization across training checkpoints for both DDIM and rectified flow
(Figure~\ref{fig:geosteer_resolution}), reducing late-stage memorization from
$51.6\%$ to $7.6\%$ and from $60.2\%$ to $8.0\%$, respectively. Importantly,
effective intervention does not require the full Jacobian. Here we observe that low-rank
sketches with $k\in\{8,\ldots,64\}$ retain sufficient directional information to substantially reduce memorization, and performance improves as geometric resolution increases. At $k=64$, aggregate memorization falls to $10.0\%$ for
DDIM and $14.0\%$ for rectified flow. These results further corroborate that the spectral signature observed in our analysis is actionable for intervention, and that useful steering signals can be recovered without full Jacobian construction—a point we will highlight more extensively in the text-to-image diffusion experiments in the next section.

\subsection{Scaling Geometric Steering to Text-to-Image Diffusion}
\label{sec:results_stable_diffusion}

The controlled experiments establish that memorization geometry is both detectable and actionable, and that effective intervention directions can be recovered under low-rank approximations. We now test whether these properties scale to Stable Diffusion v1.4, where text conditioning and high dimensionality make full Jacobian construction infeasible. Using matrix-free estimates of the guidance-residual Jacobian (Appendix~\ref{app:geosteer_geometry}), we evaluate \textsc{GeoSteer-DM} across 500 prompts with known memorized references (2,000 generations per setup).

\paragraph{Geometric steering scales to text-to-image diffusion.}
Using matrix-free estimates of local guidance geometry, \textsc{GeoSteer-DM} substantially reduces training-reference similarity and the fraction of generations exceeding the SSCD copying threshold, while  generation quality and a modest reduction in prompt alignment (Figure~\ref{fig:sd_pareto}; full metrics in Appendix~\ref{app:sd_nonmemorized_control}). This demonstrates that local geometric steering scales to large-scale text-conditioned models. Representative samples in Figure~\ref{fig:sd_qualitative} illustrate this and show that steering suppresses memorization while preserving core prompt semantics.

\paragraph{\textsc{GeoSteer-DM} complements existing mitigation strategies.}
Unlike existing methods that intervene via text conditioning, attention, model predictions, or initial noise, \textsc{GeoSteer-DM} acts directly on local Jacobian geometry. As shown in Figure~\ref{fig:sd_pareto}, adding \textsc{GeoSteer-DM} consistently shifts each baseline toward lower reference similarity with minimal impact on CLIP score or FID. Furthermore, these shifts are smaller on non-memorized control prompts (Appendix~\ref{app:sd_nonmemorized_control}), which suggests that \textsc{GeoSteer-DM} mainly targets memorization-associated geometry instead of just indiscriminately perturbing generative trajectories.

\begin{figure*}[h]
    \centering
    \includegraphics[width=0.75\textwidth]
    {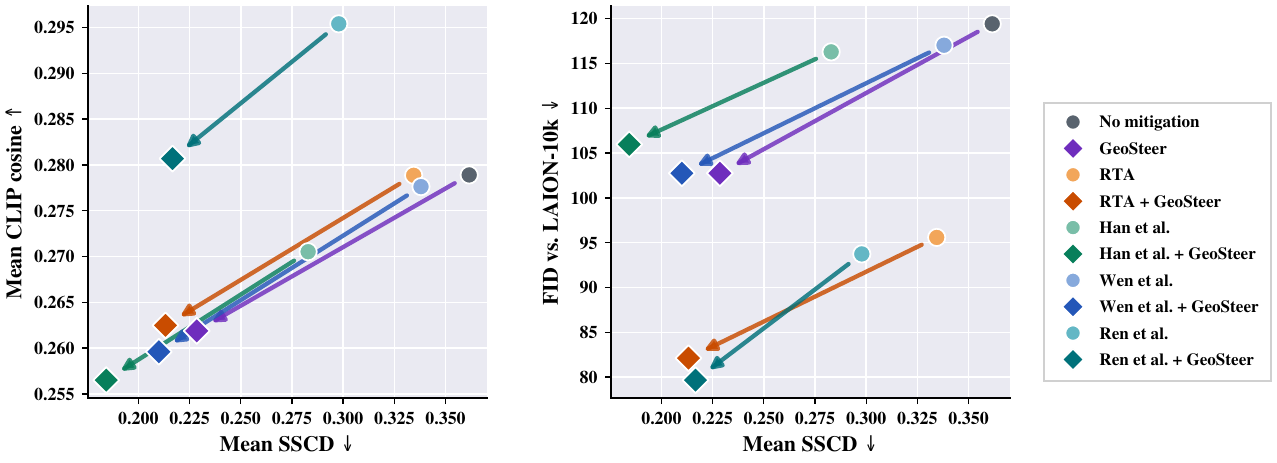}
    \caption{\textbf{\textsc{GeoSteer-DM} improves memorization--quality trade-offs across mitigation families.}
Mean SSCD to memorized training references versus CLIP prompt alignment (left) and FID (right) on memorized prompts. Connected markers show each method, including unmodified sampling, before and after adding \textsc{GeoSteer-DM}.}
    \label{fig:sd_pareto}
\end{figure*}
\section{Conclusion}

We characterized memorization in diffusion models through the local spectral geometry of the generative field. Across training and generation, memorization manifests through a distinct spectral collapse–recovery dynamic: as models memorize during training, an intermediate-noise dominant mode emerges; during generation, entering a memorization basin corresponds to a transient spectral collapse around this leading direction. Local spectral geometry thus provides a unified framework for identifying when and how trajectories become committed to memorized outcomes.
Acting on this geometric structure yields practical tools across the generative lifecycle: training-time spectral regularization, efficient detection, and inference-time trajectory steering, which scales to models as large as Stable Diffusion. While \textsc{GeoSteer-DM} employs a simple norm-preserving reflection along the leading singular mode, future work may leverage higher-order subspace projections or adaptive trajectory control. Overall, our findings demonstrate that local Jacobian geometry provides both a diagnostic and an actionable target for controlling memorization.

\section*{Acknowledgments}
We thank Eleni Triantafillou for the helpful insights throughout this project. We also acknowledge Hessian.AI and the Center for Cognition and Computation at Goethe University
for providing computational resources.



\bibliography{iclr2027_conference}
\bibliographystyle{iclr2027_conference}

\newpage
\appendix

\section*{Appendix Overview}

Appendix~A provides additional geometric derivations and diagnostics.
Appendix~B develops the JER formulation and detection protocol.
Appendix~C gives the formulation and implementation of GeoSteer-DM, and
Appendix~D provides the corresponding details for GeoReg-DM.
Appendix~E describes the experimental settings and evaluation protocols.
Appendix~F reports additional results, ablations, and controls.

\section{Additional Geometric Derivations and Diagnostics}
\label{app:geometry}

\subsection{Absorption and the Meaning of Basin Entry}
\label{app:basin_absorption}

The definition we rely on this work of $\mathcal B_t(\delta)$ in this work is intended to distinguish eventual memorization from \emph{commitment} to memorization. Specifically, we require basin membership to persist under admissible perturbations; otherwise, a trajectory could satisfy the condition at one timestep but leave the basin under a valid continuation.

The definition in Sec.~\ref{subsec:basins} ensures that basin membership is preserved along admissible trajectories. Specifically, consider $x_t\in\mathcal B_t(\delta)$ and a state $x_s=\Phi^w_{t\rightarrow s}(x_t)$ reached under an admissible perturbation sequence $w$, with $s<t$. Any admissible continuation $v\in\mathcal U_s(\delta)$ can be concatenated with $w$ to form an admissible trajectory from $x_t$. By definition of the basin, this trajectory must terminate in $\mathcal M$, and hence $\Phi^v_{s\rightarrow0}(x_s)\in\mathcal M$. Since this holds for any admissible continuation $v$, we have $x_s\in\mathcal B_s(\delta)$. Consequently, basin membership is preserved under admissible perturbations:
\begin{equation*}
\Phi^w_{t\rightarrow s}\bigl(\mathcal B_t(\delta)\bigr)
\subseteq \mathcal B_s(\delta),
\qquad s<t.
\end{equation*}

Thus, once a trajectory enters a memorization basin, no continuation within the allowed perturbation class can escape it. This makes basin entry a meaningful reference point against which we compare the spectral transition  $t^*$ and investigate whether the observed spectral changes precede, coincide with, or follow this notion of basin entry.

We would like also to highlight that the dependence on $\delta$ in our definition is also essential. The family
$\{\mathcal B_t(\delta)\}_{\delta\geq0}$ describes commitment at different
levels of intervention strength: increasing $\delta$ enlarges the admissible
set of continuations. In particular, for $\delta_1\leq\delta_2$,
\begin{equation*}
\mathcal B_t(\delta_2)
\subseteq
\mathcal B_t(\delta_1).
\end{equation*}
We will examine this
$\delta$-dependent loss of escapeability empirically in
Appendix~\ref{app:adversarial_escape}.

\subsection{Curvature-Equivalent Jacobian for Rectified Flow}
\label{app:rf_curvature}

In Sec.~\ref{subsec:geometric_objects}, we characterize local geometry through
the curvature of the time-dependent marginal density. For an
$\epsilon$-parameterized diffusion model, the input Jacobian of the model
prediction is directly proportional to the score Hessian. Rectified flow
requires an additional correction: its velocity field contains a linear
transport component whose input Jacobian is unrelated to density curvature.
We derive this correction here and show that, after removing it, the resulting
operator is proportional to the same log-density Hessian used in the diffusion
analysis.

Consider the standard linear rectified-flow path
\[
x_s
=
s x_0+(1-s)z,
\qquad
x_0\sim p_{\mathrm{data}},
\qquad
z\sim\mathcal{N}(0,I),
\qquad
s\in[0,1],
\]
where sampling proceeds from noise at $s=0$ to data at $s=1$. Let
$p_s$ denote the marginal density of $x_s$ and
\[
\mathbf{s}_s(x)
=
\nabla_x\log p_s(x)
\]
its score. The population-optimal rectified-flow velocity is
\[
v^*(x,s)
=
\mathbb{E}[x_0-z\mid x_s=x].
\]

For this Gaussian interpolation, Tweedie's identity gives
\[
\mathbb{E}[x_0\mid x_s=x]
=
\frac{x}{s}
+
\frac{(1-s)^2}{s}\mathbf{s}_s(x).
\]
Using
\[
v^*(x,s)
=
\frac{\mathbb{E}[x_0\mid x_s=x]-x}{1-s},
\]
we obtain
\[
v^*(x,s)
=
\frac{x}{s}
+
\frac{1-s}{s}\mathbf{s}_s(x),
\]
and therefore
\[
\mathbf{s}_s(x)
=
\frac{s\,v^*(x,s)-x}{1-s}.
\]
Differentiating with respect to the state yields
\[
H_s(x)
:=
\nabla_x^2\log p_s(x)
=
\frac{sJ_v^*(x,s)-I}{1-s},
\]
where
\[
J_v^*(x,s)
=
\frac{\partial v^*(x,s)}{\partial x}.
\]

This relation identifies the RF operator corresponding to density curvature.
For a learned velocity field $v_\theta$, we define
\begin{equation}
J_s^{\mathrm{RF}}
=
s\frac{\partial v_\theta(x,s)}{\partial x}-I.
\label{eq:rf_curvature_jacobian}
\end{equation}
At the population optimum,
\[
J_s^{\mathrm{RF}}
=
(1-s)H_s(x).
\]
Thus, up to a timestep-dependent scalar,   \eqref{eq:rf_curvature_jacobian}
recovers the same log-density Hessian that motivates the
$\epsilon$-Jacobian in diffusion. This is why the two model parameterizations
can be analyzed using the common notation $J_t$ in the main text.

\paragraph{Why the identity correction matters.}
The raw velocity Jacobian itself is not curvature-equivalent. Rearranging the
expression above gives
\[
J_v^*(x,s)
=
\frac{1}{s}I
+
\frac{1-s}{s}H_s(x).
\]
Its first term is induced by the linear transport path and is present
independently of the local curvature of $p_s$. Using $J_v$ directly would
therefore mix this known parameterization-dependent component with the
geometric structure we intend to measure.

This distinction matters for the spectral quantities used throughout the
paper. Stable rank, relative spectral gap, effective rank, and right-singular
subspaces are invariant to multiplication by a nonzero scalar, but not to an
additive identity shift. Removing the identity component is therefore
necessary before comparing RF geometry with the corresponding diffusion
quantities.

For this reason, we do not additionally divide \eqref{eq:rf_curvature_jacobian} by $1-s$. The omitted factor changes
only the overall scale of the operator and leaves the spectral-shape
statistics and singular subspaces used in our analysis unchanged. At a
sampler state $x_t$ with physical RF time $s_t$, the notation $J_t$ in the
main text therefore refers to
\begin{equation}
J_t
=
s_t
\frac{\partial v_\theta(x_t,s_t)}{\partial x_t}
-I.
\label{eq:rf_main_operator}
\end{equation}
Unless stated otherwise, RF spectral quantities are computed from this
curvature-equivalent operator.

\paragraph{Efficient implementation.}
The correction does not require constructing either the full Jacobian or an
explicit $d\times d$ identity matrix. For the left sketch $W$ used by
GeoReg,
\[
WJ_t
=
s_tW
\frac{\partial v_\theta(x_t,s_t)}{\partial x_t}
-W.
\]
The transport component can therefore be removed directly in the sketched
space after computing the corresponding vector--Jacobian products. The same
principle applies to stochastic Jacobian contractions used by the JER
detector: for a probe $z$,
\[
J_t^\top z
=
s_t
\left(
\frac{\partial v_\theta(x_t,s_t)}{\partial x_t}
\right)^\top z
-z.
\]
Thus, GeoReg and JER can use the curvature-equivalent RF operator without
materializing $J_t$. We evaluate this operator only at interior flow times
$s_t\in(0,1)$.

\subsection{Signed Curvature and Jacobian Symmetry}
\label{app:signed_curvature}

The geometric analysis in Sec.~\ref{subsec:geometric_objects} follows from
the connection between the denoising field and the time-dependent data
density. For an $\epsilon$-parameterized diffusion model, the model Jacobian
used in the main text satisfies
\[
J_t
=
\frac{\partial\epsilon_\theta(x_t,t)}{\partial x_t}
\approx
-b_t H_t(x_t).
\]
Its singular values therefore characterize the magnitude and organization of
local curvature, but do not retain its sign.

For the signed-curvature diagnostic in this section, we orient the operator
toward the Hessian and define
\[
\widetilde J_t
=
-J_t
=
-\frac{\partial\epsilon_\theta(x_t,t)}{\partial x_t}.
\]
At the population optimum,
$\widetilde J_t\approx b_t H_t$, so it is proportional to the
log-density Hessian by a positive timestep-dependent scalar. We decompose
\[
\widetilde J_t
=
S_t+A_t,
\qquad
S_t
=
\frac{\widetilde J_t+\widetilde J_t^\top}{2},
\qquad
A_t
=
\frac{\widetilde J_t-\widetilde J_t^\top}{2},
\]
and measure the relative asymmetric component as
\[
\rho_{\mathrm{asym}}(t)
=
\frac{\|A_t\|_F}{\|\widetilde J_t\|_F}.
\]

\begin{figure*}[t]
    \centering

    \begin{minipage}[t]{0.32\textwidth}
        \centering
         \includegraphics[width=\linewidth]
        {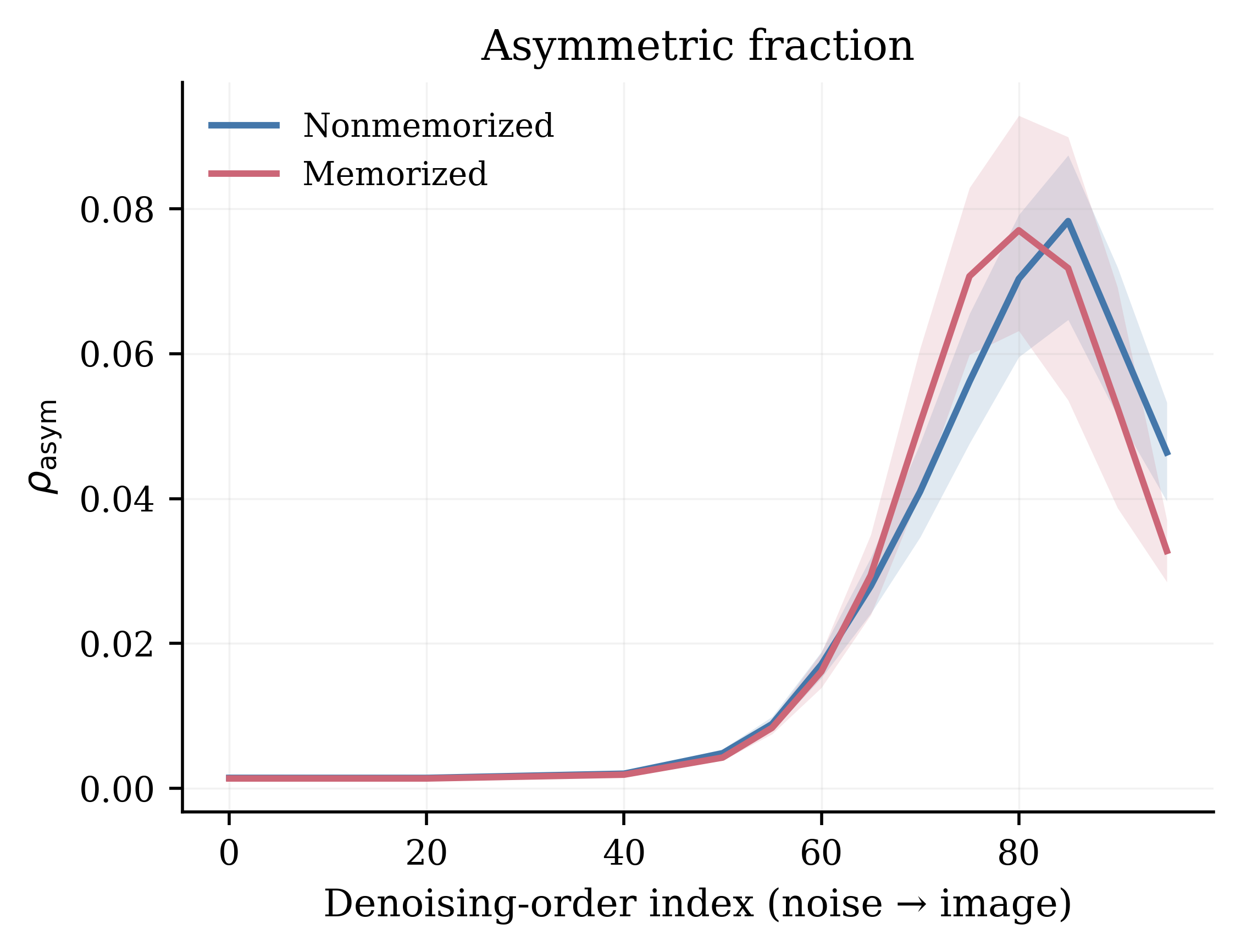}
       
        \par\smallskip
        \textbf{(A) Asymmetric fraction}
    \end{minipage}
    \hfill
    \begin{minipage}[t]{0.32\textwidth}
        \centering
         \includegraphics[width=\linewidth]
         {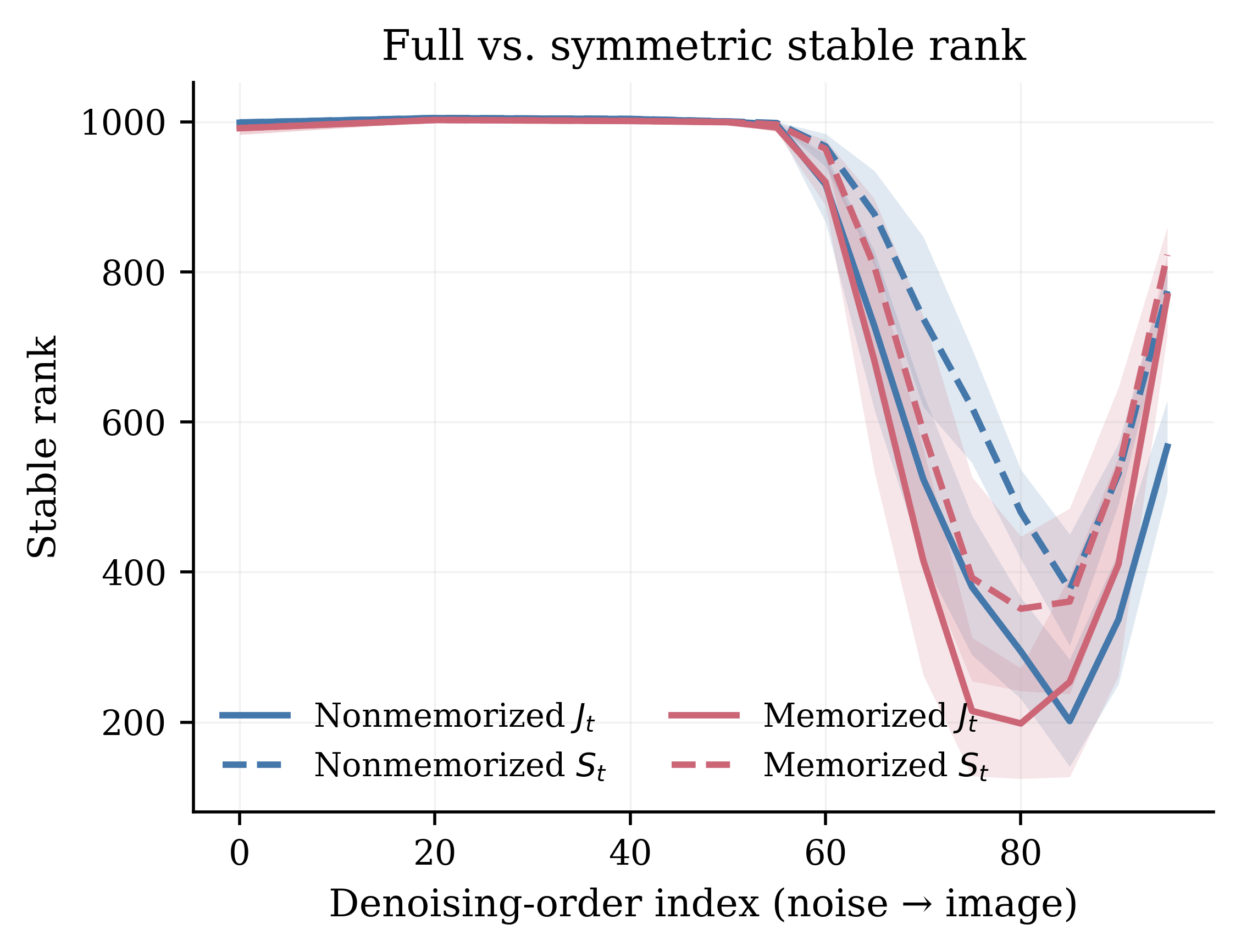}
       
        \par\smallskip
        \textbf{(B) Stable rank}
    \end{minipage}
    \hfill
    \begin{minipage}[t]{0.32\textwidth}
        \centering
         \includegraphics[width=\linewidth]
         {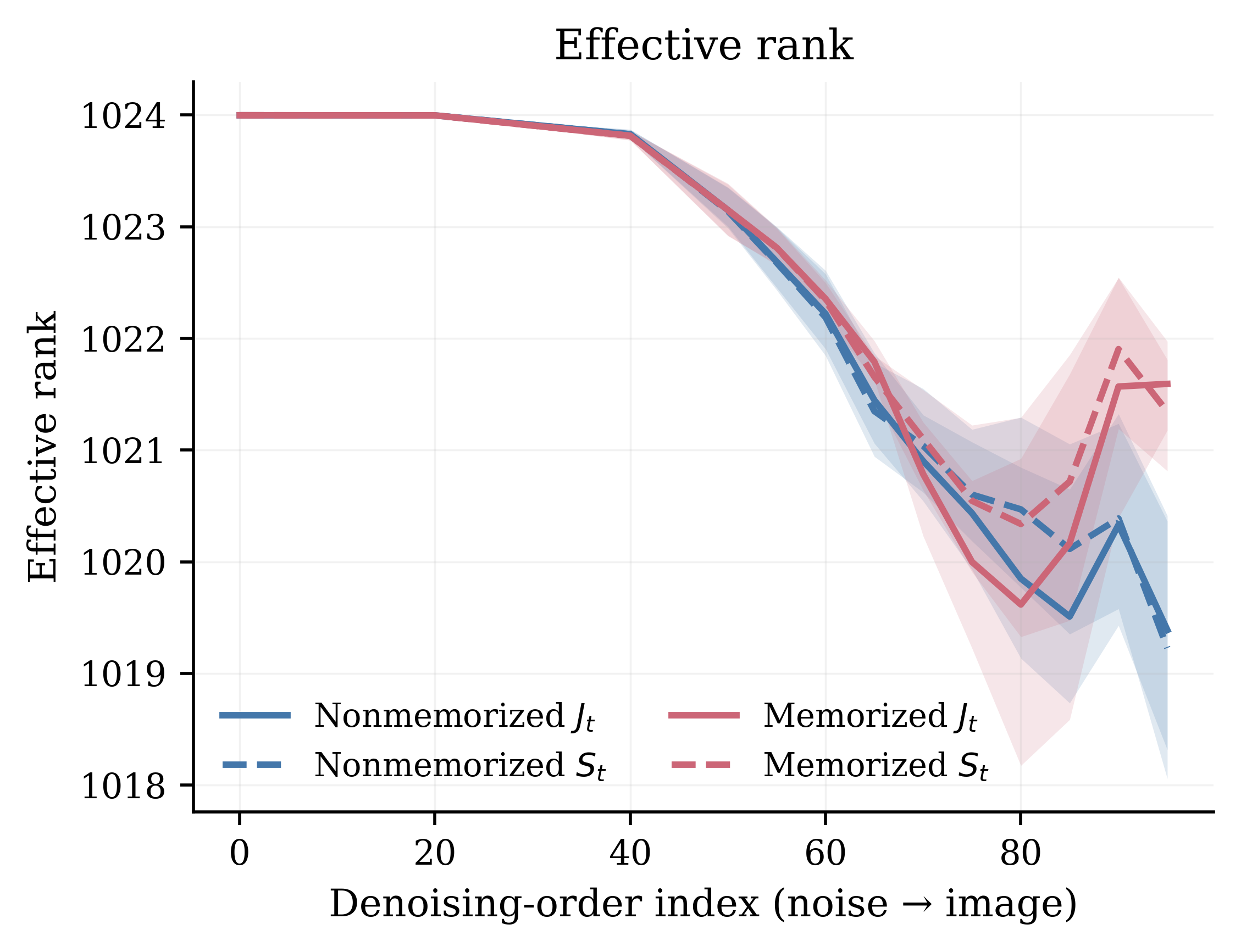}
        
        \par\smallskip
        \textbf{(C) Effective rank}
    \end{minipage}

    \caption{\textbf{Symmetry of the learned Jacobian and robustness of the
    spectral transition.}
    \textbf{(A)} Relative Frobenius norm of the antisymmetric component of the oriented curvature operator
$\widetilde J_t=-J_t=-\partial\epsilon_\theta/\partial x_t$.
    \textbf{(B--C)} Stable rank and effective rank of the full operator
    $J_t$ (solid) and its symmetric component $S_t$ (dashed) along memorized
    and non-memorized trajectories. The learned operator remains close to
    symmetric, while symmetrization preserves the characteristic spectral
    collapse and subsequent recovery.}
    \label{fig:signed_curvature}
\end{figure*}

Figure~\ref{fig:signed_curvature} shows that the learned operator remains
close to symmetric. More
importantly, symmetrization preserves the spectral organization underlying
our analysis. Across evaluated states, the stable and effective ranks of
$S_t$ correlate with their full-Jacobian counterparts at $0.989$ and $0.987$,
respectively, and retain the characteristic spectral collapse and subsequent
recovery. Although asymmetry increases around the collapse, this increase is
similar for memorized and non-memorized trajectories, with no clear
between-group separation.

The eigenvalues of $S_t$ additionally recover the curvature sign discarded
by the singular-value representation. Across the evaluated states, the
spectrum is overwhelmingly dominated by negative curvature, with negligible
spectral mass in positive directions. The observed transition therefore
reflects primarily a reorganization of negative-curvature magnitudes rather
than a change in curvature sign. Together, these results support the
curvature interpretation of the spectral organization used throughout the
main analysis.

\subsection{Spectral and Behavioral Diagnostics}
\label{app:spectral_behavioral_metrics}

We define the complementary spectral and behavioral quantities used in
Sec.~\ref{subsec:geometric_objects}.

Our primary whole-spectrum summary is the stable rank
\[
\operatorname{SR}(J_t)
=
\frac{\lVert J_t\rVert_F^2}{\sigma_1(t)^2}
=
\sum_i
\left(
\frac{\sigma_i(t)}{\sigma_1(t)}
\right)^2.
\]
Low stable rank indicates concentration in a few dominant directions, while
higher values indicate a more broadly distributed spectrum. Its invariance
under $J_t\mapsto cJ_t$ separates this directional organization from overall
curvature scale. We additionally track entropy effective rank and the
relative leading spectral gap $\Delta_{12}(J_t)$, which respectively measure
spectral participation and dominant-direction separation.

\paragraph{Complementary spectral measures.}
For singular values $\{\sigma_i\}$ of $J_t$, let
$p_i=\sigma_i/\sum_j\sigma_j$. The entropy effective rank and relative
leading spectral gap are
\[
\nu(J_t)
=
\exp\!\left(-\sum_i p_i\log p_i\right),
\qquad
\Delta_{12}(J_t)
=
\frac{\sigma_1-\sigma_2}{\sigma_1}.
\]
They respectively measure participation across the full spectrum and
separation of the dominant singular direction.

\paragraph{Perturbation sensitivity.}
At DDIM sampler step $s$, we perturb each trajectory state $x_s^{(i)}$ with
one isotropic Gaussian perturbation calibrated to $10\%$ of its norm,
\[
\widetilde{x}_s^{(i)}
=
x_s^{(i)}
+
0.1\,
\frac{\|x_s^{(i)}\|_2}
     {\|z_s^{(i)}\|_2+10^{-12}}
z_s^{(i)},
\qquad
z_s^{(i)}\sim\mathcal N(0,I).
\]
Let $\Phi_s$ denote deterministic DDIM continuation from step $s$ to the
terminal output. We define
\[
S_s^{(i)}
=
\left\|
\Phi_s(x_s^{(i)})
-
\Phi_s(\widetilde{x}_s^{(i)})
\right\|_2,
\]
which measures how strongly the perturbation changes the eventual generated
output.  We evaluate $s\in{40,50,60,70,80,90}$ using one perturbation per
trajectory and report group means with standard-error bands.

\paragraph{Prediction-improvement rate.}
At sampler step $s$, the predicted clean sample is
\[
\widehat{x}_{0,s}^{(i)}
=
\frac{
x_s^{(i)}
-
\sqrt{1-\bar{\alpha}_{t_s}}\,
\epsilon_\theta(x_s^{(i)},t_s)}
{\sqrt{\bar{\alpha}_{t_s}}},
\qquad
d_s^{(i)}
=
\left\|
\widehat{x}_{0,s}^{(i)}
-
x_{\mathrm{final}}^{(i)}
\right\|_2.
\]
If $s^-$ denotes the preceding cached sampler step, we define
\[
R_s^{(i)}
=
\frac{d_{s^-}^{(i)}-d_s^{(i)}}{s-s^-}.
\]
Predictions are cached at every sampler step in the reported experiment, so
$s^-=s-1$. Positive values indicate that the predicted clean sample has moved
closer to the trajectory's eventual output. We report rates after step $50$, averaged within each memorization group with
standard-error bands.

Neither behavioral diagnostic depends on the Jacobian spectrum or similarity
to a training example, providing independent measurements of the change in
trajectory dynamics around $t^*$.

\begin{figure}[t]
    \centering
    \includegraphics[width=\linewidth]
    {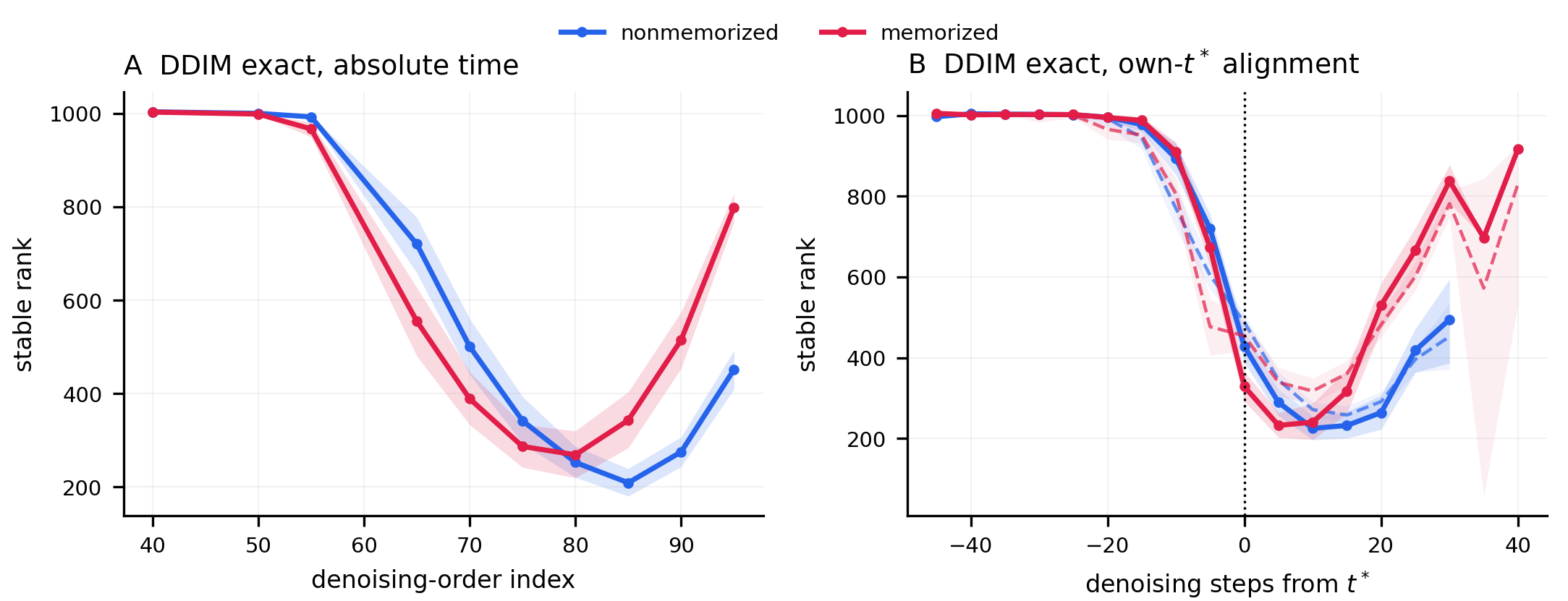}
    \caption{\textbf{Control for trajectory alignment.}
    \textbf{(a)} Stable-rank evolution of memorized and non-memorized DDIM
    trajectories in absolute sampling time, showing distinct late-stage
    evolution without alignment at $t^*$.
    \textbf{(b)} The same trajectories aligned at their individual $t^*$;
    dashed curves show the shuffled-anchor control obtained by randomly
    reassigning $t^*$ values within each group. The transition becomes easier
    to resolve after alignment, while the distinct post-transition recovery is
    not explained by trajectory-specific alignment.}
    \label{fig:alignment_controls}
\end{figure}

\paragraph{Alignment control.}
Because $t^*$ is estimated separately for each trajectory, alignment at this
data-dependent landmark can sharpen features in the averaged spectral curves.
Our analysis does not rely on this sharpness: $t^*$ provides a
trajectory-specific reference for resolving the evolution around the spectral
transition. The distinct late-stage behavior of memorized and non-memorized
trajectories is already visible in absolute sampling time
(Fig.~\ref{fig:alignment_controls}a). We additionally randomly reassign the
observed $t^*$ values across trajectories within each memorization group,
preserving the distribution of alignment times while breaking their
trajectory-specific correspondence (Fig.~\ref{fig:alignment_controls}b).

We additionally assess alignment using 10,000 within-group permutations of
$t^*$, preserving each group's distribution of alignment times while breaking
their trajectory-specific correspondence. Using a window fixed a priori at
$k=10$ sampler steps, we measure the change in prediction-improvement rate
from the $k$ steps preceding $t^*$ to the $k$ steps following it, and use the
difference in this change between memorized and non-memorized trajectories as
the test statistic. The observed contrast is $0.0160$, compared with a
shuffled mean of $-0.0050$ (95\% null interval
$[-0.0208,\,0.0109]$; permutation $p=0.0043$).  The observed contrast  exceeds
the shuffled null, showing that the change in prediction dynamics is localized
to each trajectory's own $t^*$. We therefore use $t^*$ as an observable
spectral landmark associated with trajectory-specific prediction dynamics,
while evaluating the $\delta$-dependent loss of escapeability separately in
Appendix~\ref{app:adversarial_escape}.

\subsection{Posterior-Uncertainty Interpretation of Local Curvature}
\label{app:curvature_uncertainty}

The curvature operator used in the main text also admits a posterior-uncertainty
interpretation. For the Gaussian perturbation path
\[
x_t=\alpha_t x_0+b_t z,
\qquad
z\sim\mathcal{N}(0,I),
\]
the second-order Tweedie identity gives
\[
\operatorname{Cov}[x_0\mid x_t]
=
\frac{b_t^2}{\alpha_t^2}
\left(
I+b_t^2 H_t(x_t)
\right).
\]
Thus, the eigendirections of $H_t$ also describe how uncertainty about the
underlying clean sample is distributed locally across directions. The
identity is exact for the population score; replacing it by the learned score
gives the corresponding model-based approximation
\cite{meng2021estimating}.

This provides a complementary interpretation of the spectral quantities used
in Sec.~\ref{subsec:geometric_objects}: changes in the directional
organization of curvature can equivalently be viewed as changes in the
directional organization of posterior uncertainty along the denoising
trajectory.

\paragraph{Connection to local intrinsic dimension.}
The posterior-covariance identity also clarifies the high-rank recovery
observed for memorized trajectories. Consider a neighborhood in which the
posterior concentrates on a single training example $x^{(i)}$. The
corresponding conditional noise prediction is
\[
\epsilon^*(x_t,t)
=
\frac{x_t-\sqrt{\bar\alpha_t}\,x^{(i)}}{b_t},
\]
and therefore
\[
\frac{\partial\epsilon^*(x_t,t)}{\partial x_t}
=
\frac{1}{b_t}I.
\]
All singular values are equal, so
\[
\operatorname{SR}(J_t)=d.
\]
Equivalently, the posterior covariance of the underlying clean sample
vanishes in every direction. Thus, although the local data distribution is
effectively zero-dimensional, the noise-prediction Jacobian becomes
full-rank and isotropic. The late spectral recovery observed along memorized
trajectories is therefore consistent with the endpoint geometry expected near
an isolated memorized example. This provides the connection between our
trajectory-level spectral analysis and static local-intrinsic-dimension
characterizations of memorization
\citep{ross2025geometric,stanczuk2024diffusion}.

\section{Efficient Memorization Detection}

\label{app:detection}

\subsection{Temporal Recovery and Jacobian Effective Rank}
\label{app:jer_derivation}

The temporal diagnostic in Sec.~\ref{sec:results_georeg} measures the joint recovery of
stable rank and entropy effective rank following spectral collapse. Let $J_n$
denote the Jacobian at monitored index $n$, and define
\[
\operatorname{SR}(J)
=
\frac{\sum_i\sigma_i^2}{\sigma_1^2},
\qquad
\nu(J)
=
\exp\!\left(-\sum_i p_i\log p_i\right),
\qquad
p_i=\frac{\sigma_i}{\sum_j\sigma_j}.
\]
If
\[
m_*(n)
=
\operatorname*{arg\,min}_{m\leq n}\operatorname{SR}(J_m),
\]
the corresponding relative recoveries are
\[
d_{\mathrm{SR}}(n)
=
\left[
\frac{
\operatorname{SR}(J_n)-\operatorname{SR}(J_{m_*(n)})
}{
\operatorname{SR}(J_{m_*(n)})+\varepsilon
}
\right]_+,
\]
and
\[
d_\nu(n)
=
\left[
\frac{
\nu(J_n)-\nu(J_{m_*(n)})
}{
\nu(J_{m_*(n)})+\varepsilon
}
\right]_+.
\]
The temporal recovery score is then
\begin{equation}
R_{\mathrm{sample}}
=
\max_{n\in\mathcal E}
d_{\mathrm{SR}}(n)d_\nu(n).
\label{eq:temporal_recovery_appendix}
\end{equation}
Evaluating  \eqref{eq:temporal_recovery_appendix} requires resolving the
Jacobian spectrum at multiple points along the trajectory. JER instead
targets a local spectral quantity that can be accessed without an explicit
spectral decomposition.

\paragraph{Effective-rank connection.}
For a nonzero matrix $J$, consider the R\'enyi-2 effective rank
\begin{equation}
\operatorname{ER}_2(J)
=
\frac{\lVert J\rVert_*^2}{\lVert J\rVert_F^2}
=
\frac{\left(\sum_i\sigma_i\right)^2}
     {\sum_i\sigma_i^2}
=
\frac{1}{\sum_i p_i^2}.
\label{eq:er2_appendix}
\end{equation}
This quantity is bounded by the two spectral summaries entering
 \eqref{eq:temporal_recovery_appendix}:
\begin{equation}
\operatorname{SR}(J)
\leq
\operatorname{ER}_2(J)
\leq
\nu(J).
\label{eq:er2_bounds_appendix}
\end{equation}
For the first inequality, letting
$M=\sigma_1$, $S=\sum_i\sigma_i$, and $Q=\sum_i\sigma_i^2$ gives
$Q\leq MS$, and hence
\[
\frac{Q}{M^2}
\leq
\frac{S^2}{Q}.
\]
The second follows from the ordering of R\'enyi entropies,
$H_2(p)\leq H_1(p)$, since
$\operatorname{ER}_2(J)=\exp(H_2(p))$ and
$\nu(J)=\exp(H_1(p))$.

\paragraph{Signed-trace surrogate.}
Computing $\operatorname{ER}_2$ directly still requires the nuclear norm.
For a symmetric semidefinite operator, however,
\[
\lVert J\rVert_*
=
\left|\operatorname{tr}(J)\right|.
\]
Motivated by the predominantly sign-coherent curvature observed in
Appendix~\ref{app:signed_curvature}, we replace the nuclear norm by an
oriented trace and define
\begin{equation}
\operatorname{JER}^{(a)}(J)
=
\frac{
\left[a\,\operatorname{tr}(J)\right]_+^2
}{
\lVert J\rVert_F^2+\varepsilon_F
},
\qquad
a\in\{-1,+1\}.
\label{eq:jer_exact_appendix}
\end{equation}
Since
$|\operatorname{tr}(J)|\leq\lVert J\rVert_*$,
the trace-based quantity provides a conservative proxy for
$\operatorname{ER}_2$ when the denominator is matched, becoming exact in the
symmetric sign-coherent limit. The orientation $a$ and stochastic estimates
of the trace and Frobenius energy used in practice are specified in
Appendix~\ref{app:jer_estimation}.

JER should therefore be viewed as a computationally efficient local readout
of the spectral dimensionality implicated by the temporal diagnostic, rather
than an estimator of $R_{\mathrm{sample}}$ itself.
\subsection{Stochastic Estimation and Implementation}
\label{app:jer_estimation}

The trace and Frobenius energy in
 \eqref{eq:jer_exact_appendix} can be estimated from the same stochastic
Jacobian contractions. Let
$z_1,\ldots,z_m\in\{-1,+1\}^d$ be independent Rademacher probes and define
\begin{equation}
q_\ell
=
J^\top z_\ell,
\qquad
\ell=1,\ldots,m.
\label{eq:jer_probe}
\end{equation}
We estimate
\begin{equation}
\widehat{\operatorname{tr}}_m(J)
=
\frac{1}{m}
\sum_{\ell=1}^{m}
z_\ell^\top q_\ell,
\qquad
\widehat F_m(J)
=
\frac{1}{m}
\sum_{\ell=1}^{m}
\lVert q_\ell\rVert_2^2 .
\label{eq:jer_trace_frob}
\end{equation}
Since $\mathbb{E}[z_\ell z_\ell^\top]=I$, these satisfy
\[
\mathbb{E}\!\left[\widehat{\operatorname{tr}}_m(J)\right]
=
\operatorname{tr}(J),
\qquad
\mathbb{E}\!\left[\widehat F_m(J)\right]
=
\lVert J\rVert_F^2.
\]
Crucially, both estimates reuse the same vector--Jacobian products, so no
explicit Jacobian or spectral decomposition is required. 

While $\widehat{\operatorname{tr}}_m(J)$ is unbiased for
$\operatorname{tr}(J)$, squaring its positive part introduces a
variance-dependent bias. We therefore treat stochastic JER as a computational
proxy rather than an unbiased estimator of the deterministic quantity, and
calibrate its score empirically on a disjoint calibration set.

The resulting stochastic JER score is
\begin{equation}
\widehat{\operatorname{JER}}_m(J)
=
\frac{
\left[
a\,\widehat{\operatorname{tr}}_m(J)
\right]_+^2
}{
\widehat F_m(J)+\varepsilon_m
},
\qquad
\varepsilon_m
=
10^{-8}\max\!\left(1,\widehat F_m(J)\right),
\label{eq:jer_stochastic}
\end{equation}
where $a$ fixes the orientation of the signed curvature. The sampler-specific
operators and configurations used in our experiments are
\begin{equation}
(J,a,m)
=
\begin{cases}
\left(
\displaystyle
\frac{\partial\epsilon_\theta}{\partial x},
+1,\,2
\right),
& \text{DDIM},\\[3mm]
\left(
\displaystyle
s\frac{\partial v_\theta}{\partial x}-I,
-1,\,1
\right),
& \text{rectified flow}.
\end{cases}
\label{eq:jer_sampler_specific}
\end{equation}
For rectified flow, the identity correction is the curvature-equivalent
operator derived in Appendix~\ref{app:rf_curvature}. Multiple probes are
evaluated jointly in a single batched autodiff call.

\paragraph{Static query.}
JER is evaluated once after ordinary generation. From the terminal sample, we
construct a deterministic low-noise query state at nominal noise time
$t_0=0.1$ and evaluate   \eqref{eq:jer_stochastic} at that state. No
additional random forward noise is introduced. The resulting score is used
with the threshold rule
\begin{equation}
D(x)
=
\mathbbm{1}
\left\{
\widehat{\operatorname{JER}}_m(J)\geq\tau_{\mathrm{JER}}
\right\},
\label{eq:jer_decision_appendix}
\end{equation}
where $\tau_{\mathrm{JER}}$ is determined exclusively from the calibration
split as described next.

\subsection{Detection Protocol and Baselines}
\label{app:jer_protocol}

\paragraph{Calibration and evaluation.}
We evaluate detection using disjoint calibration and held-out sets for each
sampler. The calibration set contains 1,000 trajectories from ten sampling
seeds, with 100 trajectories per seed; evaluation uses a separate 1,000
trajectories constructed in the same way. Terminal memorization labels are
defined by the nearest-neighbor criterion in
Eq. \ref{eq:memorization_threshold} and are used only for calibration and evaluation.

For each detector, the operating threshold is selected on the calibration set
to maximize F1 subject to
\[
\operatorname{Spec}_{\mathrm{pooled}}\geq0.90,
\qquad
\min_b\operatorname{Spec}_b\geq0.85,
\]
where $b$ indexes sampling seeds. The resulting threshold is fixed before
evaluation on the held-out trajectories. We report AUROC and AUPRC as
threshold-independent measures, together with recall, specificity, and
Matthews correlation coefficient (MCC) at the selected operating point.

\paragraph{Detection baselines.}
We compare JER with the full temporal recovery diagnostic in
\eqref{eq:temporal_recovery_appendix} and three local geometric baselines.
FLIPD~\cite{ross2025geometric} estimates local intrinsic dimensionality from
the score norm and a stochastic estimate of the score-Jacobian trace. The
normal-bundle estimator~\cite{stanczuk2024diffusion} constructs a local
geometric representation from independently perturbed score evaluations and
estimates its dimension through an SVD; in our $d=1024$ controlled setting,
we use the prescribed $4d=4096$ perturbed evaluations. Unless stated
otherwise, all baselines are implemented and evaluated following the
protocols and hyperparameter settings of their respective original works.

We additionally evaluate unconditional score magnitude as a local sharpness
baseline, motivated by \citet{jeon2024understanding} but distinct from their
conditional memorization detector. Since the predictive orientation of this
scalar statistic is not fixed a priori, its orientation and operating
threshold are selected on the calibration set and then frozen for held-out
evaluation.

Detector runtime is measured independently of sampling and includes the model
and derivative evaluations required to obtain each score. Temporal recovery
requires nine full-Jacobian evaluations along the trajectory, whereas JER
uses a single local query with two probe directions for DDIM and one for
rectified flow. Full detection and runtime results are reported in
Table~\ref{tab:jer_detection}.

JER is closely related to FLIPD \citep{ross2025geometric} and the
normal-bundle estimator \citep{stanczuk2024diffusion}: for $J=c\,P_N$
projecting onto a $(d-k)$-dimensional normal space,
$\mathrm{tr}(J)^2/\|J\|_F^2=d-k$, so JER recovers the codimension
$d-\mathrm{LID}$. Unlike FLIPD, which converts a trace estimate into LID
using the noise scale and a score-norm term, JER normalizes by Jacobian
energy and is invariant to $J\mapsto cJ$, separating the organization of
curvature from its magnitude. We evaluate this distinction
empirically in Sec.~\ref{sec:results_georeg}, comparing both detection
performance and computational cost.

\section{\textsc{GeoSteer-DM}: Steering Formulation and Implementation}
\label{app:geosteer}
This appendix provides the general formulation of \textsc{GeoSteer-DM} and
its specialization to the samplers used in our experiments. The central
object is the model-controlled state-space displacement introduced in
Sec.~\ref{sec:geosteer}. GeoSteer identifies a local Jacobian subspace at the
current sampler state and modifies only the component of this displacement
lying in that subspace.

Throughout, $n$ indexes sampler steps and $t_n$ denotes the corresponding
diffusion or flow time, following the notation of the main text. The
formulation below applies to both unconditional sampling and
classifier-free guidance through an appropriate choice of anchor.
\subsection{Sampler and Anchor Formulation}
\label{app:geosteer_sampler}

We write a generic sampler transition as
\begin{equation}
x_{n+1}
=
\mathcal S_{t_n}(x_n,u_n),
\label{eq:geosteer_sampler_app}
\end{equation}
where $u_n$ denotes the native model input used by the sampler update. At the
same current state $x_n$, we evaluate an anchor input $u_n^0$ and the ordinary
model input $u_n^{\mathrm{base}}$, giving
\[
x_{n+1}^{0}
=
\mathcal S_{t_n}(x_n,u_n^0),
\qquad
x_{n+1}^{\mathrm{base}}
=
\mathcal S_{t_n}(x_n,u_n^{\mathrm{base}}).
\]
Their difference
\begin{equation}
\delta_n
=
x_{n+1}^{\mathrm{base}}
-
x_{n+1}^{0}
\label{eq:geosteer_delta_app}
\end{equation}
is the model-controlled displacement on which GeoSteer acts. The anchor is
therefore a local decomposition of a single transition; it does not define a
separate sampling trajectory.

For unconditional DDIM and rectified flow, we use
\[
u_n^0=0,
\]
so that $\delta_n$ isolates the displacement induced by the model prediction.
Under classifier-free guidance,
\[
u_n^{\mathrm{base}}
=
\epsilon_u(x_n,t_n)
+
s_{\mathrm{cfg}}
\left[
\epsilon_c(x_n,t_n)-\epsilon_u(x_n,t_n)
\right],
\qquad
u_n^0
=
\epsilon_u(x_n,t_n),
\]
such that $\delta_n$ isolates the additional displacement induced by
conditional guidance while preserving the unconditional transition.

\begin{algorithm}[t]
\caption{\textsc{GeoSteer-DM}: geometric steering along a sampling trajectory}
\label{alg:geosteer_main}
\begin{algorithmic}[1]
\Require Sampler $\mathcal S$; initial state $x_{n_0}$; active steering steps
$\mathcal A$
\For{each remaining sampler step $n\geq n_0$}
    \State Evaluate the ordinary model output $u_n^{\mathrm{base}}$
    at the current state $x_n$.
    \If{$n\in\mathcal A$}
        \State Form the anchor transition $x_{n+1}^{0}$ and ordinary
        transition $x_{n+1}^{\mathrm{base}}$.
        \State
        $\delta_n
        \gets
        x_{n+1}^{\mathrm{base}}-x_{n+1}^{0}$.
        \State Compute the current Jacobian
        $J_{t_n}=U_{t_n}\Sigma_{t_n}V_{t_n}^{\top}$.
        \State
        $P_n
        \gets
        v_1(t_n)v_1(t_n)^\top$.
        \State
        $x_{n+1}
        \gets
        x_{n+1}^{0}
        +
        (I-2P_n)\delta_n$.
    \Else
        \State
        $x_{n+1}
        \gets
        \mathcal S_{t_n}(x_n,u_n^{\mathrm{base}})$.
    \EndIf
\EndFor
\State \Return the steered terminal sample.
\end{algorithmic}
\end{algorithm}

\subsection{Local Jacobian Geometry}
\label{app:geosteer_geometry}

At each active steering step, GeoSteer constructs a local state-space
subspace from the right singular vectors of the Jacobian associated with the
current sampler state. We write
\begin{equation}
J_{t_n}
=
U_{t_n}\Sigma_{t_n}V_{t_n}^{\top},
\label{eq:geosteer_svd_app}
\end{equation}
and use the leading columns of $V_{t_n}$ as the steering directions. Because
these right singular vectors lie in the sampler-state space, they can act
directly on the displacement $\delta_n$ defined in
  \eqref{eq:geosteer_delta_app}.

For the controlled DDIM experiments,
\begin{equation}
J_{t_n}^{\mathrm{DDIM}}
=
\frac{\partial\epsilon_\theta(x_n,t_n)}
     {\partial x_n}.
\label{eq:jacobian_ddim_app}
\end{equation}
For rectified flow, we use the curvature-equivalent operator derived in
Appendix~\ref{app:rf_curvature},
\begin{equation}
J_{t_n}^{\mathrm{RF}}
=
s_n
\frac{\partial v_\theta(x_n,s_n)}
     {\partial x_n}
-I,
\label{eq:jacobian_rf_app}
\end{equation}
where $s_n$ denotes the physical flow time corresponding to sampler step $n$.
Thus, the controlled DDIM and RF experiments use the same curvature-based
geometric object up to their respective model parameterizations.

At Stable Diffusion scale, we define the local geometry directly from the
classifier-free guidance residual. Let
\[
g(z_n,t_n)
=
\epsilon_c(z_n,t_n)-\epsilon_u(z_n,t_n),
\]
where $\epsilon_c$ and $\epsilon_u$ denote the conditional and unconditional
noise predictions, respectively. We use the guidance-residual Jacobian
\begin{equation}
J_{t_n}^{\mathrm{SD}}
=
J_g(z_n,t_n)
=
\frac{\partial g(z_n,t_n)}{\partial z_n}
=
\frac{\partial
\left[
\epsilon_c(z_n,t_n)-\epsilon_u(z_n,t_n)
\right]}
{\partial z_n}.
\label{eq:jacobian_sd_app}
\end{equation}
Its leading right singular vector identifies the latent direction along which
the conditional--unconditional guidance displacement changes most strongly.
This aligns the geometric operator with the quantity that GeoSteer modifies
under classifier-free guidance.

The operator also admits an approximate conditional-curvature interpretation.
Under the usual noise-prediction/score correspondence,
\[
\epsilon(z_n,t_n)
\approx
-\sigma_{t_n}\nabla_{z_n}\log p_{t_n}(z_n),
\]
the guidance residual satisfies approximately
\[
g(z_n,t_n)
\approx
-\sigma_{t_n}
\left[
\nabla_{z_n}\log p_{t_n}(z_n\mid c)
-
\nabla_{z_n}\log p_{t_n}(z_n)
\right]
=
-\sigma_{t_n}\nabla_{z_n}\log p_{t_n}(c\mid z_n),
\]
and hence
\[
J_g(z_n,t_n)
\approx
-\sigma_{t_n}
\nabla_{z_n}^2\log p_{t_n}(c\mid z_n).
\]
We use this relation only as an interpretation: the learned conditional and
unconditional predictors are approximate, and $J_g$ need not be exactly
symmetric or conservative. GeoSteer therefore operates on the leading right
singular direction of the guidance-residual Jacobian rather than assuming an
exact Hessian structure.

\subsection{Subspace Steering Operator}
\label{app:geosteer_operator}

Let
\[
V_{n,K}
=
[v_1(t_n),\ldots,v_K(t_n)]
\]
contain the leading $K$ right singular vectors of $J_{t_n}$, and define the
corresponding projector
\[
P_n
=
V_{n,K}V_{n,K}^{\top}.
\]
For a sketched Jacobian, $V_{n,K}$ is obtained from the right singular vectors
of the sketch and still lies in the original sampler-state space.

We decompose the model-controlled displacement as
\[
\delta_n
=
P_n\delta_n+(I-P_n)\delta_n.
\]
GeoSteer applies separate gains to the selected geometric subspace and its
orthogonal complement. For subspace gain $\rho_n$ and global displacement
gain $a_n\geq0$, we define
\begin{equation}
\delta_n^{\mathrm{geo}}
=
a_n
\left[
(I-P_n)\delta_n+\rho_nP_n\delta_n
\right]
=
a_n
\left[
I+(\rho_n-1)P_n
\right]\delta_n,
\qquad
x_{n+1}^{\mathrm{geo}}
=
x_{n+1}^{0}+\delta_n^{\mathrm{geo}}.
\label{eq:geosteer_operator_app}
\end{equation}
The canonical GeoSteer operator used in the controlled experiments takes
$K=1$, $a_n=1$, and $\rho_n=-1$, yielding
\begin{equation}
\delta_n^{\mathrm{geo}}
=
(I-2P_n)\delta_n,
\qquad
P_n=v_1(t_n)v_1(t_n)^\top.
\label{eq:geosteer_reflection_app}
\end{equation}
Thus, the component aligned with the dominant local geometric direction is
reversed while its orthogonal complement is preserved.

The more general form in   \eqref{eq:geosteer_operator_app} also contains
suppression ($\rho_n=0$), partial reversal ($-1<\rho_n<0$), and global
attenuation ($a_n<1$) as special cases. These variants are useful for
describing the Stable Diffusion implementation below, but the canonical
controlled-model intervention is the norm-preserving reflection in
  \eqref{eq:geosteer_reflection_app}.

\paragraph{Operator properties.}
Because $P_n$ depends only on the selected subspace, the intervention is
invariant to sign flips and orthogonal changes of basis within that subspace.
Moreover, since
$P_n\delta_n\perp(I-P_n)\delta_n$,
\begin{equation}
\left\lVert\delta_n^{\mathrm{geo}}\right\rVert_2^2
=
a_n^2
\left[
\left\lVert(I-P_n)\delta_n\right\rVert_2^2
+
\rho_n^2
\left\lVert P_n\delta_n\right\rVert_2^2
\right].
\label{eq:geosteer_norm_app}
\end{equation}
For the canonical choice $a_n=1$ and $\rho_n=-1$,
\[
\left\lVert\delta_n^{\mathrm{geo}}\right\rVert_2
=
\left\lVert\delta_n\right\rVert_2.
\]
The intervention therefore changes the direction of the model-controlled
displacement without changing its magnitude.

For this canonical case, the transformation
\[
H_n=I-2P_n
\]
is a Householder reflection satisfying
\[
H_n^\top=H_n,
\qquad
H_n^2=I,
\qquad
H_n^\top H_n=I.
\]
It reverses the selected component exactly while leaving the orthogonal
component unchanged. Reflection provides a simple norm-preserving intervention for testing whether the identified geometric direction is actionable, while more sophisticated control rules could further exploit this structure.

\subsection{Sampler-Specific Forms}
\label{app:geosteer_samplers}

The state-space formulation in   \eqref{eq:geosteer_operator_app} is
independent of how the generative model parameterizes its sampler update. For
the deterministic samplers used in our experiments, the same transformation
can also be written directly in terms of the corresponding model-controlled
quantity.

\paragraph{DDIM.}
Let $s<t$ denote consecutive diffusion times. A deterministic
$\epsilon$-parameterized DDIM transition can be written as
\[
\mathcal S_{t\rightarrow s}(x_t,\epsilon)
=
A_{t\rightarrow s}x_t+B_{t\rightarrow s}\epsilon,
\]
where
\[
A_{t\rightarrow s}
=
\sqrt{\frac{\bar\alpha_s}{\bar\alpha_t}},
\qquad
B_{t\rightarrow s}
=
\sqrt{1-\bar\alpha_s}
-
\sqrt{\frac{\bar\alpha_s}{\bar\alpha_t}}
\sqrt{1-\bar\alpha_t}.
\]
With the zero-prediction anchor,
\[
\delta_t
=
B_{t\rightarrow s}\epsilon_\theta(x_t,t).
\]
Since $B_{t\rightarrow s}$ is scalar, applying
  \eqref{eq:geosteer_operator_app} to $\delta_t$ is equivalent to
transforming the noise prediction as
\begin{equation}
\epsilon_\theta^{\mathrm{geo}}(x_t,t)
=
a_t
\left[I+(\rho_t-1)P_t\right]
\epsilon_\theta(x_t,t).
\label{eq:geosteer_ddim_equiv}
\end{equation}
For the canonical reflection, this reduces to
$\epsilon_\theta^{\mathrm{geo}}=(I-2P_t)\epsilon_\theta$.

\paragraph{Rectified flow.}
For explicit Euler sampling at physical flow time $s_n$,
\[
x_{n+1}
=
x_n+\Delta s_n\,v_\theta(x_n,s_n).
\]
The zero-velocity anchor gives
\[
x_{n+1}^{0}=x_n,
\qquad
\delta_n
=
\Delta s_n\,v_\theta(x_n,s_n).
\]
The corresponding velocity-space form is therefore
\begin{equation}
v_\theta^{\mathrm{geo}}(x_n,s_n)
=
a_n
\left[I+(\rho_n-1)P_n\right]
v_\theta(x_n,s_n).
\label{eq:geosteer_rf_equiv}
\end{equation}
Here, $P_n$ is constructed from the curvature-equivalent RF Jacobian in
  \eqref{eq:jacobian_rf_app}. Thus, the sampler displacement is transformed
in the same way as for DDIM, while the geometry defining the steering
subspace accounts for the linear transport component of the rectified-flow
parameterization.

\paragraph{Classifier-free guidance.}
Under classifier-free guidance, let
\[
g_n
=
\epsilon_c(x_n,t_n)-\epsilon_u(x_n,t_n),
\qquad
u_n^{\mathrm{base}}
=
\epsilon_u(x_n,t_n)+s_{\mathrm{cfg}}g_n.
\]
Using the unconditional prediction as the anchor isolates the
guidance-induced displacement. For deterministic DDIM, the GeoSteer
transformation is therefore equivalent to
\begin{equation}
g_n^{\mathrm{geo}}
=
a_n
\left[I+(\rho_n-1)P_n\right]g_n,
\qquad
u_n^{\mathrm{geo}}
=
\epsilon_u(x_n,t_n)
+
s_{\mathrm{cfg}}g_n^{\mathrm{geo}}.
\label{eq:geosteer_cfg_equiv}
\end{equation}
The unconditional component is preserved, while GeoSteer acts only on the
additional displacement introduced by conditional guidance.

\subsection{Experimental Configuration}
\label{app:geosteer_impl}

\paragraph{Controlled experiments.}
For CelebA32, GeoSteer uses the full $1024\times1024$ Jacobian and exact SVD
at each active steering step. Both DDIM and rectified flow use the rank-one
reflection in   \eqref{eq:geosteer_reflection_app}, with $K=1$,
$a_n=1$, and $\rho_n=-1$. The steering geometry is recomputed at the current
edited state after every modified transition, so subsequent interventions
follow the trajectory produced by GeoSteer rather than the original
unmodified path. We evaluate the intervention at training checkpoints
30k, 60k, 100k, 160k, 200k, and 300k using 500 paired generations per
checkpoint and sampler. The baseline and steered samples in each pair share
the same initial noise. Sampler-specific rewind and active steering windows
are fixed across checkpoints and are reported with the complete evaluation
protocol in Appendix~\ref{app:experiments}.

\paragraph{Stable Diffusion.}
For Stable Diffusion v1.4, GeoSteer uses the guidance-residual Jacobian
$J_g$ defined in Eq.~\ref{eq:jacobian_sd_app}. Because explicitly
constructing
\[
J_g
=
\frac{\partial(\epsilon_c-\epsilon_u)}{\partial z}
\]
in latent space is impractical, we estimate its leading right singular
direction matrix-free using power iteration on $J_g^\top J_g$.

For a unit direction $v$, each iteration first approximates the
Jacobian--vector product using a central finite difference,
\begin{equation}
J_gv
\approx
\frac{
g(z_n+h v,t_n)-g(z_n-h v,t_n)
}{
2h
},
\qquad
h=0.3,
\label{eq:sd_jg_jvp}
\end{equation}
and then evaluates
\[
J_g^\top(J_gv)
\]
with an autograd vector--Jacobian product. Normalizing the result gives the
next power-iteration direction. We use five iterations at the first active
steering step and two warm-started iterations at subsequent steps, initializing
each estimate from the direction obtained at the preceding active step. The
computation is performed in FP32 with TF32 disabled.

The resulting unit direction $v_1(t_n)$ defines
$P_n=v_1(t_n)v_1(t_n)^\top$ and is recomputed at the current, already-steered
latent at every active step. Under classifier-free guidance, GeoSteer acts on
the guidance residual according to Eq.~\ref{eq:geosteer_cfg_equiv}. We use
$K=1$, $\rho_n=-1$, and $a_n=0.7$, giving
\[
g_n^{\mathrm{geo}}
=
0.7\left(I-2P_n\right)g_n,
\qquad
u_n^{\mathrm{geo}}
=
\epsilon_u(z_n,t_n)
+
7.5\,g_n^{\mathrm{geo}}.
\]
Steering is applied over denoising-order indices $0$--$17$ of deterministic
DDIM-50 sampling.

The Stable Diffusion experiment applies GeoSteer over this prescribed
steering window without JER gating or rewind, allowing the steering operator
to be evaluated independently at this scale. Dataset construction, sampling
parameters, and evaluation metrics are given in
Appendix~\ref{app:experiments}.

\section{\textsc{GeoReg-DM}: Objective and Implementation}
\label{app:georeg}

This appendix provides implementation details for the training-time
regularizer introduced in Sec.~\ref{sec:georeg}. GeoReg selectively
regularizes the local Jacobian within each minibatch using an energy-weighted
spectral objective and a low-dimensional differentiable sketch to keep the
additional computation tractable.

\subsection{Training Objective and Selection}
\label{app:georeg_objective}

For each training example, we sample the model-specific training time and
construct the corresponding noisy or interpolated state $x_t$. Let
$\ell_{\mathrm{base}}^{(i)}$ denote the per-sample diffusion or
rectified-flow training loss and
\[
\mathcal L_{\mathrm{base}}
=
\frac{1}{B}\sum_{i=1}^{B}\ell_{\mathrm{base}}^{(i)}
\]
the ordinary minibatch objective.

GeoReg does not evaluate the geometric penalty on every training state.
Instead, we prioritize examples using the input sensitivity of the base loss.
For the mean-reduced diffusion objective, the original per-example
half-squared-error priority is recovered directly from the input gradient
obtained during the ordinary base backward,
\[
\pi^{(i)}
=
\frac{Bd}{2}
\left\|
\nabla_{x_{t_i}^{(i)}}\mathcal L_{\mathrm{base}}
\right\|_2.
\]
The common scale does not affect the resulting ranking. Among
timestep-eligible examples, we select the $m$ highest-priority states for
geometric regularization. This reuses the base-loss backward and avoids an
additional network pass for priority computation.

For each $i\in\mathcal I$, GeoReg penalizes the squared leading singular
value of the sketched Jacobian,
\[
\Phi_{\mathrm{GeoReg}}\!\left(\widehat J^{(i)}\right)
=
\sigma_1\!\left(\widehat J^{(i)}\right)^2,
\]
that is, the squared gain of the dominant direction of the sketched
Jacobian. Since $\widehat J^{(i)}\in\mathbb R^{k\times d}$, this quantity is
obtained directly as the largest eigenvalue of the small Gram matrix
\[
G^{(i)}
=
\widehat J^{(i)}\widehat J^{(i)\top}
\in\mathbb R^{k\times k}.
\]

The geometric loss is averaged over the selected states,
\begin{equation}
\mathcal L_{\mathrm{geo}}
=
\frac{1}{|\mathcal I|}
\sum_{i\in\mathcal I}
\Phi_{\mathrm{GeoReg}}\!\left(\widehat J^{(i)}\right),
\label{eq:georeg_objective_app}
\end{equation}
and the complete training objective is
\begin{equation}
\mathcal L
=
\mathcal L_{\mathrm{base}}
+
\lambda_{\mathrm{geo}}\mathcal L_{\mathrm{geo}}.
\label{eq:georeg_total_app}
\end{equation}

Selection depends only on quantities available at the current training
update. In particular, terminal nearest-neighbor distances and memorization
labels are never used by either the objective or the selection rule.

\begin{algorithm}[t]
\caption{\textsc{GeoReg-DM}: geometry-regularized training step}
\label{alg:georeg}
\begin{algorithmic}[1]
\Require Minibatch of $B$ training examples; fixed
$W\in\mathbb{R}^{k\times d}$; target count $m$; $\lambda_{\mathrm{geo}}$

\State Sample training times $t_i$ and construct corrupted states
$x_{t_i}^{(i)}$ and per-example losses $\ell_{\mathrm{base}}^{(i)}$.

\State
$\displaystyle
\mathcal L_{\mathrm{base}}
\gets
\frac{1}{B}\sum_{i=1}^{B}\ell_{\mathrm{base}}^{(i)}$.

\State Backpropagate $\mathcal L_{\mathrm{base}}$, accumulating parameter
gradients and retaining
$\nabla_{x_{t_i}^{(i)}}\mathcal L_{\mathrm{base}}$.

\State
$\displaystyle
\pi^{(i)}
\gets
\frac{Bd}{2}
\left\lVert
\nabla_{x_{t_i}^{(i)}}\mathcal L_{\mathrm{base}}
\right\rVert_2$.

\State
$\displaystyle
\mathcal I
\gets
\operatorname{TopM}
\bigl(
\{i:\operatorname{Eligible}(t_i)\},
\pi,
m
\bigr)$.

\State Detach the selected realized states
$\{x_{t_i}^{(i)}:i\in\mathcal I\}$ and re-enable input gradients.

\State Construct all $m\times k$ state--projection pairs jointly and compute
the packed Jacobian sketches
\[
\widehat J_{t_i}^{(i)}
=
WJ_{t_i}^{(i)},
\qquad
i\in\mathcal I .
\]

\For{each $i\in\mathcal I$}
    \State
    $\displaystyle
    G^{(i)}
    \gets
    \widehat J_{t_i}^{(i)}
    \widehat J_{t_i}^{(i)\top}$.

    \State
    $\displaystyle
    \Phi^{(i)}
    \gets
    \lambda_{\max}\!\left(G^{(i)}\right)$.
\EndFor

\State
$\displaystyle
\mathcal L_{\mathrm{geo}}
\gets
\frac{1}{|\mathcal I|}
\sum_{i\in\mathcal I}
\Phi^{(i)}$.

\State Backpropagate
$\lambda_{\mathrm{geo}}\mathcal L_{\mathrm{geo}}$ and accumulate its
parameter gradients with those from $\mathcal L_{\mathrm{base}}$.

\State
$\theta
\gets
\operatorname{OptimizerStep}(\theta)$.

\end{algorithmic}
\end{algorithm}

\subsection{Differentiable Jacobian Sketch}
\label{app:georeg_sketch}

Computing and differentiating through the full input Jacobian at every
regularized training state would be prohibitively expensive. GeoReg instead
uses a low-dimensional left sketch. Let
$W\in\mathbb R^{k\times d}$ have orthonormal rows and define
\begin{equation}
\widehat J
=
WJ
\in\mathbb R^{k\times d}.
\label{eq:georeg_sketch_app}
\end{equation}
The geometric objective is evaluated on $\widehat J$ rather than on the full
$d\times d$ operator.

For a uniformly random row-orthonormal sketch,
\begin{equation}
\mathbb E_W
\left[
\lVert WJ\rVert_F^2
\right]
=
\frac{k}{d}\lVert J\rVert_F^2.
\label{eq:georeg_sketch_expectation}
\end{equation}
The sketch therefore provides a low-dimensional view of the local Jacobian
while reducing the spectral computation to the $k\times k$ Gram matrix
$\widehat J\widehat J^\top$. The objective is defined directly on the
sketched operator; we do not interpret its spectral quantities as exact
estimates of those of the full Jacobian.

\paragraph{Differentiable construction.}
The sketch can be obtained without explicitly materializing $J$. For a
selected state $x_t^{(i)}$, we detach the realized state and re-enable
gradients,
\[
\widehat x^{(i)}
=
\operatorname{detach}(x_t^{(i)}),
\]
and let $w_a^\top$ denote row $a$ of $W$. Each row of the sketch is computed
as
\begin{equation}
\widehat J_{a,:}^{(i)}
=
\nabla_{\widehat x^{(i)}}
\left\langle
w_a,
f_\theta(\widehat x^{(i)},t_i)
\right\rangle,
\qquad
a=1,\ldots,k,
\label{eq:georeg_vjp_app}
\end{equation}
where $f_\theta$ denotes the model prediction associated with the training
objective: the noise prediction for diffusion and the velocity field for
rectified flow.

These vector--Jacobian products are evaluated with graph construction
enabled, so differentiating the geometric objective with respect to
$\theta$ introduces the mixed derivatives
\[
\frac{\partial^2 f_\theta}
     {\partial\theta\,\partial x}.
\]
Detaching the realized state prevents this gradient from propagating through
the forward corruption or interpolation process.

For the selected states, the $m\times k$ state--projection pairs are evaluated
jointly in a packed autodiff computation and reshaped into one $k\times d$
Jacobian sketch per state. The geometric objective is then obtained from the
largest eigenvalue of each $k\times k$ Gram matrix
$\widehat J\widehat J^\top$.

\subsection{CelebA Training and Checkpoint Stabilization}
\label{app:georeg_stabilization}

For the CelebA experiments, we train an unconditional noise-predicting
diffusion model on the 4,096-image CelebA32 subset described in
Appendix~\ref{app:celeba_setup}. GeoReg uses a fixed row-orthonormal sketch
$W\in\mathbb{R}^{8\times d}$. Among training states with raw diffusion
timestep $t\in\{90,\ldots,500\}$, we select the $m=32$ states with highest
input-loss-gradient priority and construct their Jacobian sketches using the
packed left-VJP implementation described above. The geometric objective is
the mean over the selected states, with no additional minibatch-size
rescaling. The $8\times8$ Gram matrices are diagonalized exactly, and the
geometry-dependent computation is performed in FP32.

\paragraph{Periodic checkpoint stabilization.}
To stabilize continued optimization under the geometric objective, we maintain fast and slow parameter copies initialized from a warm-up checkpoint.

The fast parameters are optimized using the GeoReg objective. After every
$H=500$ updates, the two copies are synchronized according to
\begin{equation}
\theta_s
\leftarrow
(1-\alpha_s)\theta_s+\alpha_s\theta_f,
\qquad
\theta_f
\leftarrow
\theta_s,
\qquad
\alpha_s=0.5.
\label{eq:georeg_periodic_merge}
\end{equation}

The slow parameters continue to evolve after every
synchronization and are used for evaluation at the reported major
checkpoints. Optimizer moments and counters of the fast optimizer are retained
across synchronization.

This procedure is related to the fast--slow parameter interpolation used by
Lookahead~\citep{zhang2019lookahead}, but here it is used specifically to
stabilize the GeoReg continuation. The initial coefficient $\alpha_0$ and
periodic coefficient $\alpha_s$ have distinct roles: $\alpha_0$ controls the
contribution of the regularized branch to the initial checkpoint, whereas
$\alpha_s$ controls how accumulated fast updates are incorporated into the
evolving slow parameters. The stabilization procedure is separate from the
GeoReg objective itself and uses one actively optimized model together with a
stored slow parameter copy. The stabilization was introduced to prevent rapid degradation in generation quality during continued regularization.

\begin{table}[t]
\centering
\caption{\textbf{CelebA GeoReg training and stabilization settings.}}
\label{tab:georeg_stabilization_settings}
\small
\begin{tabular}{lc}
\toprule
Setting & Value \\
\midrule
Batch size & $512$ \\
Adam learning rate & $10^{-4}$ \\
Geometry timestep window & $t\in\{90,\ldots,500\}$ \\
Selected states per update & $m=32$ \\
Sketch rank & $k=8$ \\
Initial interpolation & $\alpha_0=0.25$ \\
Synchronization interval & $H=500$ updates \\
Periodic interpolation & $\alpha_s=0.5$ \\
Geometry computation & FP32 \\
\bottomrule
\end{tabular}
\end{table}

\subsection{Computational cost}
\label{app:georeg_impl}

Table~\ref{tab:georeg_cost_app} reports the cost of the optimized GeoReg
implementation. Under the same benchmark, an ordinary DDIM training update
takes $272.36$\,ms, compared with $1045.89$\,ms for GeoReg, corresponding to
a $3.84\times$ training-time cost. Peak allocated memory increases from
$16.95$ to $19.63$\,GiB.

Nearly all of the additional computation comes from constructing and
differentiating through the Jacobian sketch. The packed sketch construction
takes $237.09$\,ms and the geometry backward $519.54$\,ms, whereas the exact
$8\times8$ Gram eigensolve and spectral objective require only $1.02$\,ms.
Thus, the spectral computation itself is negligible relative to the
Jacobian-dependent differentiation required by GeoReg.

\begin{table}[t]
\centering
\caption{\textbf{Computational cost of \textsc{GeoReg-DM}.}
CUDA-synchronized median breakdown over ten warmed DDIM training updates
on one NVIDIA A100-SXM4-40GB, using $B=512$, $m=32$ selected states, and
a fixed rank-$8$ left Jacobian sketch.}
\label{tab:georeg_cost_app}
\small
\begin{tabular}{lrr}
\toprule
Component & Time (ms) & Fraction \\
\midrule
Batch setup, base forward + loss
& 88.66 & 8.48\% \\
Base backward
& 176.11 & 16.84\% \\
State selection + geometry-mode entry
& 1.98 & 0.19\% \\
Packed left-Jacobian sketch
& 237.09 & 22.67\% \\
Exact Gram eigensolve + spectral objective
& 1.02 & 0.10\% \\
Geometry backward
& 519.54 & 49.67\% \\
Adam optimizer step
& 16.74 & 1.60\% \\
Scalar/host closure and residual
& 4.75 & 0.45\% \\
\midrule
\textbf{Total}
& \textbf{1045.89} & \textbf{100.00\%} \\
\bottomrule
\end{tabular}
\end{table}
\section{Experimental Details}
\label{app:experiments}

\subsection{GeoReg Training Baselines}
\label{app:georeg_baselines}

We compare GeoReg with several training-time baselines chosen to distinguish
its spectral intervention from simpler mechanisms that can reduce
memorization. Unless stated otherwise, all methods use the same model,
training schedule, minibatches, and evaluation protocol as GeoReg.

\paragraph{Standard training.}
The unregularized baseline continues optimization using only
$\mathcal L_{\mathrm{base}}$.

\paragraph{Tikhonov score regularization.}
Following the score regularization studied by
\citet{baptista2025memorization}, we penalize the squared magnitude of the
estimated score. In the noise-prediction parameterization this gives
\begin{equation}
\Phi_{\mathrm{Tik}}
=
\frac{
\|\epsilon_\theta(x_t,t)\|_2^2
}{
d\,b_t^2
},
\label{eq:tikhonov_baseline}
\end{equation}
where $b_t$ denotes the diffusion noise scale. This baseline regularizes the
magnitude of the learned score without explicitly constraining its local
derivatives.

\paragraph{Reduced learning rate.}
As an optimization control, we halve the Adam learning rate after 
\texttt{10k} training checkpoint. All methods share the same training history
with learning rate $10^{-4}$ through \texttt{10k}; the reduced-LR baseline
then continues ordinary diffusion training using only the base
noise-prediction objective at $5\times10^{-5}$ through \texttt{60k}.
This tests whether reduced memorization can be explained by slower continued
optimization alone.

\paragraph{Ambient-style noisy training.}
We additionally consider a fixed-noisy-bank baseline inspired by Ambient
Diffusion \cite{daras2023ambient}. At the training fork, each training example is corrupted once at
diffusion index $50$, producing a fixed noisy training set used throughout
the continuation. For timesteps up to this corruption level, training uses
the ordinary noise-prediction objective; at larger timesteps, we use the
corresponding noise-aware objective without restoring clean supervision.
This changes the information supplied during training rather than directly
regularizing the learned score or its Jacobian. Because the construction is
adapted to our uniform-time noise-prediction setting, we refer to it as an
Ambient-style baseline rather than a reproduction of the original method.

\paragraph{Regularization strengths.}
For the synthetic experiments, regularization coefficients are calibrated at
the common training fork so that the initial auxiliary-loss parameter-gradient
norm is approximately $10\%$ of the base-loss gradient norm. The reduced-learning-rate and Ambient-style controls do not
use this calibration because they modify the optimization or training-data
procedure rather than adding a regularization term.

\subsection{Synthetic Data Experiments}
\label{app:synthetic_setup}

We evaluate GeoReg on three synthetic populations of increasing geometric
complexity. In each setting, the training set is sampled once and then held
fixed throughout optimization; only diffusion noise and timesteps are
resampled during training. Evaluation uses independent samples from the known
population distribution. The model is trained only on the fixed training
coordinates and is not provided with component identities, latent
coordinates, population parameters, or other information used to construct
the datasets.

\paragraph{Gaussian80.}
The training set consists of $N=80$ independent samples from a
two-dimensional standard Gaussian,
\[
x\sim\mathcal N(0,I_2).
\]
No standardization is applied. This provides a full-dimensional unimodal
setting in which attraction to individual training coordinates can be
separated from multimodal or manifold structure.

\paragraph{Mixture32.}
We construct an eight-component Gaussian mixture in two dimensions with
uniformly weighted component means
\[
\mu_j
=
1.8
\left(
\cos\frac{2\pi j}{8},
\sin\frac{2\pi j}{8}
\right),
\qquad
j=0,\ldots,7,
\]
and covariance $0.2^2I_2$ within each component. The fixed training set
contains four independently sampled points from each component, giving
$N=32$. Population samples instead draw the component identity uniformly
before sampling from the corresponding Gaussian. The stratified training set
allows us to examine copying together with mode coverage, component balance,
and within-component variation.

\paragraph{SwissRoll8D128.}
The final synthetic setting is a noisy two-dimensional curved surface
embedded in $\mathbb R^8$. We sample
\[
u\sim\operatorname{Uniform}(1.5\pi,4.5\pi),
\qquad
v\sim\operatorname{Uniform}(-10,10),
\]
and define
\[
y
=
\left(
\frac{u\cos u}{10},
\frac{v}{10},
\frac{u\sin u}{10},
0,\ldots,0
\right).
\]
Samples are then generated as
\[
x=Qy+0.03z,
\qquad
z\sim\mathcal N(0,I_8),
\]
where $Q$ is a fixed orthogonal $8\times8$ matrix generated with seed $601$.
The training set contains $N=128$ samples. We do not whiten the thin
directions, allowing the evaluation to distinguish variation along the
surface from variation away from it.

\paragraph{Training and sampling.}
All synthetic experiments use a width-$128$ noise-prediction MLP trained
with Adam at learning rate $10^{-3}$ and batch size $512$ for
\texttt{30k} updates, with methods sharing the same initial
\texttt{5k}-update unregularized prefix. The low dimensionality permits exact
input-Jacobian computation. Geometric regularization is applied to at most
$32$ eligible states per minibatch at diffusion timesteps
$t\in\{5,\ldots,100\}$, using the selection procedure specified for the
corresponding experiment. Generation uses deterministic DDIM with $100$
sampling steps.

For each dataset and data seed, we fix $2{,}000$ validation population
samples, $10{,}000$ final population samples, and a separate $10{,}000$-sample
population-null set. The latter provides a population baseline for
nearest-training-point statistics: population samples can naturally lie near
the finite training set.

\paragraph{Evaluation.}
We distinguish copying from population-level distributional agreement.
Copying is measured relative to the fixed finite training set. Sliced
Wasserstein distance (SWD), distributional precision, and distributional
recall are instead computed against independent samples from the underlying
population. SWD measures agreement between the generated and population
distributions through one-dimensional projections, while precision measures
the fraction of generated samples lying within the estimated support of the
population and recall measures the fraction of the population support covered
by the generated distribution.

These quantities are complementary. A model can produce individually
plausible samples while covering only part of the population, resulting in
high precision but lower recall. Conversely, increased coverage can be
accompanied by additional mass outside the population support. SWD is
sensitive to broader distributional discrepancies, including changes in
relative mass across mixture components. None of these population metrics
measures copying directly, which is why we report them together with
nearest-training-point statistics.

\subsection{Controlled CelebA Setting}
\label{app:celeba_setup}

We use a controlled CelebA setting in which full input Jacobians can be
computed during sampling. The training set contains 4,096 images resized and
center-cropped to $32\times32$, converted to grayscale, and scaled to
$[0,1]$. We subtract the mean image intensity computed over this subset and
apply no variance normalization.

Both generative models use the same time-conditioned U-Net with base width
$64$, channel multipliers $(1,2,4)$, and two residual blocks per resolution.
Attention is applied at $16\times16$, $8\times8$, and the bottleneck. The
network uses four-head attention, GroupNorm, SiLU activations, dropout $0.1$,
and sinusoidal time embeddings.

\paragraph{Diffusion.}
The diffusion model is trained with the standard noise-prediction objective
\[
\mathcal L_{\mathrm{DM}}
=
\mathbb E_{x_0,\epsilon,t}
\left[
\left\|
\epsilon-
\epsilon_\theta
\left(
\sqrt{\bar\alpha_t}x_0+
\sqrt{1-\bar\alpha_t}\epsilon,t
\right)
\right\|_2^2
\right],
\]
using $T=1000$ noise levels and a linear schedule
$\beta_t\in[10^{-4},2\times10^{-2}]$. Training timesteps are sampled
uniformly.  Generation uses deterministic DDIM ($\eta=0$) with 100
linearly spaced schedule points from timestep $999$ to $0$, corresponding to
99 transitions.

\paragraph{Rectified flow.}
The rectified-flow model is trained on the linear interpolation
\[
x_t
=
(1-t)x_{\mathrm{noise}}+t x_{\mathrm{data}},
\qquad
v^\star
=
x_{\mathrm{data}}-x_{\mathrm{noise}},
\]
with uniformly sampled discretized times over $[0,1]$ and velocity-matching
objective
\[
\mathcal L_{\mathrm{RF}}
=
\mathbb E
\left[
\left\|
v_\theta(x_t,t)-v^\star
\right\|_2^2
\right].
\]
 Sampling uses explicit Euler integration on 100 uniformly spaced
points from $t=0$ to $1$, again giving 99 numerical updates.

The remaining regularizer-specific settings are given in
Appendix~\ref{app:georeg_stabilization}. Although training runs permit a longer
optimization horizon, the results in the main text use checkpoints through
300k updates.

\subsection{Memorization and Generation-Quality Metrics}
\label{app:evaluation_metrics}

For the controlled models, memorization is measured directly in the
mean-centered grayscale pixel space used for training. Given a generated
sample $y$, let $d_1(y)$ and $d_2(y)$ denote its smallest and second-smallest
Euclidean distances to the 4,096 training examples. We define the
nearest-neighbor gap ratio
\begin{equation}
r(y)
=
\frac{d_1(y)}
     {d_2(y)+10^{-10}},
\label{eq:nn_gap_ratio}
\end{equation}
and classify a generation as memorized when
\begin{equation}
r(y)<\frac{1}{3}.
\label{eq:memorization_threshold}
\end{equation}
We report memorization rate together with the corresponding nearest-neighbor
gap statistics.

Generation quality is evaluated using FID with 2,048-dimensional Inception-v3 pool-3 features. For GeoReg, FID is computed from 50,000 generations against the full 4,096-image CelebA training set; for GeoSteer, it is computed from 500 generations per arm and checkpoint against a fixed 1,024-image CelebA reference set. We treat FID as a secondary metric in this controlled setting because direct reproduction of images in the finite reference set can itself improve FID.

\subsection{Controlled-Model Evaluation}
\label{app:controlled_evaluation}

\paragraph{GeoReg.}
We evaluate all training methods using the same ordinary deterministic DDIM sampling procedure. We report results through \texttt{60k} updates, by which point the training objective has largely saturated, and do not continue these runs into the deliberately overtrained regime used separately to study memorization dynamics and GeoSteer. Training configurations and computational-cost measurements are given in Appendix~\ref{app:georeg_impl}.

\paragraph{Memorization detection.}
JER is evaluated at the 160k checkpoint for both generative dynamics in the main body. We report additional checkpoints in the appendix. 
Calibration uses 1,000 trajectories per sampler, obtained from ten sampling
seeds with 100 generations each. A disjoint set of another 1,000 trajectories
per sampler is reserved for evaluation. Threshold selection and baseline
definitions are given in Appendix~\ref{app:jer_protocol}. Throughout our analyses, memorization rates are estimated from the generation cohorts used by each
experiment; the differing evaluation sizes therefore produce small variations
in the reported control rates across sections.

\paragraph{GeoSteer.}
We evaluate steering at
$30\mathrm{k},\,
60\mathrm{k},\,
100\mathrm{k},\,
160\mathrm{k},\,
200\mathrm{k},\,
300\mathrm{k}$ updates. Each sampler--checkpoint pair contains 500 paired baseline and
steered generations, obtained from five sampling seeds with 100 generations
each. Within each pair, both trajectories begin from the same Gaussian
initial state.

For DDIM, steering resumes from sampler state 60 and is applied to transitions
60--98. For rectified flow, it resumes from state 20 and is applied to Euler
updates 20--89. The geometric operator and sampler-specific formulation are
given in Appendix~\ref{app:geosteer}.

\subsection{Stable Diffusion Evaluation}
\label{app:sd_evaluation}

We evaluate text-to-image mitigation using Stable Diffusion v1.4
at $512\times512$ resolution.
Generation uses deterministic DDIM with 50 steps, $\eta=0$, and
classifier-free guidance scale $7.5$.

\paragraph{Evaluation cohorts.}
The memorized cohort contains 500 prompts with known memorized reference
images from the Wen et al.\ benchmark. We generate four paired deterministic
samples per prompt, giving 2,000 generations per method. To evaluate behavior
away from known memorization, we additionally use 100 prompts labeled as
non-memorized in the corresponding ground-truth collection and generate one
paired sample per prompt and method.

\paragraph{Metrics.}
Training-image similarity is measured using SSCD cosine similarity. For the
memorized cohort, each generation is compared against the complete bank of
500 memorized references and assigned its maximum similarity; we additionally
report the fraction satisfying
\[
\operatorname{SSCD}\geq0.50.
\]
For the non-memorized control cohort, nearest-reference similarity is computed
against the common Webster/LAION reference bank.

Prompt fidelity is measured as the cosine similarity between normalized CLIP
ViT-B/32 prompt and image embeddings. We additionally report FID using
Inception-v3 pool-3 features against a fixed reference set of 10,000
LAION-derived images center-cropped and resized to $512\times512$.

\paragraph{Baselines.}
We compare GeoSteer with the unmodified sampler and four inference-time
mitigation methods: Ren et al.~\citep{ren2024unveiling}, RTA
\citep{somepalli2023understanding}, Wen et al.~\citep{wen2024detecting},
and Han et al.~\citep{han2026adjusting}. We additionally compose each method
with GeoSteer to evaluate whether geometric steering provides complementary
mitigation. All methods use the configurations reported for the corresponding
experiments.  To isolate the effect of the interventions, all methods are applied to every
trajectory independently of detection, and each baseline--GeoSteer pair
shares the same initial noise.

\section{Additional Results and Ablations}

\subsection{Adversarial Evidence for Memorization Basin Entry}
\label{app:adversarial_escape}

We test the basin interpretation directly by optimizing bounded perturbations
to the remaining DDIM trajectory around the geometry-defined transition
$t^*$. Starting from offset $\tau=t-t^*$, we introduce a control $u_k$
before each remaining sampler transition with
\[
\|u_k\|_2\leq\delta,
\qquad
\delta_{\mathrm{RMS}}=\frac{\delta}{\sqrt d}.
\]
We evaluate
$\delta_{\mathrm{RMS}}\in\{0.001,0.003,0.005,0.007\}$ and
$\tau\in\{-10,-5,0,5,10\}$ on 5000 held-out memorized trajectories. An
intervention escapes when the terminal sample satisfies
$r(x_0^{\mathrm{adv}})\geq1/3$.

We compare two strategies for finding an escaping continuation: end-to-end
projected gradient ascent over the remaining controls and a greedy
state-by-state optimization of the same nearest-neighbor gap objective.
Matched random perturbations provide a magnitude control.

\begin{figure}[t]
    \centering
    \includegraphics[width=\linewidth]
    {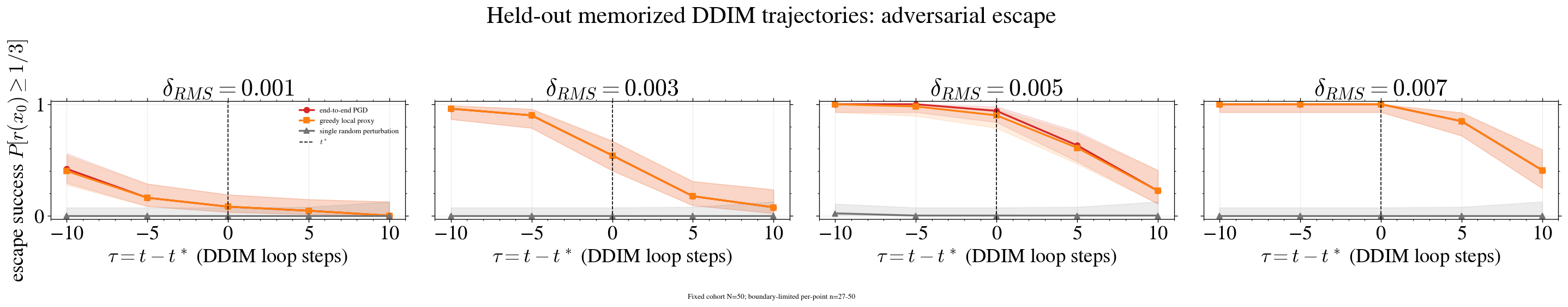}
    \caption{\textbf{Adversarial evidence for memorization basin entry.}
    Escape probability under bounded interventions relative to $t^*$.
    Escapeability decreases sharply through the transition, with larger
    perturbation bounds remaining effective later in denoising. End-to-end
    and greedy optimization yield similar profiles, while matched random
    perturbations almost never escape. Shaded regions show 95\% Wilson confidence intervals.}
    \label{fig:adversarial_escape}
\end{figure}

Figure~\ref{fig:adversarial_escape} shows a sharp loss of escapeability around
$t^*$. At $\delta_{\mathrm{RMS}}=0.003$, optimized escape probability falls
from $0.96$ at $\tau=-10$ to $0.54$ at $t^*$ and $0.07$ at $\tau=10$.
Increasing the perturbation bound shifts this transition later in sampling,
as expected from the dependence of $B_t(\delta)$ on the admissible
continuation set. End-to-end and greedy optimization agree on more than
$97\%$ of trajectory--offset pairs, whereas matched random perturbations
produce almost no escapes. As an additional control, we observe that  successful
escapes largely satisfy an independent endpoint-quality
criterion.

These results show that bounded escape becomes rapidly more difficult
around the spectral transition, with stronger interventions remaining
effective for longer. Since a finite adversarial search cannot certify
basin membership, we treat $t^*$ as an empirical marker of declining
escapeability .
\subsection{Singular-Value Dynamics of the Spectral Transition}
\label{app:singular_values}

Section~\ref{subsec:geometric_objects} summarizes the spectral transition using the relative spectral gap $\Delta_{12}$. Here, we further investigate the underlying singular-value dynamics and demonstrate how this transition manifests in stable rank. 

\begin{figure}[t]
    \centering

    \begin{subfigure}[t]{0.49\linewidth}
        \centering
        \includegraphics[width=\linewidth]
        {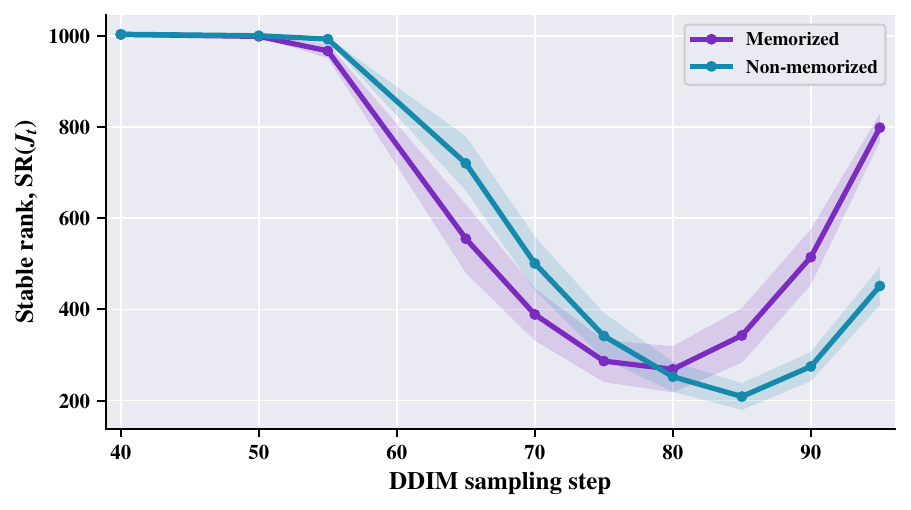}
        \caption{Stable rank}
    \end{subfigure}
    \hfill
    \begin{subfigure}[t]{0.49\linewidth}
        \centering
        \includegraphics[width=\linewidth]
        {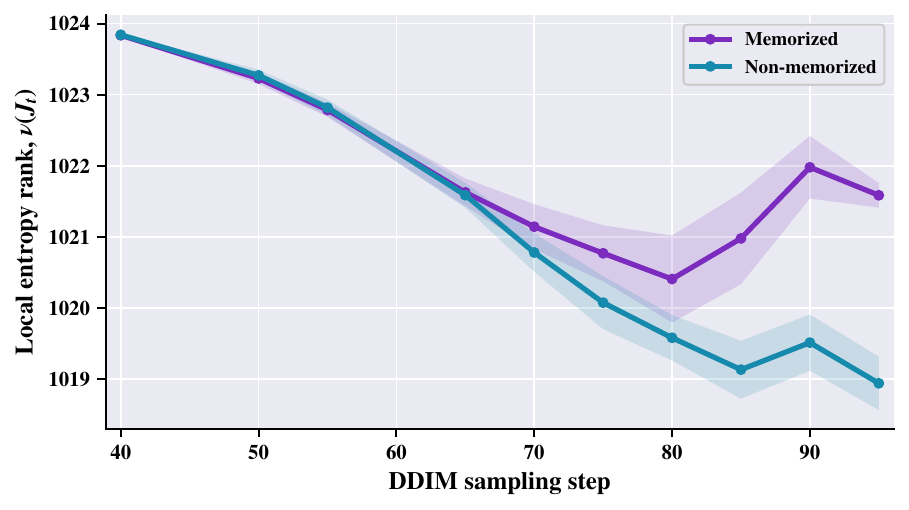}
        \caption{Entropy effective rank}
    \end{subfigure}

    \vspace{0.5em}

    \begin{subfigure}[t]{0.49\linewidth}
        \centering
        \includegraphics[width=\linewidth]
        {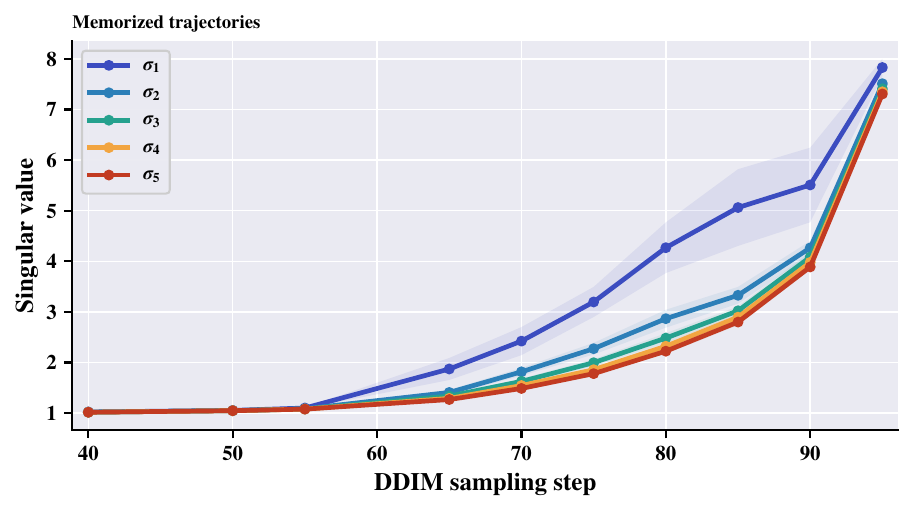}
        \caption{Leading singular values, memorized}
    \end{subfigure}
    \hfill
    \begin{subfigure}[t]{0.49\linewidth}
        \centering
        \includegraphics[width=\linewidth]
        {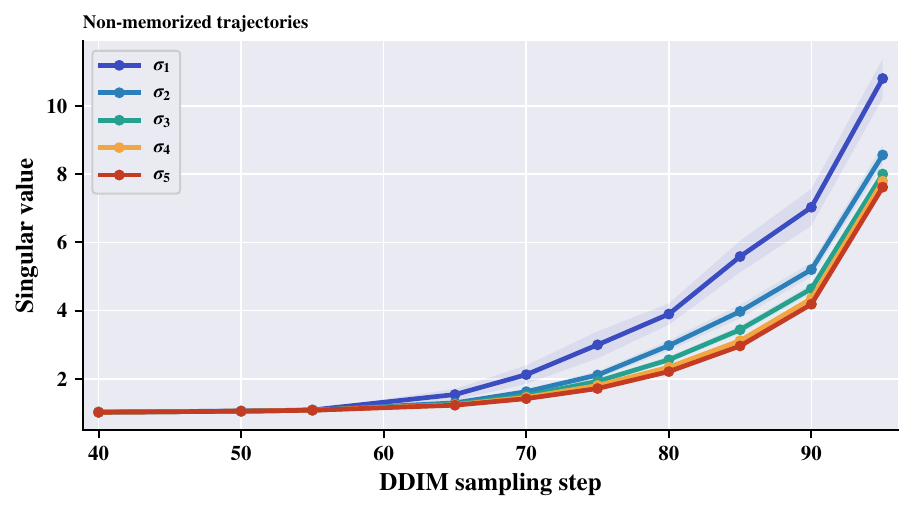}
        \caption{Leading singular values, non-memorized}
    \end{subfigure}

    \caption{\textbf{Singular-value dynamics of the spectral transition.}
    Exact full-SVD Jacobians along DDIM trajectories at the 160k checkpoint
    in absolute sampling time.
    \textbf{(a--b)} Both cohorts exhibit collapse--recovery dynamics in stable
    and entropy effective rank, with substantially stronger low-noise recovery
    for memorized trajectories.
    \textbf{(c--d)} Spectral collapse is driven by $\sigma_1$ separating from
    lower singular modes. At low noise, secondary singular values approach
    $\sigma_1$ for memorized trajectories, whereas $\sigma_1$ remains more
    strongly separated for non-memorized trajectories.}
    \label{fig:app_singular_values}
\end{figure}

\paragraph{The spectral collapse is driven by the leading singular value.}
At high noise, the spectrum remains nearly isotropic in both cohorts, with singular values near unity and stable rank approaching the ambient dimension. As denoising enters intermediate noise levels, all singular values increase, but $\sigma_1$ grows faster than the bulk spectrum (Figure~\ref{fig:app_singular_values}c--d). Because $\mathrm{SR}(J_t) = \sum_i \sigma_i^2 / \sigma_1^2$, this relative overshoot of $\sigma_1$ induces the collapse in stable rank (Figure~\ref{fig:app_singular_values}a). Spectral recovery occurs when secondary singular values subsequently catch up to $\sigma_1$.

\paragraph{Memorized trajectories exhibit earlier separation and stronger recovery.}
Along memorized trajectories, $\sigma_1$ decouples from $\sigma_2$ earlier in the sampling process and by a wider margin, causing stable rank to fall sooner. The two cohorts diverge most prominently at low noise: for memorized samples, the leading singular values reconverge toward a nearly common scale, consistent with the isotropic $1/b_t$ scaling expected near an isolated memorized state. Consequently, their stable rank recovers substantially. For non-memorized trajectories, $\sigma_1$ remains separated from the remaining spectrum, leaving the recovery incomplete. In absolute time, minimum stable rank is comparable across groups; the key distinguishing features are the earlier separation of $\sigma_1$ and its subsequent recovery, which $\Delta_{12}$ captures more directly upon alignment at $t^*$ (Figure~\ref{fig:alignment}a).

\paragraph{Local entropy rank.}
Local entropy rank $\nu(J_t)$ displays the same qualitative collapse--recovery profile across a narrower dynamic range (Figure~\ref{fig:app_singular_values}b). Because $\nu$ integrates over the full singular spectrum, it is dominated by bulk modes rather than a single leading singular value. Nevertheless, memorized trajectories systematically recover toward a higher-rank spectrum at low noise, whereas non-memorized trajectories continue to decay.

\begin{figure}[t]
  \centering
  \begin{subfigure}[t]{0.48\linewidth}
    \centering
    \includegraphics[width=\linewidth]{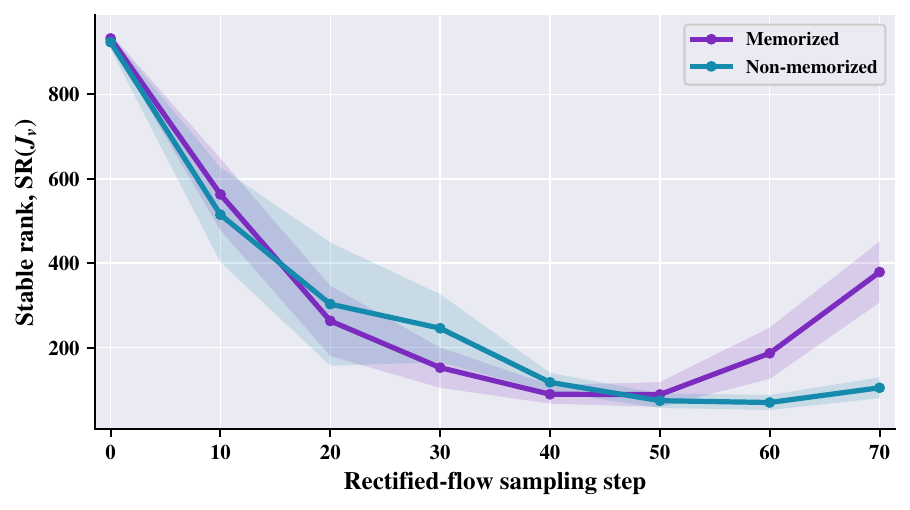}
    \caption{Stable rank}
  \end{subfigure}\hfill
  \begin{subfigure}[t]{0.48\linewidth}
    \centering
    \includegraphics[width=\linewidth]{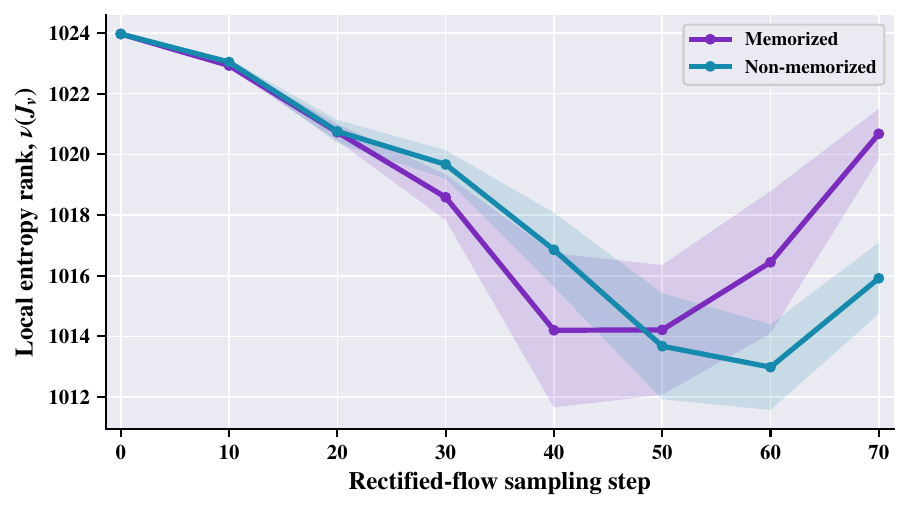}
    \caption{Entropy effective rank}
  \end{subfigure}
  \vspace{0.5em}
  \begin{subfigure}[t]{0.48\linewidth}
    \centering
    \includegraphics[width=\linewidth]{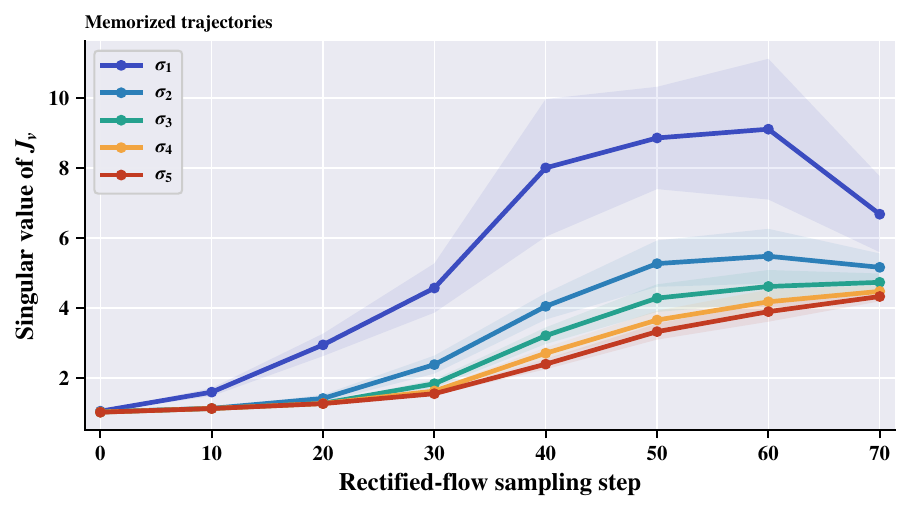}
    \caption{Leading singular values, memorized}
  \end{subfigure}\hfill
  \begin{subfigure}[t]{0.48\linewidth}
    \centering
    \includegraphics[width=\linewidth]{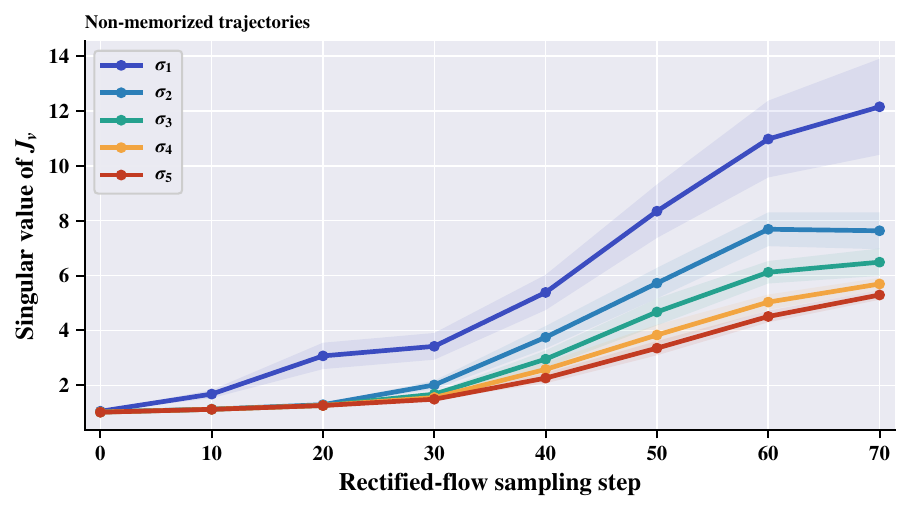}
    \caption{Leading singular values, non-memorized}
  \end{subfigure}
  \caption{\textbf{Singular-value dynamics for rectified flow.} Exact full-SVD Jacobians along rectified-flow trajectories at the 160k checkpoint in absolute sampling time. (a--b) Both cohorts exhibit spectral collapse, with substantially stronger recovery for memorized trajectories. (c--d) The collapse is driven by $\sigma_1$ separating from lower modes; for memorized trajectories, $\sigma_1$ subsequently declines toward the secondary singular values, whereas for non-memorized trajectories it continues to grow.}
  \label{fig:rf_singular_dynamics}
\end{figure}

\paragraph{Rectified flow.} Rectified-flow trajectories show a qualitatively similar picture (Figure~\ref{fig:rf_singular_dynamics}). Along memorized trajectories, $\sigma_1$ separates from the secondary singular values earlier and by a wider margin, then declines back toward them, producing a pronounced recovery in stable rank. Along non-memorized trajectories, $\sigma_1$ continues to grow and remains separated, so stable rank stays low.

\subsection{Generation Quality and Perceptual Effects of GeoReg-DM}
\label{app:georeg_quality}

\begin{figure}[t]
    \centering
    \fbox{%
        \begin{minipage}{0.6\linewidth}
            \centering
            \includegraphics[
                width=\linewidth,
                height=0.34\textheight,
                keepaspectratio
            ]{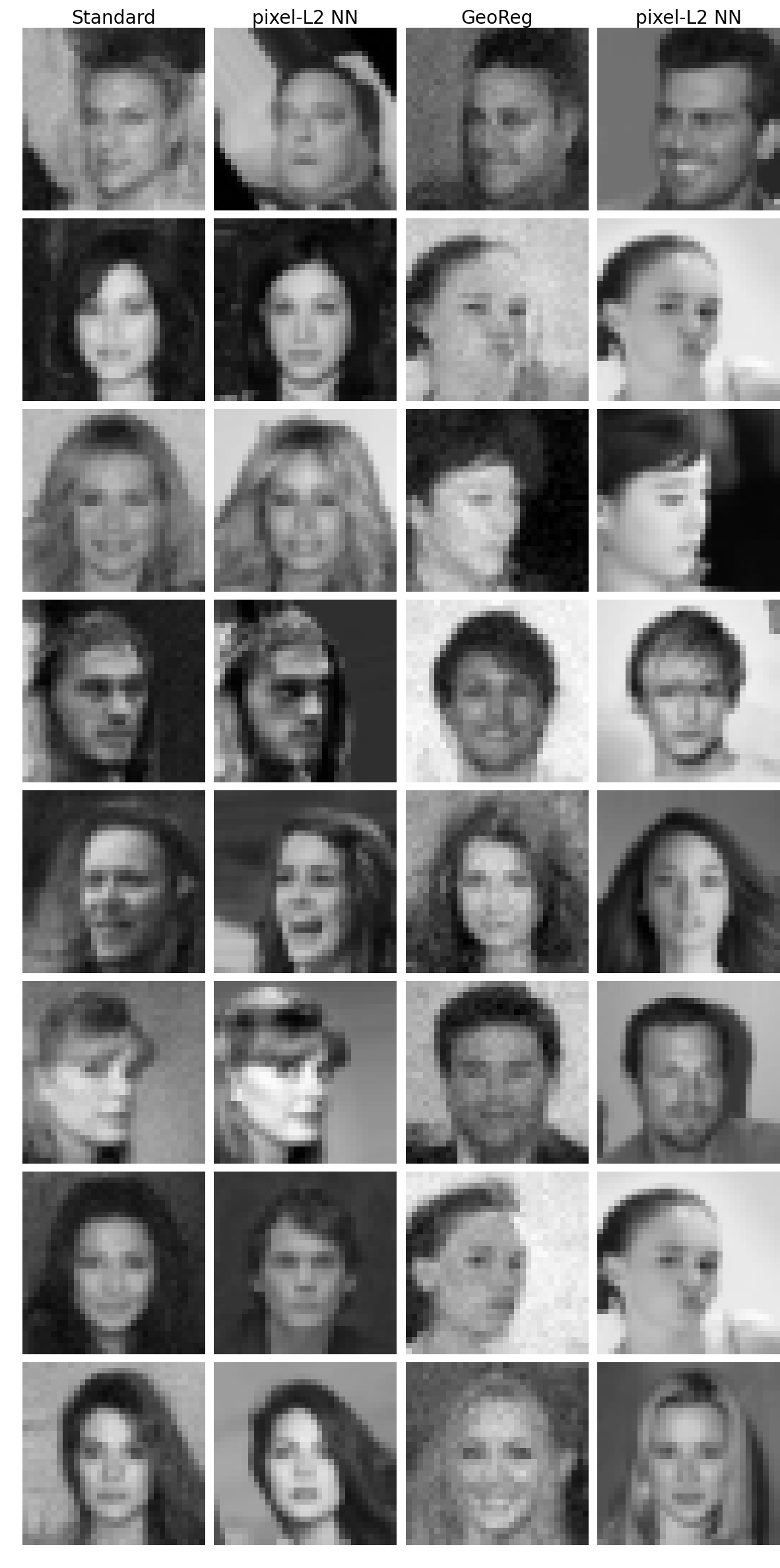}
        \end{minipage}%
    }
    \caption{\textbf{Generation quality under GeoReg.}
    Representative samples from standard training and the final GeoReg
    configuration, together with their pixel-$L_2$ nearest training examples.
    GeoReg substantially reduces copying while retaining coherent CelebA
    generations.}
    \label{fig:georeg_quality}
\end{figure}

We further examine how the reduction in copying affects the generated
distribution. On CelebA32, we compare standard training with the final GeoReg
configuration using the evaluation protocol of
Sec.~\ref{sec:results_georeg}. Figure~\ref{fig:georeg_quality} shows
representative generations from the two distributions. Despite the substantial
reduction in training-set reproduction, GeoReg continues to produce coherent
CelebA samples.

The change is not confined to samples close to the copying threshold. As shown
in Table~\ref{tab:georeg_perceptual}, the mean nearest-neighbor gap ratio
increases from $0.766$ to $0.846$, while its $5$th percentile increases from
$0.273$ to $0.470$. Nearest-training distances also increase in both pixel
and LPIPS space. GeoReg therefore shifts the distribution more broadly away
from individual training examples rather than merely moving borderline
generations across the $r=1/3$ threshold.

\begin{table}[t]
\centering
\caption{\textbf{Training-set proximity under GeoReg.}
}
\label{tab:georeg_perceptual}
\small
\begin{tabular}{lcc}
\toprule
Measure & Standard & GeoReg \\
\midrule
Copy rate ($r<1/3$) & 8.80\% & 1.56\% \\
Mean $r$             & 0.766 & 0.846 \\
Median $r$           & 0.861 & 0.910 \\
$r$ q05              & 0.273 & 0.470 \\
Pixel NN             & 3.087 & 3.833 \\
LPIPS NN             & 0.0216 & 0.0238 \\
\bottomrule
\end{tabular}
\end{table}

This stronger separation from the training set comes with a distributional trade-off: FID against the full 4,096-image CelebA training set is 54.89 for standard training and 59.77 for GeoReg. Removing memorized generations from the
standard model increases its FID to $59.60$, indicating that reproduction
memorization contributes partially to the FID difference.

\begin{figure*}[t]
    \centering
    \includegraphics[width=0.92\textwidth]
    {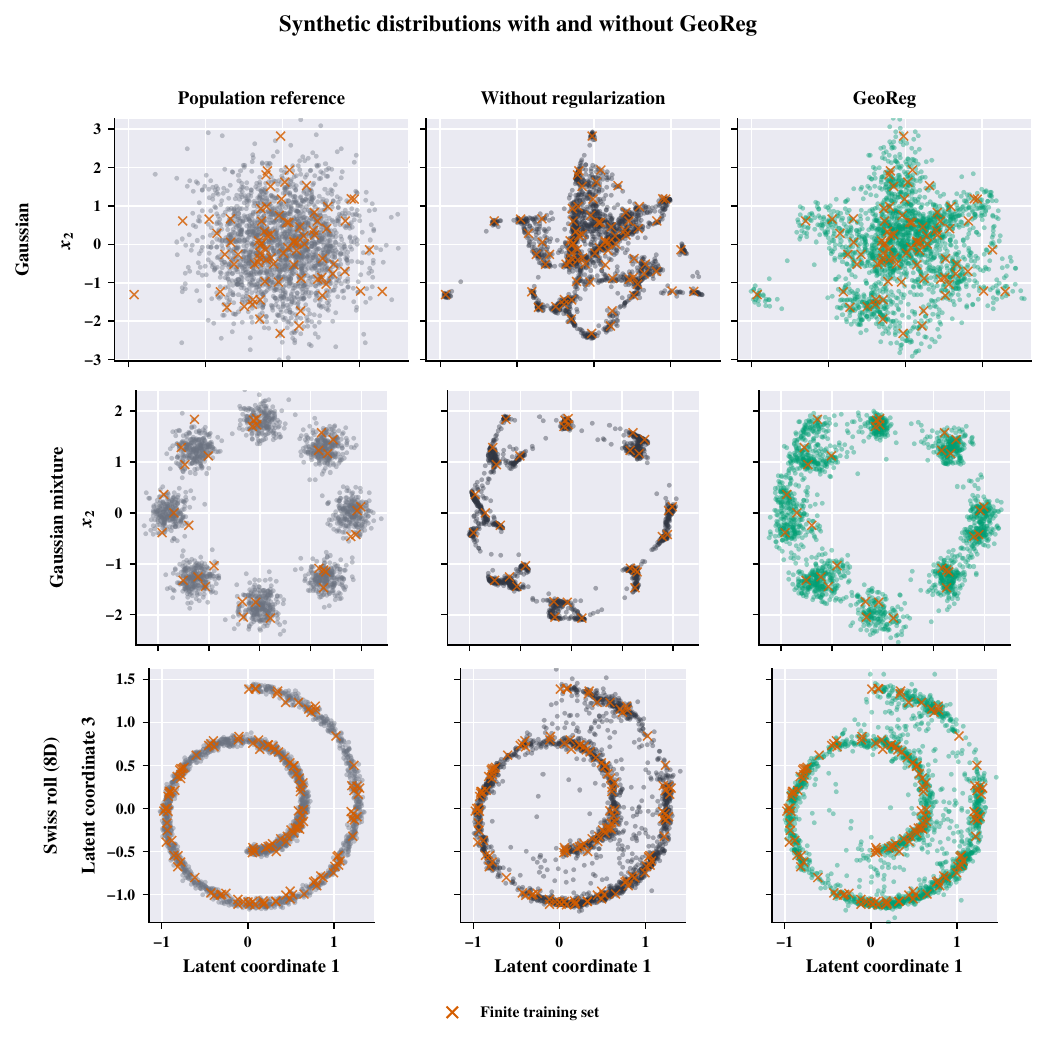}
\caption{\textbf{GeoReg improves population coverage on synthetic data.}
Rows: Gaussian80, Mixture32, and SwissRoll8D128 (shown in a 2D
projection). Columns: samples from the underlying population (left),
generations from standard training (middle), and generations from
GeoReg (right); orange crosses mark the fixed training points. Standard
training concentrates mass around individual training points, whereas
GeoReg spreads generations over the population while preserving its
structure.}
    \label{fig:georeg_synthetic_dist}
\end{figure*}

The synthetic experiments allow us to examine the quality-memorization trade-off more directly
because samples can be compared with the known population rather than only
with a finite image reference set. Figure~\ref{fig:georeg_synthetic_dist}
shows the population, standard-model generations, and GeoReg generations for
Gaussian80, Mixture32, and SwissRoll8D128. Standard training places
substantial probability mass around the finite training examples, most
visibly on Gaussian80 and Mixture32. GeoReg spreads generations more broadly
over the population while preserving its overall structure.

\subsection{Spectral Effect of GeoReg-DM}
\label{app:georeg_spectral}

\begin{figure}[t]
  \centering
  \includegraphics[width=0.75\linewidth]{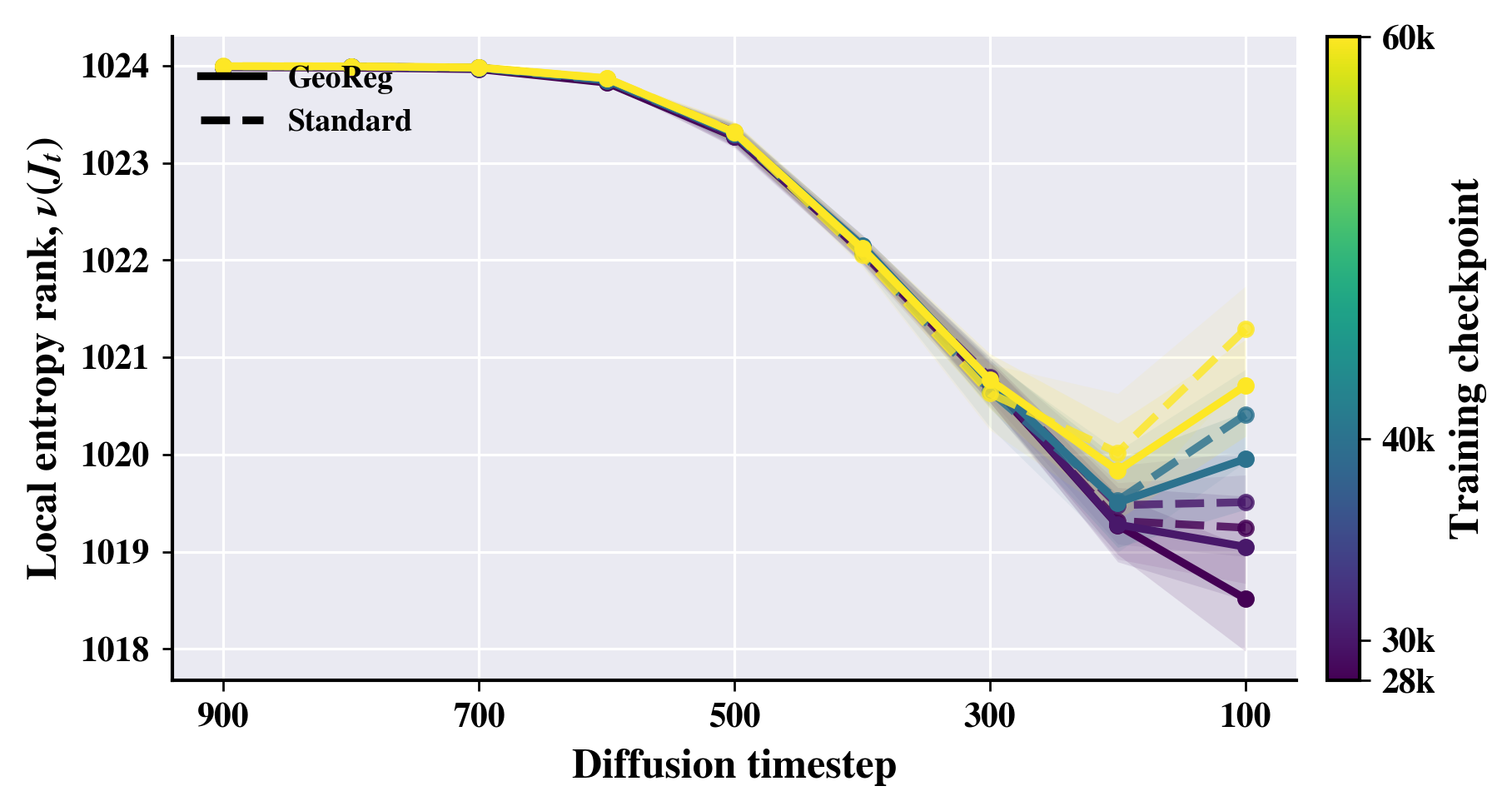}
  \caption{\textbf{Entropy effective rank under \textsc{GeoReg-DM}.}
  Local entropy effective rank $\nu(J_t)$ across diffusion time for \textsc{GeoReg-DM} (solid) and standard training (dashed) across checkpoints. Profiles align at higher noise and diverge near low-noise endpoints, where standard training exhibits stronger rank recovery. Intermediate-noise changes are reported in Table~\ref{tab:georeg_spectral}.}
  \label{fig:georeg_effective_rank}
\end{figure}

The main motivation for \textsc{GeoReg-DM} is the spectral organization that emerges as memorization develops during training. In this section, we investigate how our regularizer impacts this geometry. We evaluate \textsc{GeoReg-DM} and standard training across different fixed noised training states per timestep and checkpoint. Table~\ref{tab:georeg_spectral} reports paired differences ($\text{GeoReg} - \text{Standard}$), and Figure~\ref{fig:georeg_effective_rank} shows the corresponding local entropy rank profiles across diffusion timesteps.

\paragraph{\textsc{GeoReg-DM} reduces intermediate-noise spectral concentration.}
The clearest change appears in stable rank at intermediate noise levels. At $t=300$, \textsc{GeoReg-DM} systematically increases stable rank across all evaluated checkpoints, with mean paired differences ranging from $+34.6$ to $+95.8$ (Table~\ref{tab:georeg_spectral}). At $t=200$, this effect peaks at the 28k checkpoint ($+122.2$) and remains positive through 60k steps. The intervention thus counteracts spectral concentration in the exact intermediate-noise regime where standard training forms a dominant spectral mode. 

\paragraph{\textsc{GeoReg-DM} attenuates low-noise rank recovery.}
At lower noise ($t=100$), \textsc{GeoReg-DM} yields lower entropy effective rank than standard training across every checkpoint, with paired shifts between $-0.45$ and $-0.73$. Whereas standard training recovers toward a high-rank spectrum at low noise (Figure~\ref{fig:georeg_effective_rank}), \textsc{GeoReg-DM} systematically tempers this endpoint rank recovery.

\paragraph{Spectral changes are localized in diffusion time.}
At higher noise ($t \ge 500$), spectral geometry remains closely matched between both models: stable rank differences stay within $11.4$ (in ambient dimension $d=1024$), and entropy rank differences remain under $0.05$. Geometric modifications from \textsc{GeoReg-DM} are therefore tightly localized to the intermediate- and low-noise regimes implicated by memorization dynamics rather than applied uniformly across diffusion time.

In summary, \textsc{GeoReg-DM} weakens the intermediate-noise spectral concentration and attenuates subsequent low-noise rank recovery which empirically coincides with lower  memorization.

\begin{table}[t]
  \centering
  \caption{\textbf{Paired spectral differences under \textsc{GeoReg-DM}.}
  Mean paired differences ($\text{GeoReg} - \text{Standard}$) evaluated on 32 fixed noised training states per cell; brackets report 95\% bootstrap intervals. Bold entries denote intervals excluding zero. SR denotes stable rank, $\nu$ entropy effective rank, and $\Delta_{12}$ the relative leading spectral gap.}
  \label{tab:georeg_spectral}
  \small
  \setlength{\tabcolsep}{4pt}
  \begin{tabular}{lcccccc}
    \toprule
    & \multicolumn{4}{c}{$\Delta$SR} & $\Delta\nu$ & $\Delta\Delta_{12}$ \\
    \cmidrule(lr){2-5} \cmidrule(lr){6-6} \cmidrule(lr){7-7}
    Ckpt & $t=100$ & $t=200$ & $t=300$ & $t=400$ & $t=100$ & $t=200$ \\
    \midrule
    28k & $\mathbf{+77.4}$ & $\mathbf{+122.2}$ & $\mathbf{+95.8}$ & $\mathbf{+15.9}$ & $\mathbf{-0.73}$ & $\mathbf{-0.058}$ \\
        & {\scriptsize[63.7, 92.1]} & {\scriptsize[93.0, 153.1]} & {\scriptsize[62.0, 127.1]} & {\scriptsize[10.3, 22.2]} & {\scriptsize[$-0.86$, $-0.60$]} & {\scriptsize[$-0.090$, $-0.028$]} \\[2pt]
    30k & $\mathbf{+14.2}$ & $\mathbf{+25.4}$ & $\mathbf{+34.6}$ & $+2.3$ & $\mathbf{-0.46}$ & $-0.005$ \\
        & {\scriptsize[3.3, 25.0]} & {\scriptsize[9.2, 42.4]} & {\scriptsize[18.1, 52.2]} & {\scriptsize[$-3.6$, 8.5]} & {\scriptsize[$-0.52$, $-0.39$]} & {\scriptsize[$-0.025$, 0.014]} \\[2pt]
    40k & $+3.3$ & $\mathbf{+25.6}$ & $\mathbf{+43.3}$ & $+3.7$ & $\mathbf{-0.45}$ & $-0.003$ \\
        & {\scriptsize[$-8.5$, 15.0]} & {\scriptsize[7.9, 42.0]} & {\scriptsize[25.8, 61.8]} & {\scriptsize[$-3.7$, 11.6]} & {\scriptsize[$-0.52$, $-0.39$]} & {\scriptsize[$-0.028$, 0.022]} \\[2pt]
    60k & $-6.8$ & $+8.4$ & $\mathbf{+44.6}$ & $+20.3$ & $\mathbf{-0.58}$ & $+0.003$ \\
        & {\scriptsize[$-17.4$, 4.8]} & {\scriptsize[$-10.6$, 28.0]} & {\scriptsize[12.8, 78.0]} & {\scriptsize[$-1.0$, 46.8]} & {\scriptsize[$-0.65$, $-0.52$]} & {\scriptsize[$-0.026$, 0.032]} \\
    \bottomrule
  \end{tabular}
\end{table}

\subsection{Validation of the JER Spectral Proxy}
\label{app:jer_proxy_validation}

JER replaces the nuclear norm in the R\'enyi-2 effective rank with an oriented trace, yielding a quantity that can be estimated efficiently from
Jacobian contractions. This approximation is useful only if it preserves the spectral-dimensionality signal of the exact quantity. We therefore test this directly, independently of memorization labels and detector performance, on 128 frozen query states for which the full Jacobian spectrum can be computed.

We compare
the oriented trace-to-Frobenius proxy with the exact R\'enyi-2 effective rank
defined in Appendix~\ref{app:jer_derivation}.

As shown in Fig.~\ref{fig:jer_proxy_validation}, the oriented
trace-to-Frobenius proxy preserves the ordering of the exact R\'enyi-2
effective rank across query states, with Spearman correlations of $0.994$ for
DDIM and $0.978$ for rectified flow. The proxy is conservative in magnitude,
as expected from replacing the nuclear norm by the oriented trace, but the
relationship remains strongly monotonic across both memorized and
non-memorized states. Memorized states lie toward the high-rank end of the
distribution and particularly close to the equality line. Thus, although JER
is not intended as an unbiased numerical estimator of
$\operatorname{ER}_2$, it retains the spectral ordering relevant for
discrimination while avoiding an explicit spectral decomposition.

\begin{figure}[t]
    \centering
    \includegraphics[width=0.85\linewidth]
    {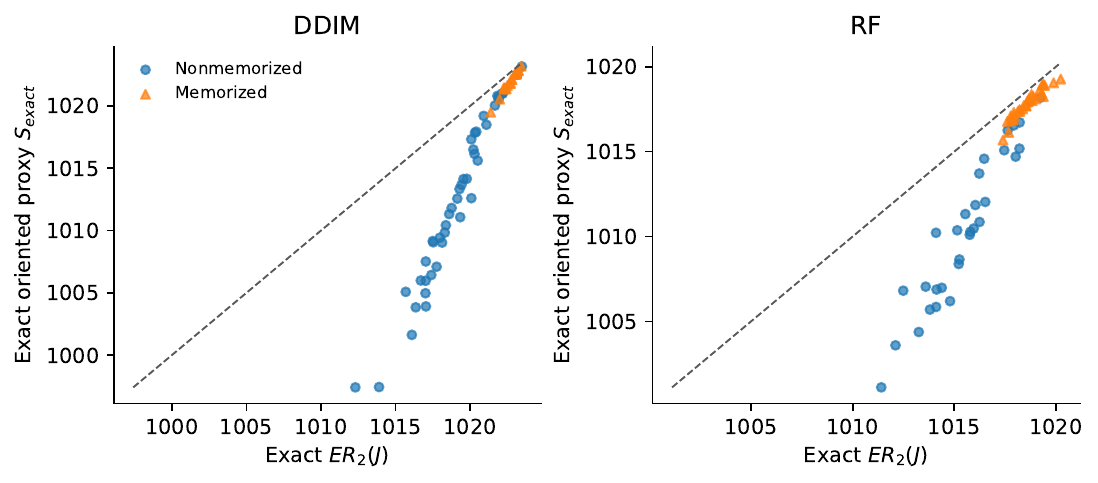}
   \caption{\textbf{Validation of the spectral proxy underlying JER.}
Exact R\'enyi-2 effective rank (horizontal axis) versus the deterministic
oriented trace-to-Frobenius proxy (vertical axis) on frozen DDIM and
rectified-flow query states. Colors distinguish memorized and non-memorized
generations, and the dashed line denotes equality. The proxy is conservative
in magnitude but closely preserves the ordering of the exact spectral
quantity across states.}
    \label{fig:jer_proxy_validation}
\end{figure}

\subsection{Detection Across Training Checkpoints}
\label{app:jer_checkpoints}

We additionally evaluate whether JER remains informative as memorization
develops during training. We repeat the DDIM detection protocol of
Appendix~B.3 at checkpoints from \texttt{60k} to \texttt{260k}, using
disjoint calibration and held-out sets of 1,000 trajectories at each
checkpoint. Across this interval, held-out memorization prevalence increases
from $7.7\%$ to $52.9\%$.

As shown in Table~\ref{tab:jer_checkpoints}, JER maintains consistently high
AUROC across checkpoints and outperforms FLIPD and score magnitude throughout
training. The low-prevalence setting provides a particularly stringent test:
at \texttt{60k}, JER reaches $0.504$ AUPRC against a memorization prevalence
of only $0.077$, compared with $0.252$ for FLIPD and $0.371$ for score
magnitude. AUPRC naturally increases as memorized generations become more
prevalent, whereas the consistently high AUROC shows that JER retains its
discriminative signal across stages of training.

\begin{table}[h]
\caption{\textbf{DDIM memorization detection across training.}
Held-out AUROC / AUPRC at successive model checkpoints. Prevalence is the
fraction of memorized generations in the held-out set; \texttt{160k}
corresponds to the DDIM evaluation in Table~2.}
\label{tab:jer_checkpoints}
\centering
\small
\begin{tabular}{lcccc}
\toprule
Ckpt & Prevalence & JER & FLIPD & Score magnitude \\
\midrule
60k  & 7.7\%  & \textbf{0.940} / \textbf{0.504} & 0.802 / 0.252 & 0.834 / 0.371 \\
100k & 21.9\% & \textbf{0.943} / \textbf{0.793} & 0.819 / 0.571 & 0.811 / 0.591 \\
160k & 33.4\% & \textbf{0.956} / \textbf{0.895} & 0.845 / 0.722 & 0.860 / 0.785 \\
260k & 52.9\% & \textbf{0.957} / \textbf{0.951} & 0.922 / 0.921 & 0.907 / 0.923 \\
\bottomrule
\end{tabular}
\end{table}

\subsection{Stable Diffusion Detection and Conditioning Sensitivity}
\label{app:sd_detection}

We further evaluate JER on Stable Diffusion v1.4 using released collections
of memorized and non-memorized prompts. Unlike the controlled experiments,
the target in this benchmark is membership in these prompt collections rather
than whether an individual generation reproduces a training example. This
distinction is important because several diffusion memorization detectors
exploit atypical conditional responses associated with memorized prompts.

\paragraph{Protocol.}
The evaluation contains 1,000 prompts, equally divided between the released
memorized and non-memorized collections. We freeze 200 prompts for calibration
and evaluate on the remaining balanced set of 800 prompts. All detectors use
identical Stable Diffusion v1.4 and DDIM generation settings. We report held-out AUROC with 95\% confidence intervals obtained from 10,000 stratified bootstrap resamples over prompts, together with the maximum
true-positive rate at FPR $\leq1\%$.

We evaluate two conditioning conventions. With \emph{masked} conditioning,
the tokenizer padding attention mask is passed to the CLIP text encoder. With
\emph{no-mask} conditioning, padding positions remain visible and the
post-EOS token tail can contribute to the conditioning representation.
Prompts, labels, splits, sampling schedules, and detector definitions are
otherwise unchanged.

\begin{table*}[t]
\centering
\caption{\textbf{Stable Diffusion prompt-level memorization detection.}
Results are evaluated on a frozen balanced set of 800 prompts. Confidence
intervals are highlighted in the brackets.
TPR@1\% is the maximum TPR at FPR $\leq1\%$. In the masked setting, padding positions are excluded from
the CLIP text encoding, whereas in the no-mask setting they remain visible,
allowing the post-EOS token tail to influence the conditioning representation.}
\label{tab:sd_detection}
\small
\setlength{\tabcolsep}{4pt}
\begin{tabular}{lcccc}
\toprule
& \multicolumn{2}{c}{Masked}
& \multicolumn{2}{c}{No mask} \\
\cmidrule(lr){2-3}
\cmidrule(lr){4-5}
Detector
& AUROC (95\% CI)
& TPR@1\%
& AUROC (95\% CI)
& TPR@1\% \\
\midrule
Conditional JER (ours)
& \textbf{0.712} [0.677, 0.747]
& 30.3\%
& 0.974 [0.967, 0.979]
& 75.5\% \\
Unconditional JER  (ours)
& 0.671 [0.633, 0.706]
& 23.8\%
& 0.914 [0.908, 0.921]
& 51.8\% \\
Conditional score norm
& 0.513 [0.474, 0.555]
& 1.8\%
& 0.102 [0.080, 0.125]
& 0.0\% \\
Unconditional score norm
& 0.701 [0.695, 0.708]
& 16.3\%
& 0.807 [0.777, 0.837]
& 10.3\% \\
FLIPD \cite{ross2025geometric}
& 0.494 [0.454, 0.534]
& 1.0\%
& 0.802 [0.770, 0.832]
& 6.5\% \\
Score-difference norm \cite{wen2024detecting}
& 0.668 [0.630, 0.705]
& 27.5\%
& 0.985 [0.976, 0.992]
& 90.8\% \\
Hessian score difference \cite{jeon2024understanding}
& 0.664 [0.624, 0.703]
& 34.8\%
& 0.990 [0.983, 0.995]
& 91.5\% \\
Anisotropy \cite{asthana2026detecting}
& 0.680 [0.643, 0.716]
& 21.3\%
& 0.989 [0.982, 0.994]
& 91.3\% \\
\bottomrule
\end{tabular}
\end{table*}

\paragraph{Detection at Stable Diffusion scale.}
Under masked conditioning, conditional JER reaches $0.712$ AUROC and
$30.3\%$ TPR at $1\%$ FPR, compared with $0.671$ for unconditional JER, and near-chance performance for the corresponding
conditional score norm ($0.513$). Thus, the local spectral statistic remains
informative at Stable Diffusion scale and captures structure beyond prediction
magnitude alone. At the stringent operating point, its $30.3\%$ TPR is also
close to the strongest detector in this setting, the Hessian score-difference
statistic at $34.8\%$.

\paragraph{What does the prompt benchmark measure?}
Performance changes markedly when the post-EOS conditioning tail is exposed.
Conditional JER rises to $0.974$ AUROC, while score-difference, Hessian, and
anisotropy statistics reach $0.985$--$0.990$. This sensitivity is relevant to
the interpretation of the benchmark. Many diffusion memorization detectors
operate by identifying unusually strong or anisotropic conditional responses:
such responses can characterize prompts associated with memorization, but do
not by themselves establish that a particular generated image copies a
training example. Indeed, the released prompt collections are themselves
highly distinguishable from text alone: a character $3$--$5$-gram classifier
reaches $0.986$ AUROC and uppercase-character count reaches $0.968$.

The conditioning comparison provides a concrete example of how this prompt
structure can enter model-based scores. Masked and no-mask CLIP
representations agree through the first EOS token, while the post-EOS tail
remains available to U-Net cross-attention when it is not masked. Removing
this tail at cross-attention reduces fixed-state epsilon-difference
discrimination from $0.970$ to $0.774$, close to its masked value of $0.783$.
We therefore view the masked results as the more conservative evidence that
JER transfers to a large pretrained diffusion model. The substantially higher
no-mask results also expose a limitation of prompt-list evaluation: success
can depend on prompt-source and conditioning structure whose relation to
generation-level copying may not transfer to less curated prompt
distributions.

\subsection{\textsc{GeoSteer-DM} Across Training Checkpoints}
\label{app:geosteer_celeba}

Table~\ref{tab:celeba_geosteer_all_checkpoints} provides the complete full-Jacobian \textsc{GeoSteer-DM} evaluation on CelebA across training checkpoints. Beyond reducing aggregate memorization rates, steering systematically shifts the full gap-ratio distribution away from training references: at 300k steps, the median gap ratio $r$ increases from $0.31$ to $0.90$ for DDIM and from $0.26$ to $0.85$ for rectified flow. 

\begin{figure}[h]
\centering
\includegraphics[width=0.32\linewidth]{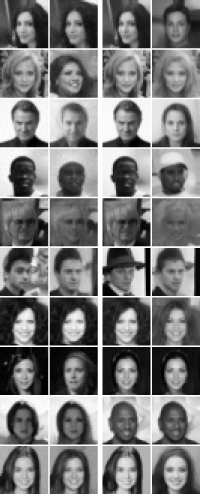}
\caption{\textbf{Qualitative examples of geometric steering on CelebA.}
Matched generations at checkpoint 160k. Columns, left to right: DDIM,
DDIM + GeoSteer, rectified flow, rectified flow + GeoSteer. Each steered
sample starts from the same initial noise as the unsteered sample to its
left. All unsteered samples are memorized ($r<1/3$), and all steered
samples are not ($r>1/3$). Steering redirects the trajectory to a
distinct, plausible CelebA generation, which can change identity
attributes.}
\label{fig:geosteer_celeba_examples}
\end{figure}

While steering increases FID across all checkpoints, this reflects a known artifact of the metric: because FID is computed using 500 generations against a fixed 1,024-image reference set, exact training-set reproduction artificially decreases FID scores (Appendix~\ref{app:georeg_quality}). We therefore treat FID as a secondary metric in this controlled setting. Qualitative examples at 160k steps (Figure~\ref{fig:geosteer_celeba_examples}) confirm that \textsc{GeoSteer-DM} redirects memorized trajectories initialized from identical noise toward distinct, plausible CelebA samples with altered facial attributes.

\begin{table}[h]
\caption{\textbf{GeoSteer on CelebA across training checkpoints.} Each
row reports unmodified sampling $\to$ GeoSteer on 500 paired generations
(five seeds, 100 samples each) sharing initial noise. FID is computed from 500
generations per arm against a fixed 1,024-image CelebA reference set
(Appendix~E.4).}
\label{tab:celeba_geosteer_all_checkpoints}
\centering
\small
\begin{tabular}{llcccc}
\toprule
Sampler & Ckpt & Copy (\%) $\downarrow$ & Mean $r$ $\uparrow$ & Median $r$ $\uparrow$ & FID $\downarrow$ \\
\midrule
DDIM & 30k  & 1.4 $\to$ 1.0  & 0.85 $\to$ 0.89 & 0.91 $\to$ 0.94 & 64.9 $\to$ 74.0 \\
     & 60k  & 8.4 $\to$ 3.2  & 0.76 $\to$ 0.85 & 0.84 $\to$ 0.92 & 60.9 $\to$ 66.6 \\
     & 100k & 24.6 $\to$ 6.0 & 0.63 $\to$ 0.83 & 0.69 $\to$ 0.90 & 56.5 $\to$ 64.6 \\
     & 160k & 36.4 $\to$ 6.8 & 0.55 $\to$ 0.83 & 0.54 $\to$ 0.91 & 52.1 $\to$ 65.5 \\
     & 200k & 42.0 $\to$ 4.4 & 0.52 $\to$ 0.84 & 0.47 $\to$ 0.91 & 52.1 $\to$ 67.0 \\
     & 300k & 51.6 $\to$ 7.6 & 0.45 $\to$ 0.82 & 0.31 $\to$ 0.90 & 47.2 $\to$ 61.3 \\
\midrule
RF   & 30k  & 7.8 $\to$ 0.4  & 0.76 $\to$ 0.89 & 0.85 $\to$ 0.93 & 39.3 $\to$ 45.4 \\
     & 60k  & 26.8 $\to$ 1.6 & 0.62 $\to$ 0.85 & 0.69 $\to$ 0.91 & 38.0 $\to$ 47.5 \\
     & 100k & 36.6 $\to$ 3.4 & 0.52 $\to$ 0.81 & 0.45 $\to$ 0.89 & 34.8 $\to$ 45.3 \\
     & 160k & 54.6 $\to$ 6.8 & 0.43 $\to$ 0.79 & 0.30 $\to$ 0.87 & 34.0 $\to$ 41.8 \\
     & 200k & 52.4 $\to$ 7.4 & 0.42 $\to$ 0.77 & 0.30 $\to$ 0.85 & 33.6 $\to$ 41.7 \\
     & 300k & 60.2 $\to$ 8.0 & 0.40 $\to$ 0.78 & 0.26 $\to$ 0.85 & 33.5 $\to$ 42.5 \\
\bottomrule
\end{tabular}
\end{table}

\subsection{Computational Cost of GeoSteer-DM}
\label{app:geosteer_resolution}

We evaluate how the dimensionality of the Jacobian sketch affects the
steering signal by varying $k\in\{8,16,32,64\}$ and comparing against the
full-Jacobian intervention. Table~\ref{tab:geosteer_resolution} reports
memorization rates pooled across the six controlled-model checkpoints,
together with the median runtime of each sketched steering configuration.

Increasing geometric resolution generally improves steering effectiveness.
For rectified flow, memorization decreases consistently from $32.7\%$ at
$k=8$ to $14.0\%$ at $k=64$, compared with $6.3\%$ using the full
Jacobian. DDIM exhibits the same overall dependence on resolution, although
the smaller sketches are not monotonically ordered: memorization remains
near $18$--$20\%$ for $k\in\{8,16,32\}$ before decreasing to $10.0\%$ at
$k=64$, compared with $3.0\%$ under full-spectrum steering.

The computational cost varies comparatively little across the tested sketch
dimensions. Median runtime ranges from $3.47$--$3.82$\,s for DDIM and
$5.96$--$6.84$\,s for rectified flow. Thus, increasing $k$ substantially
improves the geometric resolution available to the intervention without a
corresponding increase of the same magnitude in runtime. Full-Jacobian
timings were not recorded in this benchmark and are therefore not included
in the runtime comparison.

\begin{table}[t]
\centering
\caption{\textbf{Effect of geometric resolution on steering.}
Memorization rates are pooled across the six controlled-model checkpoints.
Runtime denotes the median wall-clock time per evaluated steering arm across
checkpoints. Full-spectrum timing was not recorded in this benchmark.}
\label{tab:geosteer_resolution}
\small
\begin{tabular}{lcccc}
\toprule
Geometry
& DDIM mem. $\downarrow$
& RF mem. $\downarrow$
& DDIM time (s) $\downarrow$
& RF time (s) $\downarrow$ \\
\midrule
$k=8$
& 18.0\%
& 32.7\%
& 3.50
& 5.96 \\
$k=16$
& 19.7\%
& 24.7\%
& 3.47
& 6.29 \\
$k=32$
& 19.0\%
& 18.0\%
& 3.52
& 6.21 \\
$k=64$
& 10.0\%
& 14.0\%
& 3.82
& 6.84 \\
Full spectrum
& \textbf{3.0\%}
& \textbf{6.3\%}
& -- & -- \\
\bottomrule
\end{tabular}
\end{table}
\subsection{Direction-Specificity of \textsc{GeoSteer-DM}}
\label{app:geosteer_random_control}

A key control question is whether the anti-memorization effect of \textsc{GeoSteer-DM} arises specifically from the identified spectral direction, or whether perturbing trajectories by a comparable magnitude produces a similar reduction. We test this hypothesis using a matched random-direction control that preserves the instantaneous magnitude and rotation angle of \textsc{GeoSteer-DM} while randomizing its orientation.

At each active sampler step, let $\delta = x_{\mathrm{base}} - x_{\mathrm{anchor}}$ denote the unsteered model displacement, and let $\delta_{\mathrm{geo}}$ denote the steered displacement. We construct a random control matching both $\|\delta_{\mathrm{geo}}\|_2$ and the deflection angle $\theta = \angle(\delta, \delta_{\mathrm{geo}})$. Sampling a random Gaussian direction $q$, projecting it onto the orthogonal complement of $\delta$, and normalizing yields:
\begin{equation}
\delta_{\mathrm{rand}} = \|\delta_{\mathrm{geo}}\|_2 \left[ \cos\theta \, \widehat{\delta} + \sin\theta \, q \right], \qquad q^\top \widehat{\delta} = 0.
\label{eq:matched_random_rotation}
\end{equation}
By construction, $\|\delta_{\mathrm{rand}}\|_2 = \|\delta_{\mathrm{geo}}\|_2$ and $\angle(\delta, \delta_{\mathrm{rand}}) = \angle(\delta, \delta_{\mathrm{geo}})$, matching the Euclidean edit magnitude $\|\delta_{\mathrm{rand}} - \delta\|_2$ to numerical precision. All intervention windows, initial noise vectors, and sampler settings remain identical, with local geometry recomputed dynamically along each edited trajectory.

We evaluate this control on 500 paired trajectories at the \texttt{300k} checkpoint. For DDIM, \textsc{GeoSteer-DM} reflects across the leading direction of the full noise-prediction Jacobian over steps 60--98. 

As detailed in Table~\ref{tab:geosteer_random_control}, matching intervention magnitude alone is insufficient to escape memorization basins. For DDIM, baseline memorization ($51.6\%$) remains essentially unchanged under matched random rotation ($50.8\%$), whereas \textsc{GeoSteer-DM} reduces it to $7.6\%$. Similarly, for rectified flow, memorization shifts from $60.2\%$ to $58.6\%$ under random control, compared to $8.0\%$ under \textsc{GeoSteer-DM}. Paired bootstrap analysis confirms this gap: relative to the random control, \textsc{GeoSteer-DM} reduces copying by $43.16$ percentage points on DDIM (95\% CI $[-47.8, -38.4]$) and $50.58$ percentage points on rectified flow (95\% CI $[-55.4, -46.0]$). In contrast, the random rotation induces negligible change relative to baseline (DDIM: $-0.81\%$, 95\% CI $[-2.6, 1.0]$; RF: $-1.59\%$, 95\% CI $[-3.2, 0.0]$).

Crucially, these interventions are locally small: the median deflection angle is only $4.00^\circ$ for DDIM and $4.06^\circ$ for rectified flow. Applying these minor rotations along the leading spectral direction systematically redirects trajectories away from memorization basins, whereas equally strong rotations in random planes produce almost no effect. This confirms that the anti-memorization property of \textsc{GeoSteer-DM} is strictly direction-specific.

\begin{table}[t]
\centering
\caption{\textbf{Direction-specificity of geometric steering.}
Matched comparison at the \texttt{300k} checkpoint across 500 trajectories per sampler. The random-plane control matches the instantaneous displacement norm and rotation angle of \textsc{GeoSteer-DM} relative to the unmodified update while randomizing the orientation plane.}
\label{tab:geosteer_random_control}
\small
\begin{tabular}{llccc}
\toprule
Sampler & Method & Copy rate $\downarrow$ & Mean $r$ $\uparrow$ & Median $r$ $\uparrow$ \\
\midrule
DDIM & Baseline & 51.6\% & 0.454 & 0.314 \\
& Matched random rotation & 50.8\% & 0.460 & 0.326 \\
& \textsc{GeoSteer-DM} & \textbf{7.6\%} & \textbf{0.821} & \textbf{0.897} \\
\midrule
RF & Baseline & 60.2\% & 0.397 & 0.257 \\
& Matched random rotation & 58.6\% & 0.401 & 0.262 \\
& \textsc{GeoSteer-DM} & \textbf{8.0\%} & \textbf{0.778} & \textbf{0.851} \\
\bottomrule
\end{tabular}
\end{table}

\subsection{Guidance-Residual versus Feature-Jacobian Steering}
\label{app:sd_jg_jf}

At Stable Diffusion scale, multiple local Jacobian operators can provide
directions for geometric steering. We compare the guidance-residual Jacobian
$J_g$ used in our primary method against a feature-space alternative,
\[
J_F
=
\frac{\partial F_c(z_t,t)}{\partial z_t},
\]
where $F_c$ denotes the conditional U-Net mid-block representation. For
$J_F$, we obtain the steering direction from the leading right singular
vector of a fixed rank-$32$ left sketch $WJ_F$. For $J_g$, we use the
matrix-free power iteration described in
Appendix~\ref{app:geosteer_impl}. All other components of the steering rule
are held fixed: both methods act on the classifier-free guidance residual via
\[
g_{\mathrm{geo}}
=
0.7(I-2v_1v_1^\top)g
\]
over denoising-order indices $0$--$17$.

The two operators capture distinct notions of local sensitivity. $J_F$
identifies latent directions that most strongly affect an internal
conditional representation, whereas $J_g$ directly differentiates the
conditional--unconditional residual modified by GeoSteer. Consistent with
this distinction, a separate pilot comparison finds a squared subspace
overlap of only approximately $0.286$ between their leading directions.

Table~\ref{tab:sd_jg_jf} reports a matched full-scale comparison on the
memorized-prompt cohort. Replacing $J_F$ with $J_g$ produces a stronger
reduction in training-reference similarity: mean SSCD decreases from $0.259$
to $0.229$, while the fraction of generations with
$\mathrm{SSCD}\geq0.50$ decreases from $14.15\%$ to $9.60\%$, compared
with $30.30\%$ for unmodified generation. This stronger intervention is
accompanied by a larger reduction in CLIP prompt alignment.

The same comparison on non-memorized prompts shows that neither intervention
is confined to memorized generations. Mean SSCD changes from $0.191$ for the
unmodified sampler to $0.172$ under $J_F$ and $0.167$ under $J_g$, while
CLIP changes from $0.268$ to $0.257$ and $0.251$, respectively. Thus,
directly differentiating the guidance residual yields a stronger intervention
on the memorized cohort, but also produces measurable changes on
non-memorized prompts.
\begin{table}[t]
\centering
\caption{\textbf{Stable Diffusion steering with alternative geometric operators.}
Results on 500 memorized prompts with four paired generations per prompt.
$J_F$ denotes the conditional mid-block feature Jacobian, while $J_g$
denotes the Jacobian of the classifier-free guidance residual. Lower SSCD
indicates lower similarity to memorized training references.}
\label{tab:sd_jg_jf}
\small
\begin{tabular}{lcccc}
\toprule
Method
& Mean SSCD $\downarrow$
& SSCD $\geq 0.50$ $\downarrow$
& CLIP $\uparrow$
& FID $\downarrow$ \\
\midrule
No mitigation
& 0.362
& 30.30\%
& 0.279
& 119.42 \\

$J_F$ GeoSteer
& 0.259
& 14.15\%
& 0.270
& 109.15 \\

$J_g$ GeoSteer
& \textbf{0.229}
& \textbf{9.60\%}
& 0.262
& \textbf{102.74} \\
\bottomrule
\end{tabular}
\end{table}

\subsection{Control Study on Non-Memorized Generations}
\label{app:sd_nonmemorized_control}

Table~\ref{tab:sd_complementarity} shows that $J_g$-based GeoSteer
substantially reduces training-reference similarity on the memorized-prompt
cohort, both independently and when composed with existing mitigation
methods. We next test whether these reductions could instead reflect a
generic tendency to perturb all generations away from their reference
images.

We construct a separate control cohort of 100 prompts that are not memorized
by the unmodified Stable Diffusion model. For these prompts, lower SSCD is not
itself an objective; rather, the measurements quantify how strongly the
interventions change reference similarity when memorization is absent. The
unmodified control cohort has a mean SSCD of $0.191$, a $90$th percentile of
$0.262$, and only $1\%$ of generations above the $\mathrm{SSCD}\geq0.50$
threshold.

Under $J_g$-based GeoSteer, mean SSCD decreases from $0.191$ to $0.167$ and
CLIP alignment from $0.268$ to $0.251$ (Table~\ref{tab:sd_nonmem}). These
changes are substantially smaller than on the memorized cohort, where mean
SSCD decreases from $0.362$ to $0.229$ and the fraction above the
$\mathrm{SSCD}\geq0.50$ threshold decreases from $30.30\%$ to $9.60\%$.
The same qualitative asymmetry is observed when GeoSteer is composed with
the existing mitigation methods.  \textsc{GeoSteer-DM} is not inert on non-memorized prompts, but its impact on reference similarity is substantially stronger on memorized ones. This contrast suggests that the reductions on memorized samples are not driven by generic trajectory noise.

\begin{table*}[t]
\centering
\caption{\textbf{Guidance-residual GeoSteer complements existing Stable
Diffusion mitigation methods.}
Results on the memorized-prompt cohort. The first block compares unmodified
generation with $J_g$-based GeoSteer. In the composition block, each entry
reports the corresponding mitigation method alone $\rightarrow$ the same
method combined with GeoSteer. Lower mean SSCD and a smaller fraction of
generations with $\mathrm{SSCD}\geq0.50$ indicate lower training-reference
similarity.}
\label{tab:sd_complementarity}
\small
\begin{tabular}{lcccc}
\toprule
Method
& Mean SSCD $\downarrow$
& SSCD $\geq 0.50$ $\downarrow$
& CLIP $\uparrow$
& FID $\downarrow$ \\
\midrule

No mitigation
& 0.362
& 30.30\%
& 0.279
& 119.42 \\

$J_g$ GeoSteer
& 0.229
& 9.60\%
& 0.262
& 102.74 \\

\midrule
\multicolumn{5}{l}{\emph{Composition with existing mitigation methods}} \\

Ren et al.\ \cite{ren2024unveiling}
& $0.298 \rightarrow 0.217$
& $13.80\% \rightarrow 3.80\%$
& $0.295 \rightarrow 0.281$
& $93.75 \rightarrow 79.67$ \\

RTA \cite{somepalli2023understanding}
& $0.334 \rightarrow 0.213$
& $23.95\% \rightarrow 5.50\%$
& $0.279 \rightarrow 0.263$
& $95.59 \rightarrow 82.12$ \\

Wen et al.\ \cite{wen2024detecting}
& $0.338 \rightarrow 0.210$
& $26.80\% \rightarrow 6.60\%$
& $0.278 \rightarrow 0.260$
& $117.03 \rightarrow 102.75$ \\

Han et al.\ \cite{han2026adjusting}
& $0.283 \rightarrow 0.184$
& $19.05\% \rightarrow 2.50\%$
& $0.271 \rightarrow 0.257$
& $116.29 \rightarrow 105.97$ \\

\bottomrule
\end{tabular}
\end{table*}

\begin{table}[t]
\centering
\caption{\textbf{Control experiment on non-memorized prompts.}
Results on 100 prompts that are not memorized by the unmodified Stable
Diffusion model.}
\label{tab:sd_nonmem}
\small
\begin{tabular}{lccc}
\toprule
Method
& Mean SSCD
& SSCD $q_{0.90}$
& CLIP \\
\midrule
No mitigation
& 0.191 & 0.262 & 0.268 \\
$J_g$ GeoSteer
& 0.167 & 0.214 & 0.251 \\
\midrule
RTA
& 0.204 & 0.298 & 0.281 \\
RTA + GeoSteer
& 0.183 & 0.238 & 0.271 \\
Wen et al.
& 0.184 & 0.271 & 0.268 \\
Wen et al. + GeoSteer
& 0.178 & 0.228 & 0.264 \\
Han et al.
& 0.176 & 0.226 & 0.255 \\
Han et al. + GeoSteer
& 0.174 & 0.225 & 0.247 \\
\bottomrule
\end{tabular}
\end{table}

\subsection{Separating Geometric Reflection from Guidance Attenuation}
\label{app:sd_reflection_attenuation}

The Stable Diffusion setup in our main experiments applies both a geometric reflection and a global gain reduction to the classifier-free guidance displacement. To isolate these mechanisms, we perform a paired four-arm ablation comparing:
\[
I, \qquad 0.7I, \qquad I - 2P, \qquad 0.7(I - 2P),
\]
where $P$ denotes the rank-one projector onto the leading singular direction of the guidance-residual Jacobian. For every prompt and seed, all four arms are evaluated from identical initial noise states.

We evaluate on a collection of memorized and non-memorized prompt families used in the other experiments. On memorized prompts, adding reflection at a fixed gain of $0.7$ reduces mean SSCD by $0.222$ relative to attenuation alone (95\% paired-bootstrap CI $[-0.377, -0.081]$) and lowers the copying rate by $26.7$ percentage points (95\% CI $[-53.3, -6.7]$), with a corresponding CLIP score shift of $-0.016$ (95\% CI $[-0.042, 0.004]$).

Geometric reflection is also effective without attenuation: compared to the unmodified sampler ($I$), norm-preserving reflection ($I - 2P$) reduces mean SSCD by $0.254$ (95\% CI $[-0.440, -0.076]$) and the copying rate by $33.3$ percentage points (95\% CI $[-60.0, -13.3]$).

In the main experiments, we use $a_n = 0.7$ because it slightly improves prompt alignment over un-attenuated reflection while preserving most of the anti-memorization effect. Because attenuation alone fails to suppress memorization, this confirms that the reduction is driven primarily by geometric reflection.

On non-memorized control prompts, no sample crosses the copying threshold under any arm, and SSCD shifts remain noticeably smaller than on memorized prompts. These results confirm that the observed anti-copying effect is not driven by guidance attenuation alone; rather, geometric reflection provides substantial, independent suppression of training-reference similarity. 
\begin{table}[t]
\centering
\caption{\textbf{Reflection versus attenuation on Stable Diffusion.}
Paired ablation on memorized prompt families.}
\label{tab:sd_reflection_attenuation}
\small
\begin{tabular}{lccc}
\toprule
Method & Mean SSCD $\downarrow$ & Copy rate $\downarrow$ & CLIP $\uparrow$ \\
\midrule
Baseline                 & 0.847 & 93.3\% & 0.325 \\
Attenuation only         & 0.870 & 93.3\% & 0.324 \\
Reflection only          & 0.593 & 60.0\% & 0.302 \\
Reflection + attenuation & 0.649 & 66.7\% & 0.308 \\
\bottomrule
\end{tabular}
\end{table}

\subsection{Qualitative Stable Diffusion Examples}
\label{app:sd_qualitative}

\begin{figure*}[t]
    \centering
    \includegraphics[width=\textwidth]
    {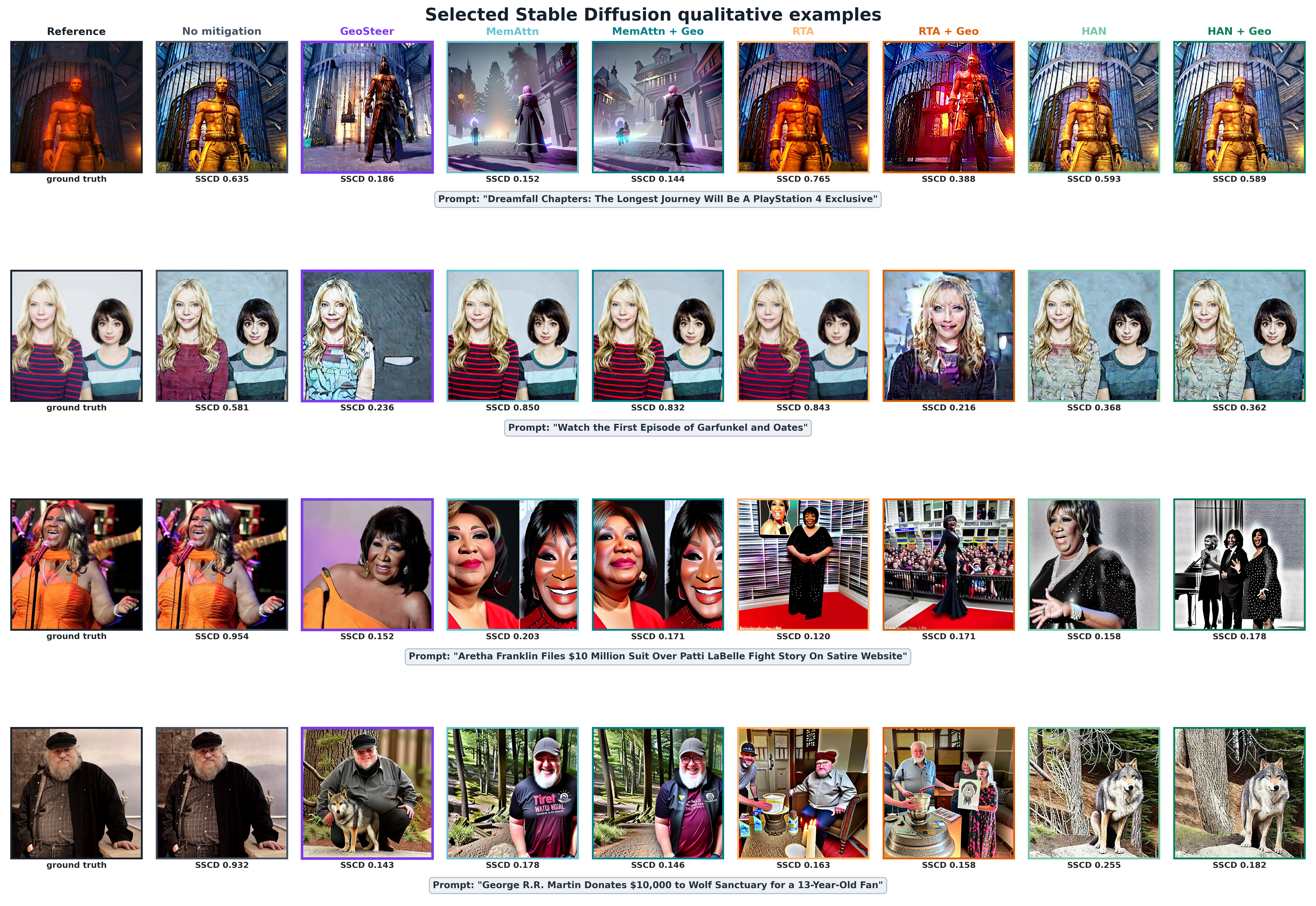}
    \caption{\textbf{Qualitative Stable Diffusion examples.}
    Representative generations from the memorized-prompt evaluation.
    For each example, the training reference, unmodified generation, and
    GeoSteer generation are shown for the same prompt and initial noise.
    GeoSteer reduces training-reference similarity while retaining the
    principal semantic content of the prompt.}
    \label{fig:sd_qualitative}
\end{figure*}

Figure~\ref{fig:sd_qualitative} shows representative paired generations from
the Stable Diffusion evaluation. For each example, the unmodified and
GeoSteer generations use the same prompt and initial noise. GeoSteer moves
outputs away from the memorized training reference while largely preserving
the semantic content of the prompt, complementing the aggregate SSCD and CLIP
results in Sec.~\ref{sec:results_stable_diffusion}.

\end{document}